%% file: main.tex
\documentclass[10pt,journal,compsoc]{IEEEtran}
\usepackage[utf8]{inputenc}
\usepackage[T1]{fontenc}
\usepackage{microtype}
\usepackage{url}
\usepackage[table]{xcolor} 

\usepackage{amsmath}
\usepackage{amssymb}
\usepackage{amsfonts}
\usepackage{amsthm}
\usepackage{nccmath}
\usepackage{mathtools}
\usepackage{nicefrac}
\usepackage{bm}
\usepackage{bbm}
\usepackage{physics}
\usepackage{mathdots}
\usepackage{yhmath}
\usepackage{cancel}
\usepackage{gensymb}
\usepackage{extarrows}
\usepackage{proof}
\usepackage{siunitx}

\usepackage{graphicx}
\usepackage{epstopdf}
\usepackage{float}
\usepackage{caption}
\usepackage{subcaption} 
\usepackage{tikz}
\usepackage{tikz-cd}
\usetikzlibrary{fadings, patterns, shadows.blur, shapes}

\usepackage{booktabs}
\usepackage{multirow}
\usepackage{array}
\usepackage{tabularx}
\usepackage{makecell}

\usepackage[linesnumbered,ruled,vlined]{algorithm2e}
\usepackage{titletoc}
\usepackage{tocloft}
\usepackage{balance}
\usepackage{multicol}
\usepackage{ragged2e}
\usepackage{chngcntr}

\usepackage{lipsum}
\usepackage{atbegshi}
\usepackage{etoolbox}
\usepackage{newclude}
\usepackage{marvosym}
\usepackage{pifont}
\usepackage{listings}
\usepackage{enumitem}
\usepackage[normalem]{ulem}
\usepackage{tcolorbox}
\usepackage{verbatim}

\usepackage{hyperref}

\usepackage{xcolor,colortbl}
\definecolor{ballblue}{rgb}{0.13, 0.67, 0.8}
\definecolor{lightseagreen}{rgb}{0.13, 0.7, 0.67}
\definecolor{darkturq}{HTML}{00CED1}
\definecolor{org}{HTML}{F8A145}
\definecolor{blu}{HTML}{63ACE5}
\definecolor{c1}{HTML}{8ECFC9}
\definecolor{c2}{HTML}{FFBE7A}
\definecolor{c3}{HTML}{FA7F6F}
\definecolor{softred}{HTML}{FE8A71}
\definecolor{softblue}{HTML}{63ACE5}

\definecolor{lightblue}{RGB}{200, 231, 250}
\definecolor{cvprblue}{rgb}{0.21,0.49,0.74}
\definecolor{matplotlibblue}{HTML}{1f77b4}
\definecolor{matplotliborange}{HTML}{ff7f0e}
\definecolor{matplotlibgreen}{HTML}{2ca02c}
\definecolor{matplotlibred}{HTML}{d62728}
\definecolor{matplotlibpurple}{HTML}{9467bd}
\definecolor{matplotlibbrown}{HTML}{8c564b}

\newcommand{\RowStrut}{\rule{0pt}{2.6ex}\rule[-1.0ex]{0pt}{0pt}}
\newcommand{\BoldRowStrut}{\rule{0pt}{3.6ex}\rule[-1.0ex]{0pt}{0pt}}

\definecolor{blue}{RGB}{68,114,196}
\definecolor{orange}{RGB}{237,125,49}
\definecolor{gray}{RGB}{165,165,165}

\hypersetup
{
    colorlinks=true,
    linkcolor=blue,
    citecolor=darkturq 	 	
}

\usepackage{algorithmic}

\usepackage{thm-restate}

\newcommand{\Eqref}[1]{Eq.~(\ref{#1})}
\newcommand{\Tbref}[1]{Tab.~\ref{#1}}
\newcommand{\Fgref}[1]{Fig.~\ref{#1}}

\newcommand{\Algref}[1]{Alg.~\ref{#1}}

\newcommand{\cref}[1]{Sec.~\ref{#1}}

\def \vc {\bm{v}_c}
\def \vp {\bm{v}_p}
\def \vf {\bm{v}_f}

\def \zp {\bm{z}_p}
\def \zf {\bm{z}_f}

\def \Acp {\bm{A}^{c\rightarrow p}}
\def \Acf {\bm{A}^{c\rightarrow f}}
\def \Spc {\bm{S}^{p\rightarrow c}}
\def \Scf {\bm{S}^{c\rightarrow f}}
\def \Qcp {\bm{Q}_c^p}
\def \Kcp {\bm{K}_c^p}
\def \Vcp {\bm{V}_c^p}
\def \Qcf {\bm{Q}_c^f}
\def \Kcf {\bm{K}_c^f}
\def \Vcf {\bm{V}_c^f}

\def \zcsp {\bm{z}_c^p}
\def \zcsf {\bm{z}_c^f}

\def \lpt {\mathcal{L}_{pt}}
\def \lft {\mathcal{L}_{ft}}
\def \lpf {\mathcal{L}_{pf}}

\def \methodname {VT-CoRe}
\def \methodnameb {\textbf{VT-CoRe}}

\definecolor{soft_red}{RGB}{240, 110, 110}
\definecolor{soft_blue}{RGB}{100, 170, 220}
\definecolor{define_red}{RGB}{255, 20, 20}
\definecolor{define_green}{RGB}{55, 100, 0}

\definecolor{responseblue}{RGB}{0,0,255}
\definecolor{responsered}{RGB}{255,0,0}

\usepackage[framemethod=tikz]{mdframed}
\usepackage{lipsum}

\ifCLASSOPTIONcompsoc
  \usepackage[nocompress]{cite}
\else
  \usepackage{cite}
\fi
\ifCLASSINFOpdf
\else
\fi
\begin{document}

\title{Variation-aware Temporal Correspondence for Self-Supervised Video Representation Learning}

%
%
%
%

\author{Yang Liu, 
        Qianqian Xu\IEEEauthorrefmark{1},~\IEEEmembership{Senior Member,~IEEE,} Peisong Wen,\\
        Siran Dai and~Qingming Huang\IEEEauthorrefmark{1},~\IEEEmembership{Fellow,~IEEE}
\IEEEcompsocitemizethanks{
\IEEEcompsocthanksitem This work was supported in part by National Natural Science Foundation of China: 62525212, U23B2051, 62236008, 62441232, 62521007, 62602635, and U21B2038, in part by Youth Innovation Promotion Association CAS, in part by the Strategic Priority Research Program of the Chinese Academy of Sciences, Grant No. XDB0680201, in part by the China National Postdoctoral Program for Innovative Talents under Grant BX20250377, in part by the China Postdoctoral Science Foundation under Grant 2026M791710, and in part by the project ZR2025ZD01 supported by Shandong Provincial Natural Science Foundation.
\IEEEcompsocthanksitem Yang Liu is with the School of Computer Science and Technology, University of Chinese Academy of Sciences, Beijing 101408, China (email: \texttt{liuyang232@mails.ucas.ac.cn}).
\IEEEcompsocthanksitem Qianqian Xu is with the State Key Laboratory of AI Safety, Institute of Computing Technology, Chinese Academy of Sciences, Beijing 100190, China, and also with the Beijing Academy of Artificial Intelligence (email: \texttt{xuqianqian@ict.ac.cn}).
\IEEEcompsocthanksitem Peisong Wen is with the School of Computer Science and Technology, University of Chinese Academy of Sciences, Beijing 101408, China (email: \texttt{wenpeisong@ucas.ac.cn}).
\IEEEcompsocthanksitem Siran Dai is with the Institute of Information Engineering, Chinese Academy of Sciences,
Beijing 100093, China, and also with the School of Cyber Security,
University of Chinese Academy of Sciences, Beijing 100049, China (email: \texttt{daisiran@iie.ac.cn}).
\IEEEcompsocthanksitem Qingming Huang is with the School of Computer Science and Technology, University of Chinese Academy of Sciences, Beijing 101408, China, and also with the State Key Laboratory of AI Safety, Institute of Computing Technology, Chinese Academy of Sciences, Beijing 100190, China (e-mail: \texttt{qmhuang@ucas.ac.cn}).\protect\\
}
\thanks{Corresponding authors: Qianqian Xu and Qingming Huang.}
}

%
%

\markboth{Journal of \LaTeX\ Class Files,~Vol.~14, No.~8, August~2015}%
{Shell \MakeLowercase{\textit{et al.}}: Bare Demo of IEEEtran.cls for Computer Society Journals}
%




\IEEEtitleabstractindextext{%
\justifying

\include{sec/0_abstract}
}

\maketitle
\IEEEdisplaynontitleabstractindextext
\IEEEpeerreviewmaketitle
\include*{sec/1_introduction}

\include*{sec/2_related_work}

\include*{sec/3_method}

\include*{sec/4_experiments}
\include*{sec/5_conclusion}


%



\ifCLASSOPTIONcaptionsoff
  \newpage
\fi



%
\bibliographystyle{abbrv}
\bibliography{main}

%
\input{sec/bio}

\clearpage

\include{sec/X_appendix}

\end{document}

%% file: sec/0_abstract.tex
\begin{abstract}

Self-supervised video representation learning has achieved remarkable progress by leveraging large-scale unlabeled video data.
Recent methods commonly extend \textit{Masked Image Modeling} (MIM) to \textit{Masked Video Modeling} (MVM), where video representations are learned by capturing semantic information from temporal contexts during masked content recovery.
Unlike static images, videos exhibit inherent \textbf{temporal variation}, as visual objects continuously change their positions, appearances, and motion states across frames.
Therefore, MVM needs to preserve semantic consistency while capturing temporal changes of corresponding regions, which raises two central challenges:
\textbf{1)} how to establish reliable temporal correspondence across frames, and
\textbf{2)} how to formulate an appropriate representation learning objective for modeling both semantic consistency and temporal variation.
To address these challenges, we propose \textit{\textbf{V}ariation-aware \textbf{T}emporal \textbf{Co}rrespondence for video \textbf{Re}presentation learning} (\methodnameb{}) framework.
For challenge \textbf{1)}, \methodname{} introduces a sandwich sampling strategy to construct two-sided temporal references and a Patch Registration Module (PRM) to establish reliable patch-level correspondence.
For challenge \textbf{2)}, \methodname{} imposes latent-space semantic alignment on registered representations through a self-distillation architecture and further proposes temporal variation modeling to constrain representation trajectories over time.
Experiments on 8 downstream tasks across 16 benchmarks demonstrate the effectiveness and applicability of \methodname{} across pixel-level, frame-level, video-level, and vision-based robotic learning scenarios.
Complementary representation analysis shows that \methodname{} learns a more structured latent space for variation-aware temporal correspondence learning.

\end{abstract}

%% file: sec/1_introduction.tex
\section{Introduction}
\label{sec:introduction}

Self-supervised representation learning has achieved remarkable progress by effectively leveraging large-scale visual data without human annotations~\cite{MoCov1,BYOL,MAE,ijepa,DINO,iBOT}.
One of the most promising self-supervised paradigms is \textit{\textbf{M}asked \textbf{I}mage \textbf{M}odeling (\textbf{MIM})}~\cite{MAE,ijepa,BEiTv1,BEiTv2}, where the model learns representations by recovering masked image regions from visible context with an encoder-decoder architecture.
This process encourages semantically related visual regions to form consistent representations in the latent space.
Benefiting from the success of MIM, recent methods have extended this paradigm to videos by incorporating temporal contexts from nearby frames for masked content recovery, thereby giving rise to \textit{\textbf{M}asked \textbf{V}ideo \textbf{M}odeling (\textbf{MVM})}~\cite{MAE_ST,VideoMAE,SiamMAE,CropMAE,RSP,vjepa2,vjepa2_1}.

\begin{figure*}
  \centering
    \includegraphics[width=1.0\linewidth]{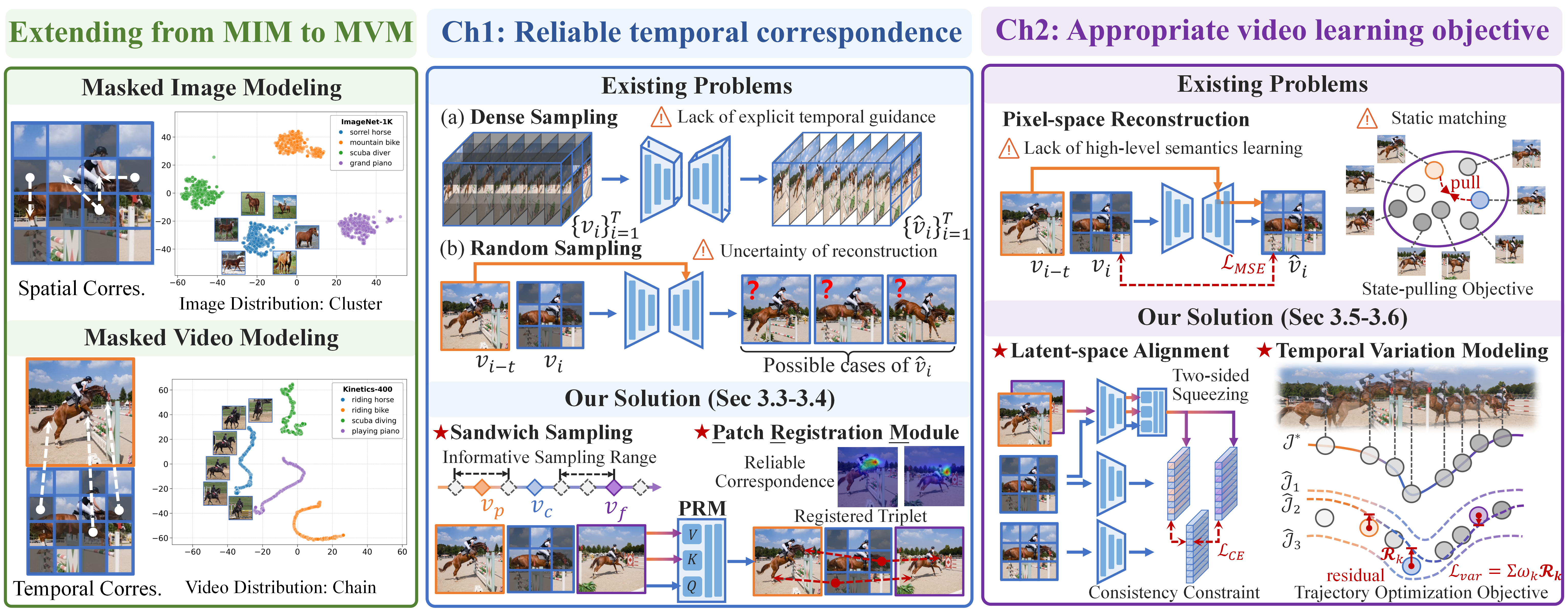}
    \vspace{-3pt}
    \caption{
        Motivation of this work.
        Extending masked image modeling (MIM) to masked video modeling (MVM) requires handling temporal variation beyond static semantic consistency, as video representations tend to evolve along the temporal sequence.
        This extension raises two key challenges: establishing reliable temporal correspondence across frames and designing an appropriate objective for video representation learning.
        \methodname{} addresses these challenges by introducing sandwich sampling and patch registration to construct reliable temporal triplets, and further combines latent-space semantic alignment with variation-aware temporal modeling.
    }
    \label{fig:intro}
    \vspace{-12pt}
\end{figure*}

However, extending masked modeling from images to videos is not straightforward.
In contrast to static images, videos exhibit substantial \textbf{temporal variation}, as visual objects continuously change their positions, appearances, and motion states over time.
When nearby frames are used as temporal contexts, recovering a masked target often relies on temporally corresponding regions across frames, while such correspondence is not directly available in the self-supervised setting, making it difficult to obtain reliable temporal references for recovery.
Therefore, effective MVM should preserve semantic consistency among corresponding regions while capturing their temporal changes.
This requirement raises two central challenges:
\textbf{1)} how to establish \textbf{reliable temporal correspondence} across frames, and \textbf{2)} how to formulate an appropriate \textbf{representation learning objective} for modeling both semantic consistency and temporal variation in videos, as illustrated in \Fgref{fig:intro}.

Existing MVM methods construct temporal context mainly through dense sampling or conditional prediction.
Early efforts primarily follow the MIM design by densely sampling video frames and reconstructing masked video contents with a joint spatiotemporal encoder~\cite{MAE_ST,VideoMAE,VideoMAEv2,BEVT,DropMAE}.
However, without explicit temporal guidance, these methods rely on highly redundant inputs to capture visual co-occurrence patterns, rather than establishing fine-grained correspondence between temporally related regions.
More recent methods have shifted to sparse random sampling, where a conditional frame or visible spatiotemporal context is used as prior information to recover masked targets~\cite{SiamMAE,CropMAE,STP,RSP,vjepa2,vjepa2_1}.
While more efficient, such conditional prediction can be compatible with multiple plausible target states when inferred from the single-sided temporal context, making the induced correspondence less reliable in complex dynamic scenarios~\cite{uncertainty_1,uncertainty_2}.

Regarding the representation learning objective, mainstream MVM methods typically inherit the recovery objective of MIM, where the recovered masked regions are aligned with their original unmasked counterparts~\cite{MAE,ijepa} or with another augmented version~\cite{DINO,iBOT,DINOv2}.
When extended to videos, MVM incorporates nearby temporal context into the masked reconstruction process and aligns the recovered regions with the original observation at the same timestep~\cite{VideoMAE,SiamMAE,CropMAE,RSP,vjepa2,vjepa2_1}.
In this way, the objective mainly encourages representations inferred from temporal context to match the target state, thereby implicitly pulling adjacent temporal states closer to preserve semantic consistency across temporally related regions.
However, such a matching objective is naturally suited for image representation learning, where semantically related regions typically form compact clusters, but can be suboptimal for video representations, where frame representations from the same video are expected to evolve along chain-like structures over time.
This issue becomes more evident in pixel-space reconstruction, where strict pixel matching provides limited information compression for high-level semantic abstraction, thus making it less effective for temporal variation learning.
This suggests that video representation learning should go beyond static matching objectives and further model dynamic changes.

To address challenge \textbf{1)}, we first construct more stable temporal references with a \textbf{sandwich sampling strategy}.
For a masked current frame, we sample two auxiliary frames from its past and future contexts, respectively.
The two-sided references convert masked reconstruction from single-sided inference into temporal interpolation, thereby reducing correspondence uncertainty through bidirectional constraints.
Based on the sampled frames, we further introduce an auxiliary branch and design a \textit{\textbf{P}atch \textbf{R}egistration \textbf{M}odule (\textbf{PRM})} to register past and future patch-level representations to the spatial layout of the current frame, and simultaneously select high-confidence registered patches as reliable temporal references for subsequent representation learning.

To handle challenge \textbf{2)}, we first impose local semantic consistency on the registered representations through DINO-based self-distillation~\cite{DINO,DINOv2}.
The registered past and future representations are aligned with the current representation from the teacher branch, so that temporally corresponding regions preserve essential semantic consistency across nearby frames.
To further model the temporal variation of the representation in the latent space, we use the registered past-current-future triplet to estimate the latent trajectory of a corresponding region.
Since this triplet provides three ordered temporal states, we apply a second-order Taylor expansion around the current state to derive a finite-difference relation among the registered states and construct a residual constraint on the latent trajectory.
The residual is then parameterized with $K$ groups of learnable coefficients to adaptively capture different variation patterns.
Thus, the overall representation learning objective aligns temporally corresponding states for semantic consistency and constrains how the representations temporally change.

By integrating the above components, we propose a self-supervised framework called \textit{\textbf{V}ariation-aware \textbf{T}emporal \textbf{Co}rrespondence for video \textbf{Re}presentation learning (\methodnameb{})}.
\methodname{} integrates temporal correspondence modeling with variation-aware constraints, enabling the representations to preserve local semantic consistency while better reflecting coherent changes across temporally corresponding regions.
Experiments on 8 downstream tasks across 16 benchmarks demonstrate the effectiveness of the proposed framework across pixel-level, frame-level, video-level, and vision-based robotic learning scenarios.
Representation analysis via metrics and visualizations shows that \methodname{} learns a more structured representation space, with favorable properties for variation-aware temporal correspondence learning.

This work extends our previous CVPR 2025 paper~\cite{tcore}, which presents an early exploration of latent-space temporal correspondence for dense self-supervised video representation learning.
This journal version includes systematic improvements in methodology, experiments, and analysis in the following aspects.

\textbf{1) Framework Extension.}
This journal version further extends the framework from local temporal alignment to variation-aware temporal modeling.
To this end, we enhance the correspondence learning module with a \textit{\textbf{P}atch \textbf{R}egistration \textbf{M}odule (\textbf{PRM})} in \cref{subsec:patch_matching}, which registers past and future patch-level representations to the current-frame layout and selects high-confidence registered patches for more reliable temporal correspondence.
Based on the registered triplet representations, we introduce a residual constraint derived from a second-order Taylor expansion in a separate projected space, with learnable parameter groups to adapt to different variation patterns in \cref{subsec:temporal_var}.
These extensions complement local semantic consistency by regularizing how temporally corresponding representations change over time.

\textbf{2) More Comprehensive Evaluation.}
The empirical evaluation is extended from the original 3 dense prediction benchmarks to 16 benchmarks on 8 downstream tasks across pixel-level, frame-level, video-level, and vision-based robotic learning settings in \cref{subsec:main_results}.
The results show consistent improvements on multiple downstream tasks, demonstrating the effectiveness of \methodname{} under different evaluation protocols.
We also provide efficiency analysis and ablation studies in \cref{subsec:efficiency_analysis} and \cref{subsec:ablation_study}, yielding a more complete empirical evaluation of the model design.

\textbf{3) New Representation Analysis.}
Beyond downstream evaluation, this version provides a more detailed analysis of the learned representations in \cref{subsec:representation_analysis}.
We include quantitative metrics and qualitative visualizations to examine whether the representations preserve semantic consistency while reflecting local temporal variation.
The results show that \methodname{} learns a more structured representation space, providing additional evidence for the effectiveness of variation-aware temporal correspondence learning.

%% file: sec/2_related_work.tex
\section{Related Work}
\label{sec:related_work}

\subsection{Self-supervised Video Representation Learning}
\label{subsec:related_work_SSL}

Self-supervised learning aims to learn transferable visual representations from unlabeled data by exploiting the intrinsic structure of visual signals.
Existing methods mainly differ in the design of supervisory signals for capturing spatial appearance and temporal information, leading to several representative pre-training paradigms.

\textbf{Contrastive learning methods.}
The core objective of contrastive learning is to construct a consistent representation space by pulling semantically related instances or views closer while separating unrelated ones.
Foundational image-based methods construct such supervision from augmented image views~\cite{SimCLRv1,MoCov1} or asymmetric network designs~\cite{BYOL,SimSiam}.
Extending these image-based paradigms, video contrastive learning further constructs positive pairs from temporally related frames, clips, or augmented video views~\cite{SlowFast,HiCo_2,ad2}.
By exploiting temporal continuity and appearance correlations in videos, these methods encourage feature consistency across related views to capture robust global semantics, yielding strong performance on multiple discriminative tasks~\cite{Kinetics,SSV2,UCF101,HMDB51}.

\textbf{Masked modeling methods.}
As a dominant paradigm in self-supervised video representation learning, masked modeling aims to learn fine-grained spatiotemporal representations by recovering missing information from partial observations in the pixel space or latent space~\cite{MIM_survey}.
\textbf{1) Pixel-space-oriented methods} reconstruct masked RGB patches with an encoder-decoder structure~\cite{MAE,SimMIM}.
When extended to videos, a common strategy is to mask and reconstruct video tubes, forcing the model to aggregate contextual information across both space and time~\cite{VideoMAE,VideoMAEv2,MAE_ST}.
These methods are typically built upon spatiotemporal Transformer architectures~\cite{TimeSformer,ViViT}, and subsequent studies improve efficiency or reconstruction quality through sparse sampling~\cite{DropMAE,OmniMAE} or attention-based restoration~\cite{SiamMAE,CropMAE,RSP}, allowing missing content to be inferred from limited temporal observations.
Despite their straightforward design, pixel-level reconstruction objectives tend to emphasize low-level visual details, leaving higher-level semantic structures insufficiently modeled.
\textbf{2) Latent-space-oriented methods} reduce the reliance on low-level reconstruction by defining prediction targets in the representation space, where the model is optimized to match stable semantic embeddings instead of raw visual details.
Initially introduced in the image domain, DINO-based methods adopt self-distillation architectures, where a student network is trained under the supervision of a momentum-updated teacher to align semantic representations across different views~\cite{DINO,iBOT,DINOv2,DINOv3,wen2025semantic}.
In the video domain, related methods further introduce temporal modeling to adapt self-distillation architectures to video inputs~\cite{Dino-foresight,DISCOVR}.
Another line of work is Joint-Embedding Predictive Architectures (JEPA)~\cite{ijepa,vjepa2,vjepa2_1}, which predicts the latent representations of randomly sampled target regions from visible context.
By exploiting visual co-occurrence relationships under masking, masked modeling methods learn dense-level representations that support fine-grained video tasks~\cite{DAVIS17,VIP,JHMDB}.

\subsection{Temporal Correspondence Modeling}
\label{subsec:related_work_correspondence}

The temporal continuity of videos provides a natural source of supervision for video representation learning.
A common assumption is that visual entities preserve their semantic identity over short time intervals, allowing temporal correspondence to serve as an effective constraint for representation learning~\cite{continuity,ad1}.
Accordingly, temporal consistency is often enforced by encouraging adjacent frames or corresponding regions to remain close in the latent space.
Early methods localize sparse patch-level or object-level targets, track them forward and backward across frames, and penalize deviations of the returned positions or representations from their initial states~\cite{cycle_patch_1,cycle_patch_2,cycle_patch_3}.
Subsequent work extends this idea to semantically related videos by cyclically propagating frame correspondences across sequences and minimizing mismatches between the resulting alignment distributions~\cite{cycle_match_1,cycle_match_2}.
More recent approaches model dense frame correspondences as transition matrices on space-time graphs and constrain their product along a cyclic walk to approach the identity matrix, returning patches to their starting positions~\cite{cycle_walk_1,cycle_walk_2,cycle_walk_3}.
Related cross-modal work learns correspondences between compressed and raw visual representations through contrastive alignment~\cite{low_level_motion_traj_3,ad5}.
These methods learn temporal correspondence by enforcing local semantic consistency across nearby frames, yet provide limited constraints on longer-range temporal evolution.

To go beyond local consistency preservation, recent work has introduced more explicit constraints on temporal evolution.
Some methods exploit low-level motion cues as auxiliary supervision, such as motion vectors and masked motion targets~\cite{low_level_motion_traj_1,low_level_motion_traj_2}.
Other methods draw on point tracking to learn long-range temporal propagation with point-level correspondence as supervision~\cite{corres_traj_1,corres_traj_2,corres_traj_3}.
More recent studies further model longer-range temporal relations through recurrent state propagation or representation correction over time~\cite{corres_RNN_1,corres_RNN_2,Chirality}.
By modeling temporal evolution more explicitly, these methods improve correspondence learning in complex scenarios.

\subsection{Temporal Evolution of Video Representations}
\label{subsec:related_work_dynamic}

Real-world visual processes usually exhibit coherent changes over short time intervals, making it meaningful to describe video content as evolving states rather than as a collection of independent frames.
In the study of physical systems, ordinary differential equations (ODEs) have long been used to characterize continuous state transitions.
This idea was later extended to continuous-time latent state models, where hidden states are represented as trajectories evolving over time, and further developed into second-order formulations that jointly characterize the state and its temporal derivative~\cite{ode_1,ode_2,ode_3}.
This perspective has gradually influenced video representation learning.
Leveraging video continuity, recent methods exploit frame differences or Taylor expansions to explicitly model inter-frame variations, where temporal differences at different orders can be viewed as discrete approximations of velocity and acceleration, thereby introducing change-sensitive cues into video tasks~\cite{video_ode_1,video_ode_2}.
Further studies combine visual encoders with differentiable ODE solvers to estimate latent states or physical parameters from video observations through latent-space prediction under known second-order ODE systems~\cite{video_ode_physics_1,video_ode_physics_2,video_ode_physics_3}.
These studies motivate modeling temporal variation in video representation learning.
Accordingly, we further constrain the representation trajectories of temporally corresponding regions, encouraging them to preserve semantic consistency while capturing structured variation over time.

\begin{figure*}[htbp]
  \centering
    \includegraphics[width=1.0\linewidth]{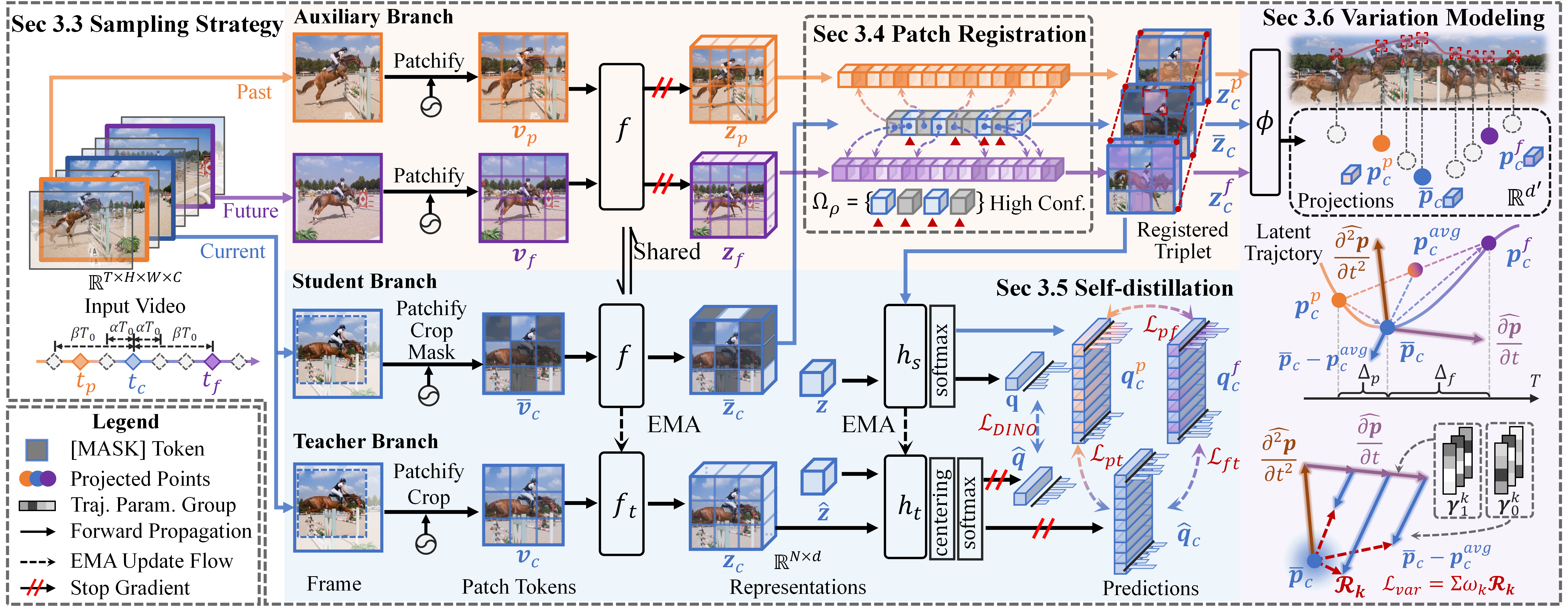}
    \caption{The architecture of the proposed framework. Given a training video, we introduce a sandwich sampling strategy to select past, current, and future frames, with the past and future frames providing two-sided temporal references via the auxiliary branch. We propose the Patch Registration Module to align the past and future representations with the current one, thereby establishing reliable correspondence for latent-space reconstruction. We further project the registered representations into a separate space for variation-aware temporal modeling, where we use a second-order Taylor approximation to construct local representation trajectories with multiple parameter groups to adaptively fit different patterns.}
    \label{fig:pipeline}
    \vspace{-8pt}
\end{figure*}

%% file: sec/3_method.tex
\section{Methodology}
\label{sec:method}

\subsection{Task Definition}
\label{subsec:task_definition}

An input video is represented as a sequence of frames $\bm{V} = \{\bm{v}_j \in \mathbb{R}^{H \times W \times C}\}_{j=1}^{T}$, where $H$, $W$, and $C$ denote the height, width, and channel number of each frame respectively, and $T$ indicates the total number of frames in the video. For clarity, we denote the frame at the $t$-th time step as $\bm{V}(t)$. 
Each frame $\bm{v}$ is divided into $N$ non-overlapping patches with patch size $p \times p$, where each patch contains $p^2$ spatial pixels.
The goal of video representation learning is to obtain a representation for each patch in the video. 

In this work, we focus on a frame-level encoder $f: \mathbb{R}^{H \times W \times C} \rightarrow \mathbb{R}^{N \times d}$, where $d$ denotes the embedding dimension.
Unlike image representation learning, the encoder is pre-trained with video sequences rather than isolated images, enabling the learned representations to encode temporally aware visual information in a self-supervised manner.
After pre-training, the encoder $f$ can be used to initialize various downstream video tasks.

Following the masked video modeling paradigm, the model is required to infer missing information from visible contexts.
Instead of performing reconstruction in the pixel space, we conduct masked reconstruction in the latent space and further exploit temporally corresponding patch representations across frames to model temporal variation.

\subsection{Overview}
\label{subsec:overview}

As illustrated in \Fgref{fig:pipeline}, the proposed framework follows a self-distillation architecture with three branches: an \textit{auxiliary} branch, a \textit{student} branch, and a \textit{teacher} branch.
Given a training video, we sample three ordered frames, which are designated as the past, current, and future frames.
For clarity, the past and future frames are referred to as \textit{auxiliary frames}.
The masked current frame is fed into the student branch, while the unmasked current frame is fed into the teacher branch to provide latent supervision through an exponential moving average of the student model.
Meanwhile, the auxiliary frames are processed by the auxiliary branch to provide two-sided temporal references for reconstructing the masked current-frame representations.

The framework is designed around two key objectives in self-supervised video representation learning: \textbf{1)} establishing reliable temporal correspondence across frames, and \textbf{2)} learning representations that preserve local semantic consistency while modeling temporal variation among corresponding regions.
For the first aspect, as described in \cref{subsec:temporal_cor}, the masked current frame and the auxiliary frames are processed by the student branch and the auxiliary branch, respectively, allowing the model to exploit two-sided temporal references.
In \cref{subsec:patch_matching}, we introduce a \textit{Patch Registration Module} to establish patch-level correspondence between the auxiliary frames and the current frame in the latent space, enabling dense inter-frame registration for subsequent temporal modeling.
For the second aspect, as outlined in \cref{subsec:temporal_recon}, the reconstructed representations are aligned with the teacher representations in a self-distillation manner, encouraging local semantic consistency in the latent space.
Finally, in \cref{subsec:temporal_var}, the registered representations are further projected into a separate space, where temporal variation modeling is imposed to constrain their latent trajectories over time.

\subsection{Uncertainty Reduction via Sandwich Sampling}
\label{subsec:temporal_cor}

As discussed in Sec.~\ref{sec:introduction}, recent video reconstruction methods commonly rely on random temporal sampling, where the past frame serves as a single-sided reference for recovering the current frame.
However, this setting may provide unreliable temporal guidance for reconstruction, since one past observation can lead to multiple plausible current states.

Although predicting the future from the past is difficult, inferring an intermediate state becomes easier when both the starting and ending states are available. Based on this idea, we propose a \textit{sandwich sampling strategy}. 
To reconstruct the masked current frame, we sample a past and a future frame to provide two-sided temporal references. 
Specifically, we first select the current frame $\vc = \bm{V}(t_c)$ at time step $t_c$ and obtain its masked version $\bar {\bm {v}}_c$. 
Then, we uniformly sample the past and future timestamps with offsets in $[\alpha T_0,\beta T_0]$, where $T_0$ denotes the temporal length of the input video, measured in the same time units as $t_c$.

\begin{equation}
    \begin{aligned}
        t_p &\sim \mathcal{U}(t_c - \beta \cdot T_0, t_c - \alpha \cdot T_0), \\
        t_f &\sim \mathcal{U}(t_c + \alpha \cdot T_0, t_c + \beta \cdot T_0).
    \end{aligned}
\end{equation}

The past and future frames are then obtained as $\vp = \bm{V}(t_p)$ and $\vf = \bm{V}(t_f)$, respectively.
These frames are passed through the encoder $f$ to obtain their representations:
$\bar{\bm{z}}_c = f(\bar{\bm{v}}_c)$, $\zp = f(\vp)$, and $\zf = f(\vf)$.

\subsection{Patch Registration for Dense Correspondence}
\label{subsec:patch_matching}

\begin{figure}
  \centering
    \includegraphics[width=0.98\linewidth]{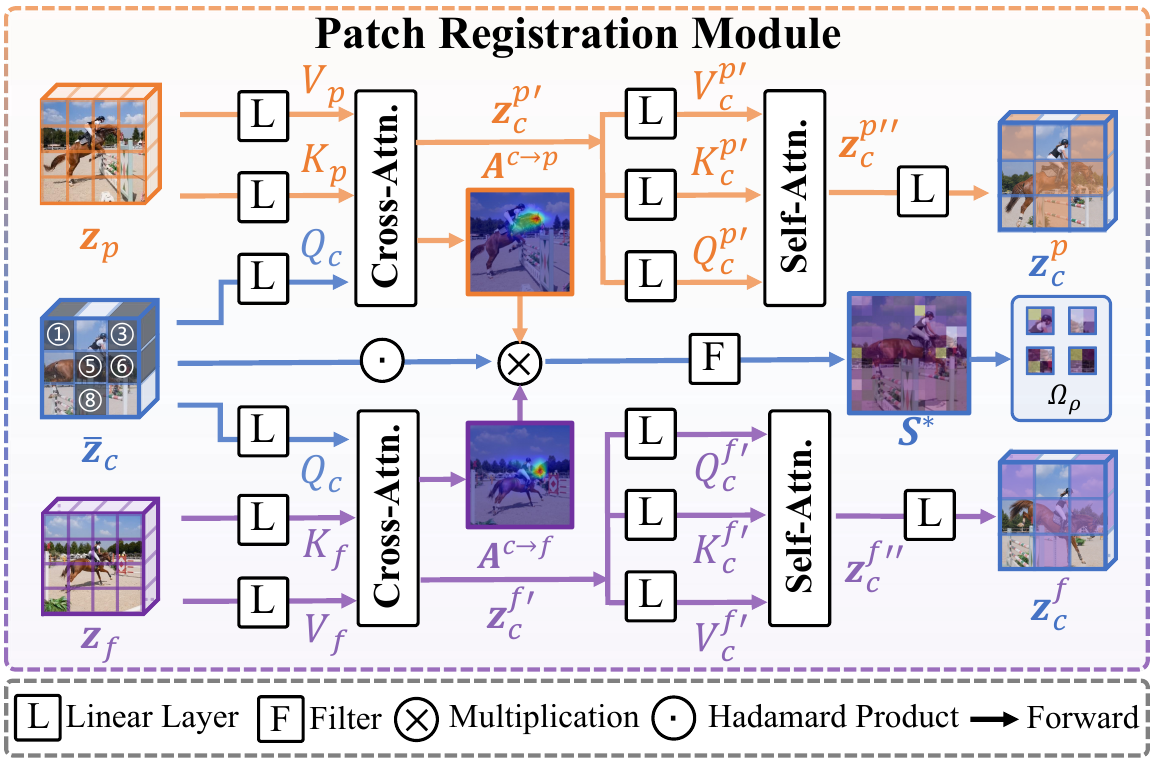}
    \caption{Structure of the Patch Registration Module. Given the past, current, and future representations, we register the auxiliary representations to the current-frame patch layout through attention modules, producing $\zcsp = g(\zp, \bar{\bm z}_c)$ and $\zcsf = g(\zf, \bar{\bm z}_c)$. This module establishes current-anchored temporal correspondence and selects high-confidence patch positions $\Omega_\rho$ for subsequent temporal modeling.}
    \label{fig:PRM}
    \vspace{-8pt}
\end{figure}

In this subsection, we focus on establishing dense correspondence between the extracted frames in the latent space.
While previous works have attempted to build correspondence in the pixel space~\cite{SiamMAE,RSP}, such a formulation may retain excessive low-level details due to the high redundancy of video information.
To address this issue, we conduct patch registration in a compact latent space.
Here, patch registration aims to map auxiliary representations onto the patch layout of the current frame, so that the past, current, and future representations are organized under the same current-frame positions for subsequent temporal modeling.
However, establishing precise correspondence between representations presents challenges due to the low resolution of latent tokens.
To overcome this, we introduce a \textit{Patch Registration Module} (PRM) $g$, which uses the current frame as the registration anchor and aligns the auxiliary representations to the current-frame patch layout.
As shown in \Fgref{fig:PRM}, the PRM consists of two main components: a cross-attention block for current-anchored inter-frame registration and a self-attention block for intra-frame contextual aggregation.

Specifically, the first step is to register auxiliary patches to the current-frame patch layout.
Using the masked current representations $\bar{\bm{z}}_c$ as the query and the auxiliary representations $\zp$ ($\zf$) as the key and value, cross-attention blocks are used to establish patch correspondence between the current frame and the two auxiliary frames:
\begin{equation}
    \begin{aligned}
    \Acp &= \text{softmax}_\tau \big(\frac{Q_c K_p^\top}{\sqrt{d}}\big), 
    & {\zcsp}' &= \Acp V_p, \\
    \Acf &= \text{softmax}_\tau \big(\frac{Q_c K_f^\top}{\sqrt{d}}\big), 
    & {\zcsf}' &= \Acf V_f ,
    \end{aligned}
    \label{eq:cross-attn}
\end{equation}
where $Q_c=\bar{\bm{z}}_c W_q$, $K_p=\zp W_k$, $V_p=\zp W_v$, $K_f=\zf W_k$, and $V_f=\zf W_v$, with $W_q, W_k, W_v \in \mathbb{R}^{d \times d}$ being weight matrices.
The softmax is applied along the last dimension:
\begin{align}
    \text{softmax}_\tau(\bm x)^i = \frac{\exp(\bm x^i/\tau)}{\sum_{j=1}^N\exp(\bm x^j/\tau)},
\end{align}
where $\tau > 0$ is a temperature hyperparameter.
In \Eqref{eq:cross-attn}, $\Acp$ and $\Acf$ can be regarded as normalized patch-transition matrices, where each row quantifies the confidence distribution from a current-frame patch to all candidate patches in the past and future frames, thereby providing dense inter-frame correspondences within the latent space.

Building upon this, we further apply self-attention blocks to model contextual relations among the registered patches on the current-frame layout:
\begin{equation}
    \begin{aligned}
    {\zcsp}'' &= \text{softmax}_\tau \big(\frac{{\Qcp}' {\Kcp}'^\top}{\sqrt{d}}\big) {\Vcp}' ,\\
    {\zcsf}'' &= \text{softmax}_\tau \big(\frac{{\Qcf}' {\Kcf}'^\top}{\sqrt{d}}\big) {\Vcf}' ,
    \end{aligned}
    \label{eq:self-attn}
\end{equation}
where ${\Qcp}'={\zcsp}' W'_q$, ${\Kcp}'={\zcsp}' W'_k$, ${\Vcp}'={\zcsp}' W'_v$, and ${\Qcf}'={\zcsf}' W'_q$, ${\Kcf}'={\zcsf}' W'_k$, ${\Vcf}'={\zcsf}' W'_v$, with $W'_q, W'_k, W'_v \in \mathbb{R}^{d \times d}$ being weight matrices.

Finally, a feed-forward network (FFN) is applied to calibrate the registered representations in the latent space:
\begin{equation}
    \begin{aligned}
    \zcsp &= \text{FFN}({\zcsp}''), \\
    \zcsf &= \text{FFN}({\zcsf}'').
    \end{aligned}
    \label{eq:linear}
\end{equation}
In practice, we adopt a fully connected layer as the FFN, and apply layer normalization and skip connections to each block for training stability.

The generated $\zcsp$ and $\zcsf$ denote the current-anchored past and future representations registered to the current-frame patch layout.
Since not all current-frame patch positions can be reliably matched across frames, we further select high-confidence positions from the two-sided transition matrices to reduce ambiguous matches.
For the $i$-th patch position on the current-frame layout, we compute
\begin{equation}
    \begin{aligned}
    \Spc_{i} &= \max_j \big((\Acp)^\top\big)_{j,i}, \\
    \Scf_{i} &= \max_k (\Acf)_{i,k},
    \end{aligned}
    \label{eq:patch_confidence}
\end{equation}

where $\Spc_i$ and $\Scf_i$ measure the strongest matching responses of the $i$-th current patch position with the past and future frames, respectively.
Then, we define the joint confidence as
\begin{equation}
    \begin{aligned}
    \bm{S}^{\ast}_i &= \Spc_i \Scf_i, \\
    \Omega_\rho &= \{i \mid \bm{S}^{\ast}_i \geq \eta_\rho,\ 1 \leq i \leq N\},
    \end{aligned}
    \label{eq:high_confidence_set}
\end{equation}
where $\rho$ denotes the retained patch ratio and $\eta_\rho$ is the corresponding confidence threshold, chosen such that $\Omega_\rho$ contains the top $\rho$ proportion of positions ranked by $\{\bm{S}^{\ast}_i\}_{i=1}^{N}$.
The selected set $\Omega_\rho$ thus defines reliable current-frame patch positions, on which $\bar{\bm{z}}_c$, $\zcsp$, and $\zcsf$ form temporally corresponding states for subsequent temporal modeling.

\subsection{Local Consistency via Self-distillation Alignment}
\label{subsec:temporal_recon}

In the previous subsections, we have obtained auxiliary representations registered to the current-frame patch layout. 
In this subsection, we describe how these registered representations are aligned with the teacher representations to learn local semantic consistency.

\noindent\textbf{Patch-level self-distillation alignment.} 
We formulate the patch-level alignment objective based on a self-distillation framework~\cite{iBOT}. Specifically, we introduce a teacher model (denoted as $f_t$) to guide the student branch. 
Inspired by the widely used mean-teacher technique~\cite{MoCov1}, we update the teacher model using the exponential moving average (EMA) of the student weights:
\begin{align}
\label{eq:ema}
    \bm \theta_t \xleftarrow{} m \cdot \bm \theta_t + (1-m) \cdot \bm \theta ,
\end{align}
where $\bm \theta$ and $\bm \theta_t$ are the parameters of the student model $f$ and the teacher model $f_t$, respectively.
As a temporal ensemble of the student model, the mean teacher provides a stable supervision signal for self-supervised learning~\cite{BYOL,DINO,iBOT,MoCov3}.

Following prior self-distillation methods, we introduce projection heads $h$ and $h_t$ on top of the student and teacher models.
To stabilize training and prevent model collapse~\cite{dai2026exploring}, we further apply centering and sharpening techniques~\cite{DINO,iBOT} to the outputs of the projection heads:

\begin{equation}
    \begin{aligned}
        \bm q_c^p &= \text{softmax}_{\tau_s}\left(h(\zcsp)\right), \\
        \bm q_c^f &= \text{softmax}_{\tau_s}\left(h(\zcsf)\right), \\
        \hat {\bm q}_c &= \text{softmax}_{\tau_t}\left(h_t(f_t(\bm v_c)) - \bm \mu_t\right), 
        \label{eq:proj}
    \end{aligned}
\end{equation}
where $\tau_s > 0$ and $\tau_t > 0$ are the student and teacher temperatures, and $\bm \mu_t$ is the running center estimated from the teacher outputs.
The patch-level self-distillation losses are then computed on the masked patch positions:

\begin{equation}
    \begin{aligned}
        \lpt &= -\frac{1}{|\mathcal M(\bm v_c)|}\sum_{i \in \mathcal{M}(\bm v_c)} (\hat {\bm q}_c)^i \log(\bm q_c^p)^i  , \\
        \lft &= -\frac{1}{|\mathcal M(\bm v_c)|}\sum_{i \in \mathcal{M}(\bm v_c)} (\hat {\bm q}_c)^i \log (\bm q_c^f)^i .
    \label{eq:loss1_main}
    \end{aligned}
\end{equation}

\noindent\textbf{Two-sided local consistency.} 
In addition to aligning each registered representation with the teacher target, we further encourage the past-side and future-side predictions to maintain consistent local semantics:
\begin{equation}
    \lpf = \frac{1}{|\mathcal M(\bm v_c)|}\sum_{i \in \mathcal{M}(\bm v_c)} \| (\bm q_c^p)^i - (\bm q_c^f)^i \|_2^2 .
\label{eq:loss2_main}
\end{equation}
This term reduces inconsistent semantic predictions from the two temporal references, making the registered patch states more comparable in the latent space.

\noindent\textbf{Frame-level alignment.}
Following prior self-distillation methods~\cite{DINO,iBOT,DINOv2}, we further apply a frame-level alignment objective on the \texttt{[CLS]} embeddings to preserve global semantic consistency.
Let $\bm q$ denote the sharpened \texttt{[CLS]} prediction of the student model, and let $\tilde{\bm q}$ denote the centered and sharpened \texttt{[CLS]} prediction of the teacher model on $\vc$.
The self-distillation loss is formulated as
\begin{equation}
    \mathcal{L}_{\mathrm{DINO}} = -\frac{1}{2} \left( \bm q \log \tilde{\bm q} +    \tilde{\bm q} \log \bm q \right).
    \label{eq:dino_loss}
\end{equation}

\subsection{Variation-aware Temporal Modeling}
\label{subsec:temporal_var}

The previous subsections establish dense temporal correspondence among the past, current, and future frames, and enforce local consistency on the registered representations. 
Since real-world targets usually evolve continuously within short temporal intervals, their visual observations also change in a locally smooth manner. 
With an effective visual encoder, such temporal changes should be reflected in the representation space, where corresponding regions are expected to form smooth latent trajectories rather than isolated states.
Therefore, modeling temporal variation essentially requires learning how the representations evolve in the latent space.
This leads to two key questions: \textbf{1)} how to formulate a unified latent trajectory constraint for corresponding representations, and \textbf{2)} how to incorporate this constraint within the proposed self-supervised learning framework.

\subsubsection{Representation Trajectory Modeling}
\label{subsubsec:trajectories_estimation}
Let $\zcsp$, $\zcsf$, and $\bar{\bm z}_c$ be the (registered) representations of the past frame, future frame, and the masked current frame.
We first apply a linear mapping to these representations to separate the semantic alignment and variation modeling:
\begin{equation}
    \bm p_c^{p} = \phi({\zcsp}), \quad
    \bar{\bm p}_c = \phi(\bar{\bm z}_c), \quad
    \bm p_c^{f} = \phi({\zcsf}),
    \label{eq:temporal_projection}
\end{equation}
where $\phi:\mathbb{R}^{d}\rightarrow\mathbb{R}^{d'}$ is implemented as a linear projection followed by layer normalization.

Since we only observe the representations of the same registered patch position across three ordered frames, fitting a non-parametric trajectory from such limited observations is challenging. To exploit the continuity and smoothness of the trajectory, we approximate it locally in time using a Taylor expansion.
Specifically, given a reference state $\bm P(t_0)$, a nearby state $\bm P(t)$ can be approximated as
\begin{equation}
    \bm P(t) = \bm P(t_0) + (t-t_0)\frac{\partial \bm P}{\partial t}(t_0) + \frac{(t-t_0)^2}{2} \frac{\partial^2 \bm P}{\partial t^2}(t_0) + \bm R_3(t),
    \label{eq:taylor_general}
\end{equation}
where $\bm R_3(t)$ denotes the third-order remainder terms.
Under local temporal sampling, we neglect the higher-order term by setting $\bm R_3(t) \approx 0$, and approximate the first- and second-order temporal variations at $t_c$ by finite differences. Specifically, let $t_p=t_c-\Delta_p$ and $t_f = t_c+\Delta_f$ be the past and future timestamps, where $\Delta_p, \Delta_f>0$ are the temporal intervals, we estimate $\frac{\partial \bm p}{\partial t}(t_c)$ and $\frac{\partial^2 \bm p}{\partial t^2}(t_c) $ as follows:
\begin{equation}
    \begin{aligned}
        \widehat{\frac{\partial \bm p}{\partial t}}(t_c)
         &= \frac{\Delta_p}{\Delta_p+\Delta_f}\frac{\bm p_c^f-\bar{\bm p}_c}{\Delta_f}
        +
        \frac{\Delta_f}{\Delta_p+\Delta_f}\frac{\bar{\bm p}_c-\bm p_c^p}{\Delta_p} \\
        & =
        \frac{\Delta_p^2 \bm p_c^f-\Delta_f^2 \bm p_c^p+(\Delta_f^2-\Delta_p^2)\bar{\bm p}_c}{\Delta_p\Delta_f(\Delta_p+\Delta_f)} ,
    \end{aligned}
    \label{eq:first_order_difference}
\end{equation}
\begin{equation}
    \begin{aligned}
    \widehat{\frac{\partial^2 \bm p}{\partial t^2}}(t_c) 
    & = \frac{2}{\Delta_p+\Delta_f}\left(
    \frac{\bm p_c^f-\bar{\bm p}_c}{\Delta_f} - \frac{\bar{\bm p}_c-\bm p_c^p}{\Delta_p} \right) \\
    &= \frac{2\Delta_f \bm p_c^p-2(\Delta_p+\Delta_f)\bar{\bm p}_c+2\Delta_p \bm p_c^f}{\Delta_p\Delta_f(\Delta_p+\Delta_f)} .
    \end{aligned}
    \label{eq:second_order_difference}
\end{equation}
On top of this, we further approximate $\bm p_c^{p} = \phi({\zcsp})$ and $\bm p_c^{f}$
with the second-order Taylor expansion around $t_c$:
\begin{equation}
    \bm p_c^{p} = \bar{\bm p}_c - \Delta_p
    \widehat{\frac{\partial \bm p}{\partial t}}(t_c) + \frac{(\Delta_p)^2}{2}
    \widehat{\frac{\partial^2 \bm p}{\partial t^2}}(t_c),
    \label{eq:taylor_past}
\end{equation}
\begin{equation}
    \bm p_c^{f} = \bar{\bm p}_c + \Delta_f
    \widehat{\frac{\partial \bm p}{\partial t}}(t_c) + \frac{(\Delta_f)^2}{2}
    \widehat{\frac{\partial^2 \bm p}{\partial t^2}}(t_c).
    \label{eq:taylor_future}
\end{equation}
The two Taylor expansions motivate a second-order local relation among the registered triplet:
\begin{equation}
    \widehat{\frac{\partial^2 \bm p}{\partial t^2}}(t_c)
    +
    \gamma_1
    \widehat{\frac{\partial \bm p}{\partial t}}(t_c)
    +
    \gamma_0
    \bm p^{\delta}
    =
    \bm 0,
    \label{eq:taylor_ode_relation}
\end{equation}
where $\bm p^{\delta}=\bar{\bm p}_c-(\bm p_c^{p}+\bm p_c^{f})/2$ denotes the deviation of the current state from the two-sided average.
The coefficients $\gamma_1$ and $\gamma_0$ characterize the local trajectory relation and can be derived from the temporal intervals $\Delta_p$ and $\Delta_f$ under the Taylor approximation.
By solving for these coefficients, we formulate the trajectory constraint.



However, due to approximation errors, a single observation provides limited evidence for learning a common temporal evolution pattern.
Since the local trajectory relation is governed by two parameters $(\gamma_0,\gamma_1)$, observations with similar local variation patterns can share these parameters and follow a common relation, which can be interpreted as a latent temporal evolution prototype.
Under the associated local continuous model, shared coefficients determine the same characteristic roots and solution family, with different trajectories arising from different initial conditions.
Accordingly, given $B$ observations with similar local variation patterns,
$\{(\widehat{\frac{\partial^2 \bm p^{(i)}}{\partial t^2}}, \widehat{\frac{\partial \bm p^{(i)}}{\partial t}}, \bm p^{\delta,(i)})\}_{i=1}^B$,
we assume that they share the same parameters $(\bar{\gamma}_0,\bar{\gamma}_1)$, which are estimated by solving:
\begin{equation}
\begin{aligned}
\text{(OP0)} \quad
&\min_{\bar{\gamma}_0,\bar{\gamma}_1}
\quad
\mathcal{R}(t_c)
=
\frac{1}{B}
\sum_{i=1}^{B}
\left\|
\bm \epsilon^{(i)}
\right\|_F^2
\\
&~~~~~=
\frac{1}{B}
\sum_{i=1}^{B}
\left\|
    \widehat{\frac{\partial^2 \bm p^{(i)}}{\partial t^2}}
    +
    \bar{\gamma}_1
    \widehat{\frac{\partial \bm p^{(i)}}{\partial t}}
    +
    \bar{\gamma}_0
    \bm p^{\delta,(i)}
\right\|_F^2 .
\end{aligned}
\label{eq:op0_batch_trajectory}
\end{equation}

Jointly estimating the shared coefficients from multiple observations can reduce sensitivity to individual approximation errors and improve prototype estimation.
Solving this problem provides a shared local trajectory constraint for regularizing the temporal evolution of representations.

\subsubsection{Learning with Adaptive Trajectory Constraints}
\label{subsubsec:adaptive_constraints}

The formulation in \cref{subsubsec:trajectories_estimation} turns temporal variation modeling into fitting a unified second-order trajectory over registered triplets. In this part, we provide an efficient implementation to solve $\text{(OP0)}$ and constrain the representations.

Directly solving \Eqref{eq:op0_batch_trajectory} is impractical during training, since it is difficult to sample observations with similar local variation patterns from the entire dataset.
We therefore replace the coefficients with learnable channel-wise parameters and optimize them jointly with the representation learning objective.
To further accommodate diverse temporal variation patterns, we maintain $K$ groups of learnable parameter vectors
$\bm\Gamma = \{(\bm{\gamma}_0^{k},\bm{\gamma}_1^{k})\}_{k=1}^{K}$, where
$\bm{\gamma}_0^{k}, \bm{\gamma}_1^{k} \in \mathbb{R}^{d'}$ are defined in the projected temporal space.

For the $k$-th parameter group, the learnable parameters and finite-difference estimates yield the following residual:
\begin{equation}
\bm r^{k,(i)}(t_c) =
\widehat{\frac{\partial^2 \bm p^{(i)}}{\partial t^2}}(t_c)
+ \bm{\gamma}_1^{k} \odot
\widehat{\frac{\partial \bm p^{(i)}}{\partial t}}(t_c)
+ \bm{\gamma}_0^{k} \odot \bm p^{\delta,(i)}.
\label{eq:ode_residual_matrix}
\end{equation}
Here, $\bm r^{k,(i)}(t_c)\in\mathbb{R}^{N\times d'}$ measures the mismatch between the finite-difference local evolution and the $k$-th learnable second-order trajectory constraint.
The operator $\odot$ denotes element-wise multiplication, and $\bm{\gamma}_0^{k},\bm{\gamma}_1^{k}\in\mathbb{R}^{d'}$ are broadcast along the token dimension.

Based on the high-confidence patch set $\Omega_\rho^{(i)}$ defined in \Eqref{eq:high_confidence_set}, we construct a binary patch map $\bm M_\rho^{(i)}\in\{0,1\}^{N\times 1}$, where the entries corresponding to $\Omega_\rho^{(i)}$ are set to $1$.
For the $i$-th sample and the $k$-th parameter group, the mean squared residual over high-confidence patches is computed as
\begin{equation}
    \begin{aligned}
    \mathcal{R}^{k,(i)}(t_c)
    & =
    \frac{1}{d'|\Omega_\rho^{(i)}|}
    \left\|
    \bm M_\rho^{(i)} \odot \bm r^{k,(i)}(t_c)
    \right\|_F^2 .
    \end{aligned}
    \label{eq:op1_residual}
\end{equation}
Here, the Frobenius norm aggregates the channel-wise deviations in the projected temporal space, and the mask restricts the residual computation to reliable registered patches.

With multiple trajectory parameter groups, the adaptive trajectory fitting objective can be written as
\begin{equation}
    \begin{aligned}
    \text{(OP1)} \quad
    \min_{\bm\Gamma}
    \quad
    \frac{1}{B}
    \sum_{i=1}^{B}
    \min_{k\in\{1,\ldots,K\}}
    \mathcal{R}^{k,(i)}(t_c).
    \end{aligned}
    \label{eq:op1_adaptive_trajectory}
\end{equation}
The inner minimization performs a hard selection of the parameter group that best fits the selected patch trajectories of each sample.
To make this selection differentiable and stable during training, we use a soft assignment over group-wise residuals for each sample:
\begin{equation}
    \omega_k^{(i)}
    =
    \frac{\exp(-\tau_o \mathcal{R}^{k,(i)}(t_c))}
    {\sum_{q=1}^{K}\exp(-\tau_o \mathcal{R}^{q,(i)}(t_c))},
    \label{eq:ode_soft_assignment}
\end{equation}
where $\tau_o$ is a sharpness hyperparameter for the assignment.
The weights are computed separately for each sampled frame triplet and shared by its selected patches, while the parameter groups $\bm\Gamma$ are shared across samples.

The temporal variation loss is then obtained by averaging the sample-wise weighted residuals over the batch:
\begin{equation}
    \begin{aligned}
    \mathcal L_{\mathrm{var}}(t_c)
     =
    \frac{1}{B}
    \sum_{i=1}^{B}
    \sum_{k=1}^{K}
    \omega_k^{(i)} \mathcal{R}^{k,(i)}(t_c).
    \end{aligned}
    \label{eq:variation_loss}
\end{equation}

This soft assignment gives larger weights to the parameter groups that better fit the selected patch trajectories of each sample, enabling differentiable selection of suitable variation patterns.
This formulation does not require samples within a mini-batch to share similar temporal variation patterns.
Together with the local consistency objective, $\mathcal L_{\mathrm{var}}$ encourages temporally corresponding representations to preserve local semantics while following coherent changes across frames.

The total training objective is formulated as
\begin{equation}
    \mathcal{L} = \underbrace{\lambda_1(\lpt+\lft)+\lambda_2\lpf+\lambda_3\mathcal{L}_{\mathrm{DINO}}}_{\text{local consistency}} + \underbrace{\lambda_4\mathcal{L}_{\mathrm{var}}}_{\text{temporal variation}},
\label{eq:total_loss}
\end{equation}
where $\lambda_j$ $(j=1,\ldots,4)$ are hyperparameters that control the contributions of different loss terms.

The overall optimization algorithm of the methods presented in this section is summarized in \Algref{alg:algo}. The inner loop can be implemented with matrix operations in practice.

\begin{algorithm}[tb]
\caption{\methodname{} Optimization Algorithm}
\label{alg:algo}
\LinesNumbered
\KwIn{Unlabeled dataset $\mathcal{D}$, maximum iterations $L$.}
\KwOut{Pre-trained student model parameters.}
Initialize model parameters.

\For{$l=1$ {\bfseries to} $L$}{
    Sample a batch of videos $\{\bm{V}_i\}_{i=1}^B$.

    \For{$i=1$ {\bfseries to} $B$}{
        \textcolor{cvprblue}{$\triangleright$ \cref{subsec:temporal_cor}: Sandwich Sampling}
        
        Select three frames $\vc$, $\vp$, and $\vf$ from $\bm{V}_i$.

        Get the masked current frame $\bar{\bm v}_c$.
    
        Extract dense representations $\bar{\bm z}_c$, $\zp$, $\zf$.

        \textcolor{cvprblue}{$\triangleright$ \cref{subsec:patch_matching}: Patch Registration}

        Register auxiliary representations via PRM.

        Build the high-confidence set $\Omega_\rho$ by \Eqref{eq:high_confidence_set}.

        \textcolor{cvprblue}{$\triangleright$  \cref{subsec:temporal_recon}: Local Consistency Alignment}
        
        Generate $\bm q_c^p$, $\bm q_c^f$, and $\hat{\bm q}_c$ by \Eqref{eq:proj}.

        Distill from the teacher by $\lpt$, $\lft$ in \Eqref{eq:loss1_main}.

        Align $\bm q_c^p$ with $\bm q_c^f$ by $\lpf$ in \Eqref{eq:loss2_main}.

        \textcolor{cvprblue}{$\triangleright$ \cref{subsec:temporal_var}: Temporal Variation Modeling}

        Project $\zcsp$, $\bar{\bm z}_c$, and $\zcsf$ via $\phi(\cdot)$ in \Eqref{eq:temporal_projection}.

        Estimate $\widehat{\frac{\partial \bm p}{\partial t}}(t_c)$, $\widehat{\frac{\partial^2 \bm p}{\partial t^2}}(t_c)$  by \Eqref{eq:first_order_difference}, \Eqref{eq:second_order_difference}.

        Build $K$-group residuals $\{\bm r^{k}\}_{k=1}^{K}$ by \Eqref{eq:ode_residual_matrix}.

        Compute residuals on the set $\Omega_\rho$ by \Eqref{eq:op1_residual}.

        Compute $\mathcal{L}_{\mathrm{var}}$ by soft assignment in \Eqref{eq:variation_loss}.
    }

    Update the student model with $\mathcal{L}$ (\Eqref{eq:total_loss}).

    Update the teacher model with EMA (\Eqref{eq:ema}).
}
\Return

\end{algorithm}

%% file: sec/4_experiments.tex
\section{Experiments}
\label{sec:experiments}

In this section, we evaluate the proposed framework from four complementary aspects: downstream task performance, computational efficiency, component effectiveness, and representation analysis.
We first introduce the implementation details and experimental settings in \cref{subsec:implementation_details}, followed by the main results across multiple types of downstream tasks in \cref{subsec:main_results}.
We then examine the pre-training efficiency in \cref{subsec:efficiency_analysis} and conduct ablation studies in \cref{subsec:ablation_study} to assess the contribution of key design choices.
Finally, \cref{subsec:representation_analysis} examines the learned representations through quantitative representation metrics and qualitative visualizations, further analyzing whether \methodname{} learns a more structured latent space for variation-aware temporal correspondence learning.

\subsection{Implementation Details}
\label{subsec:implementation_details}


\textbf{Network architecture.}
We adopt ViT-Small (ViT-S) and ViT-Base (ViT-B)~\cite{Vision_Transformer} with a patch size of 16 as the backbone models.
The feature dimension is 384 for ViT-S/16 and 768 for ViT-B/16.
By default, the number of trajectory parameter groups is set to $K=3$, and the projection dimension is set to $d'=256$.
The soft-assignment sharpness parameter is set to a moderate value of $\tau_o=2.0$.

\textbf{Pre-training settings.} 
For pre-training, we extract frames from each video at $2$ FPS. 
For each training sample, the current frame and its past/future reference frames are sampled according to the past/future offset range, with $\alpha=0.10$ and $\beta=0.20$ by default.
For the current frame, we generate $2$ global views of size $224\times224$ and $8$ local views of size $96\times96$. 
Following~\cite{DINOv2}, we select $50\%$ of the global crops in the student branch and randomly mask $10\%\sim50\%$ of their patches. 
The ViT-S/16 model is trained on Kinetics-400~\cite{Kinetics} with a batch size of $256$ for $400$ epochs. 
The ViT-B/16 model is trained with a batch size of $128$ for $200$ epochs for pixel-level evaluations and $400$ epochs for other evaluations.


\textbf{Optimization strategy.}
The student branch is optimized by AdamW.
The base learning rate $\textit{blr}$ is set to $2\times10^{-3}$ for ViT-S/16 and $1\times10^{-3}$ for ViT-B/16, and decays to $1\times10^{-6}$ with a cosine schedule.
The actual learning rate is computed as $\textit{lr}=\textit{blr}\sqrt{\textit{bs}/256}$, where $\textit{bs}$ denotes the batch size.
For the PRM, the learning rate is set to $0.1\times\textit{lr}$.
The loss weights are set to $\lambda_1=0.8$, $\lambda_2=20$, $\lambda_3=1.0$, and $\lambda_4=0.5$.
Here, $\lambda_1$, $\lambda_2$, and $\lambda_3$ follow the default settings of~\cite{tcore}, while only $\lambda_4$ is tuned in our experiments.
The variation-aware loss $\mathcal{L}_{\mathrm{var}}$ is introduced at $\xi=0.6$ ($\xi_{\mathrm{s}}$), where $\xi\in[0,1]$ denotes normalized pre-training progress.
By default, its weight follows a piecewise cosine schedule, increasing from $0$ to $\lambda_4$ at $\xi=0.8$ ($\xi_{\mathrm{p}}$) and decreasing to $0.2\times\lambda_4$ at $\xi=1$.

\textbf{Competitors.}
We compare the proposed method with representative self-supervised video representation learning methods within two categories: 
\textbf{1) }\textit{Contrastive learning methods}, including SimCLR~\cite{SimCLRv1}, MoCo v3~\cite{MoCov3}, DINO~\cite{DINO}; 
\textbf{2) }\textit{Masked modeling methods}, including MAE~\cite{MAE}, MAE-ST~\cite{MAE_ST}, VideoMAE~\cite{VideoMAE}, DropMAE~\cite{DropMAE}, SiamMAE~\cite{SiamMAE}, CropMAE~\cite{CropMAE}, RSP~\cite{RSP}, iBOT~\cite{iBOT}, DINOv2~\cite{DINOv2}, Ada-iBOT~\cite{co_settle}, Ada-DINOv2~\cite{co_settle}, and V-JEPA 2.1~\cite{vjepa2_1}. 
Among them, iBOT and DINOv2 are pre-trained on Kinetics-400 from scratch, while Ada-iBOT and Ada-DINOv2 are initialized from ImageNet-pretrained models and further adapted on Kinetics-400.

\subsection{Main Results on Video Downstream Tasks}
\label{subsec:main_results}

We evaluate the representation capabilities of the proposed model on 8 downstream tasks across 16 benchmarks, covering pixel-level, frame-level, video-level, and vision-based robotic learning scenarios.
These tasks assess different aspects of video representations, ranging from dense correspondence and temporal decoding to global semantic recognition and vision-based control. 
An overview of the downstream tasks and evaluation protocols is provided in \Tbref{tab:overview_tasks}.

\begin{table}[t]
  \centering
  \caption{Overview of downstream tasks and evaluation protocols, covering 8 downstream tasks across 4 task categories.}
  \vspace{-3pt}
  \setlength{\tabcolsep}{3pt}
  \resizebox{\columnwidth}{!}{%
    \begin{tabular}{c|ccc}
    \toprule
    \makecell[c]{\RowStrut\textbf{Task}\\\textbf{Category}}
    & \makecell[c]{\RowStrut\textbf{Downstream}\\\textbf{Task}}
    & \makecell[c]{\RowStrut\textbf{Evaluation}\\\textbf{Protocol}}
    & \makecell[c]{\RowStrut\textbf{Benchmarks}} \\
    \midrule

    \multirow[c]{2}{*}[-0.4ex]{\makecell[c]{\BoldRowStrut \textbf{Pixel-level}\\(\cref{subsec:evaluation_pixel})\\{$\mathbb{R}^{T \times N \times D}$}}}
    & \makecell[c]{\RowStrut\textbf{Object / Part}\\\textbf{Segmentation}}
    & \makecell[c]{\RowStrut Semi-supervised\\ propagation}
    & \makecell[c]{\RowStrut DAVIS-2017~\cite{DAVIS17}\\VIP~\cite{VIP}} \\
    \cmidrule(lr){2-4}

    &
    \makecell[c]{\RowStrut\textbf{Human Pose}\\\textbf{Estimation}}
    & \makecell[c]{\RowStrut Semi-supervised\\ propagation}
    & \makecell[c]{\RowStrut JHMDB~\cite{JHMDB}} \\
    \midrule

    \multirow[c]{2}{*}[-0.4ex]{\makecell[c]{\BoldRowStrut \textbf{Frame-level}\\(\cref{subsec:evaluation_frame})\\{$\mathbb{R}^{T \times D}$}}}
    & \makecell[c]{\RowStrut\textbf{Action}\\\textbf{Segmentation}}
    & \makecell[c]{\RowStrut Frozen features with\\ temporal decoding}
    & \makecell[c]{\RowStrut Breakfast~\cite{Breakfast}\\EgoProceL~\cite{EgoProceL}} \\
    \cmidrule(lr){2-4}

    &
    \makecell[c]{\RowStrut\textbf{Temporal}\\\textbf{Localization}}
    & \makecell[c]{\RowStrut Frozen features with\\ temporal decoding}
    & \makecell[c]{\RowStrut THUMOS14~\cite{THUMOS14}} \\
    \midrule

    \multirow[c]{3}{*}[-2.4ex]{\makecell[c]{\BoldRowStrut \textbf{Video-level}\\(\cref{subsec:evaluation_video})\\{$\mathbb{R}^{D}$}}}
    & \makecell[c]{\RowStrut\textbf{Action}\\\textbf{Classification}}
    & \makecell[c]{\RowStrut Full fine-tuning}
    & \makecell[c]{\RowStrut Kinetics-400~\cite{Kinetics}\\HMDB51~\cite{HMDB51}} \\
    \cmidrule(lr){2-4}

    &
    \makecell[c]{\RowStrut\textbf{Video}\\\textbf{Retrieval}}
    & \makecell[c]{\RowStrut Zero-shot retrieval}
    & \makecell[c]{\RowStrut UCF101~\cite{UCF101}\\HMDB51~\cite{HMDB51}} \\
    \cmidrule(lr){2-4}

    &
    \makecell[c]{\RowStrut\textbf{Chiral}\\\textbf{Recognition}}
    & \makecell[c]{\RowStrut Linear probing}
    & \makecell[c]{\RowStrut Chiral SSv2~\cite{Chirality}\\Chiral Charades~\cite{Chirality}} \\
    \midrule

    \makecell[c]{\BoldRowStrut \textbf{Robotic learning}\\(\cref{subsec:evaluation_robotic})\\{$\mathbb{R}^{T \times D}$}}
    & \makecell[c]{\RowStrut\textbf{Vision-based}\\\textbf{Robotic Learning}}
    & \makecell[c]{\RowStrut Frozen features with\\ imitation learning}
    & \makecell[c]{\RowStrut CortexBench~\cite{CortexBench_1}\\(Adroit~\cite{Adroit}, DMC~\cite{DMC}, \\ TriFinger~\cite{TriFinger}, \\  MetaWorld~\cite{MetaWorld})} \\
    \bottomrule
    \end{tabular}%
  }
  \label{tab:overview_tasks}
  \vspace{-12pt}
\end{table}

\begin{table*}[t]
  \centering
   \caption{Evaluation results on pixel-level downstream tasks. Results marked with $^\dagger$ are directly taken from previous studies, while $^\ddagger$ denotes evaluations based on officially released pre-trained weights. $^\ast$ indicates that the model is distilled from ViT-G. K400 and VMix163M denote Kinetics-400~\cite{Kinetics} and VisionMix-163M~\cite{vjepa2_1}, respectively. The best and second-best results are highlighted in \textbf{bold} and \underline{underlined}, respectively.}

   \renewcommand{\arraystretch}{1.15}
   \large
    \resizebox{0.98\textwidth}{!}{%
      \begin{tabular}{cc|lccc|ccc|c|cc}
      \toprule
      \multicolumn{2}{c|}{\multirow{3}[2]{*}{Type}} & \multirow{3}[2]{*}{Method} & \multirow{3}[2]{*}{Backbone} & \multirow{3}[2]{*}{\multirow{2}{*}{\centering \shortstack{Training \\ Dataset}}} & \multirow{3}[2]{*}{Epoch} & \multicolumn{4}{c|}{\rule[-1.0ex]{0pt}{1.5ex}\textbf{Object/Part Segmentation}} & \multicolumn{2}{c}{\textbf{Human Pose Estimation}} \\
    \cline{7-12}          &       &    &    &       &       & \multicolumn{3}{c|}{\rule[-1.0ex]{0pt}{3.5ex}DAVIS-2017} & VIP   & \multicolumn{2}{c}{JHMDB} \\
    \cline{7-12}          &       &    &    &       &       & \rule[-1.0ex]{0pt}{3.5ex}$\mathcal{J}\&\mathcal{F}_{\mathrm{m}}$ & $\mathcal{J}_{\mathrm{m}}$ & $\mathcal{F}_{\mathrm{m}}$ & mIoU  & PCK@0.1 & PCK@0.2 \\
    \midrule
     \multicolumn{2}{c|}{\multirow{3}[2]{*}{\centering \shortstack{Contrastive \\ Learning}}} 
          & SimCLR$^\dagger$~\cite{SimCLRv1}  & ViT-S/16 & K400 & 400  & \cellcolor[rgb]{ .918,  .953,  .98}53.9  & \cellcolor[rgb]{ .91,  .949,  .98}51.7  & \cellcolor[rgb]{ .918,  .953,  .98}56.2  & \cellcolor[rgb]{ .918,  .953,  .98}31.9  & \cellcolor[rgb]{ .965,  .98,  .957}37.9  & \cellcolor[rgb]{ .965,  .98,  .957}66.1  \\
          & & MoCo v3$^\dagger$~\cite{MoCov3}   & ViT-S/16 & K400 &  400  & \cellcolor[rgb]{ .882,  .929,  .973}57.7  & \cellcolor[rgb]{ .882,  .933,  .973}54.6  & \cellcolor[rgb]{ .878,  .929,  .973}60.8  & \cellcolor[rgb]{ .906,  .945,  .98}32.4  & \cellcolor[rgb]{ .961,  .976,  .949}38.4  & \cellcolor[rgb]{ .945,  .969,  .929}67.6  \\
          & & DINO$^\dagger$~\cite{DINO}   & ViT-S/16 & K400 &  400  & \cellcolor[rgb]{ .851,  .91,  .965}59.5  & \cellcolor[rgb]{ .859,  .918,  .969}56.5  & \cellcolor[rgb]{ .839,  .906,  .965}62.5  & \cellcolor[rgb]{ .875,  .925,  .973}33.4  & \cellcolor[rgb]{ .918,  .957,  .894}41.1  & \cellcolor[rgb]{ .902,  .949,  .875}70.3  \\
    \cmidrule(lr){1-12}
    \multicolumn{1}{c|}{\multirow{8}[1]{*}{\centering \shortstack{Masked \\ Modeling}}} & \multirow{5}[1]{*}{\centering \shortstack{Pixel \\ Space}} 
          & MAE$^\dagger$~\cite{MAE}   & ViT-S/16 & K400 &  400 & \cellcolor[rgb]{ .918,  .953,  .98}53.5  & \cellcolor[rgb]{ .918,  .953,  .98}50.4  & \cellcolor[rgb]{ .914,  .953,  .98}56.7  & \cellcolor[rgb]{ .902,  .945,  .976}32.5  & \cellcolor[rgb]{ .886,  .937,  .855}43.0  & \cellcolor[rgb]{ .886,  .937,  .855}71.3  \\
         \multicolumn{1}{c|}{} & & VideoMAE$^\dagger$~\cite{VideoMAE}   & ViT-S/16 & K400 & 1600  & \cellcolor[rgb]{ .961,  .98,  .992}39.3  & \cellcolor[rgb]{ .961,  .98,  .992}39.7  & \cellcolor[rgb]{ .961,  .98,  .992}38.9  & \cellcolor[rgb]{ .961,  .98,  .992}22.3  & \cellcolor[rgb]{ .918,  .957,  .898}41.0  & \cellcolor[rgb]{ .941,  .969,  .922}67.9  \\
         \multicolumn{1}{c|}{} & & SiamMAE$^\dagger$~\cite{SiamMAE}   & ViT-S/16 & K400 & 400   & \cellcolor[rgb]{ .878,  .929,  .973}57.9  & \cellcolor[rgb]{ .871,  .925,  .973}56.0  & \cellcolor[rgb]{ .886,  .933,  .976}60.0  & \cellcolor[rgb]{ .878,  .929,  .973}33.2  & \cellcolor[rgb]{ .725,  .851,  .643}46.1  & \cellcolor[rgb]{ .761,  .871,  .686}74.0  \\
         \multicolumn{1}{c|}{} & & CropMAE$^\dagger$~\cite{CropMAE}   & ViT-S/16 & K400 & 400   & \cellcolor[rgb]{ .875,  .925,  .973}58.6  & \cellcolor[rgb]{ .875,  .925,  .973}55.8  & \cellcolor[rgb]{ .875,  .925,  .973}61.4  & \cellcolor[rgb]{ .863,  .918,  .969}33.7  & \cellcolor[rgb]{ .89,  .941,  .859}42.9  & \cellcolor[rgb]{ .89,  .941,  .863}71.1  \\
         \multicolumn{1}{c|}{} & & RSP$^\dagger$~\cite{RSP}    & ViT-S/16 & K400 & 400   & \cellcolor[rgb]{ .824,  .894,  .961}60.1  & \cellcolor[rgb]{ .824,  .898,  .961}57.4  & \cellcolor[rgb]{ .827,  .898,  .961}62.8  & \cellcolor[rgb]{ .855,  .918,  .969}33.8  & \cellcolor[rgb]{ .804,  .894,  .745}44.6  & \cellcolor[rgb]{ .788,  .886,  .725}73.4  \\
   \cmidrule(lr){2-12}
   \multicolumn{1}{c|}{} & \multirow{3}[1]{*}{\centering \shortstack{Latent \\ Space}} 
          & iBOT~\cite{iBOT}   & ViT-S/16 & K400 & 400  &  \cellcolor[rgb]{ .678,  .804,  .922}\underline{63.5} & \cellcolor[rgb]{ .663,  .796,  .918}61.7  & \cellcolor[rgb]{ .694,  .816,  .925}\underline{65.4}  & \cellcolor[rgb]{ .675,  .804,  .922}\underline{37.4}  & \cellcolor[rgb]{ .812,  .898,  .753}44.5  & \cellcolor[rgb]{ .749,  .863,  .675}74.2  \\
         \multicolumn{1}{c|}{} & & DINO v2~\cite{DINOv2}   & ViT-S/16 & K400 & 400   & \cellcolor[rgb]{ .698,  .82,  .925}63.0  & \cellcolor[rgb]{ .651,  .788,  .914}\underline{62.0}  & \cellcolor[rgb]{ .761,  .855,  .941}64.1  & \cellcolor[rgb]{ .694,  .816,  .925}37.0  & \cellcolor[rgb]{ .702,  .839,  .608}\underline{46.6}  & \cellcolor[rgb]{ .722,  .847,  .635}\underline{74.8}  \\ 
         \multicolumn{1}{c|}{} & & \methodnameb{} \textbf{(Ours)}     & ViT-S/16 & K400 & 400   & \cellcolor[rgb]{ .608,  .761,  .902}\textbf{65.1} & \cellcolor[rgb]{ .608,  .761,  .902}\textbf{63.1} & \cellcolor[rgb]{ .608,  .761,  .902}\textbf{67.1} & \cellcolor[rgb]{ .608,  .761,  .902}\textbf{38.7} & \cellcolor[rgb]{ .663,  .816,  .557}\textbf{47.3} & \cellcolor[rgb]{ .663,  .816,  .557}\textbf{76.0} \\

     \midrule
     \midrule
     \multicolumn{1}{c|}{\multirow{10}[1]{*}{\centering \shortstack{Masked \\ Modeling}}} & \multirow{6}[1]{*}{\centering \shortstack{Pixel \\ Space}} 
     &  VideoMAE$^\dagger$~\cite{VideoMAE}     & ViT-L/16 & K400 & 1600  & \cellcolor[rgb]{ .961,  .98,  .992}45.0  & \cellcolor[rgb]{ .961,  .98,  .992}43.6  & \cellcolor[rgb]{ .961,  .98,  .992}46.5  & \cellcolor[rgb]{ .961,  .98,  .992}25.6  & \cellcolor[rgb]{ .945,  .969,  .929}43.3  & \cellcolor[rgb]{ .949,  .973,  .933}70.5  \\
     \multicolumn{1}{c|}{} & &  MAE-ST$^\dagger$~\cite{MAE_ST}     & ViT-L/16 & K400 & 1600  & \cellcolor[rgb]{ .91,  .949,  .98}54.6  & \cellcolor[rgb]{ .894,  .941,  .976}55.5  & \cellcolor[rgb]{ .918,  .953,  .98}53.6  & \cellcolor[rgb]{ .898,  .941,  .976}33.2  & \cellcolor[rgb]{ .918,  .957,  .898}44.4  & \cellcolor[rgb]{ .922,  .957,  .898}72.5  \\
     \multicolumn{1}{c|}{} & & DropMAE$^\dagger$~\cite{DropMAE}  & ViT-B/16 & K400 & 1600  & \cellcolor[rgb]{ .918,  .953,  .98}53.4  & \cellcolor[rgb]{ .918,  .953,  .98}51.8  & \cellcolor[rgb]{ .914,  .949,  .98}55.0  & \cellcolor[rgb]{ .918,  .953,  .98}31.1  & \cellcolor[rgb]{ .965,  .98,  .957}42.3  & \cellcolor[rgb]{ .965,  .98,  .957}69.2  \\
     \multicolumn{1}{c|}{} & & SiamMAE~\cite{SiamMAE}  & ViT-B/16   & K400 & 400  &  \cellcolor[rgb]{ .867,  .922,  .969}60.9 &  \cellcolor[rgb]{ .867,  .922,  .969}59.4 &  \cellcolor[rgb]{ .875,  .925,  .973}62.4 & \cellcolor[rgb]{ .867,  .922,  .969}36.1  & \cellcolor[rgb]{ .745,  .859,  .663}{47.0} & \cellcolor[rgb]{ .871,  .929,  .831}74.9  \\
     \multicolumn{1}{c|}{} & & CropMAE~\cite{CropMAE}   & ViT-B/16  & K400 & 400  &  \cellcolor[rgb]{ .89,  .937,  .976}57.8 & \cellcolor[rgb]{ .886,  .933,  .973}56.9  & \cellcolor[rgb]{ .894,  .937,  .976}58.7  & \cellcolor[rgb]{ .902,  .941,  .976}33.0  & \cellcolor[rgb]{ .898,  .945,  .871}45.3 & \cellcolor[rgb]{ .91,  .949,  .882}73.3  \\
     \multicolumn{1}{c|}{} & & RSP$^\dagger$~\cite{RSP}   & ViT-B/16 & K400 & 400   & \cellcolor[rgb]{ .871,  .925,  .973}60.5  & \cellcolor[rgb]{ .878,  .929,  .973}57.8  & \cellcolor[rgb]{ .867,  .922,  .969}63.2  & \cellcolor[rgb]{ .89,  .937,  .976}34.0  & \cellcolor[rgb]{ .855,  .922,  .816}46.0  & \cellcolor[rgb]{ .89,  .941,  .859}74.6  \\

   \cmidrule(lr){2-12}
   \multicolumn{1}{c|}{} & \multirow{4}[1]{*}{\centering \shortstack{Latent \\ Space}} 
         & iBOT~\cite{iBOT}  & ViT-B/16 & K400 & 400 &  \cellcolor[rgb]{ .776,  .867,  .945}63.1 & \cellcolor[rgb]{ .78,  .867,  .949}61.4  & \cellcolor[rgb]{ .796,  .878,  .953}64.7  & \cellcolor[rgb]{ .761,  .855,  .941}38.0 & \cellcolor[rgb]{ .788,  .886,  .725}46.6  & \cellcolor[rgb]{ .804,  .894,  .745}75.4  \\
         \multicolumn{1}{c|}{} & & DINO v2~\cite{DINOv2}   & ViT-B/16 & K400 & 200   & \cellcolor[rgb]{ .698,  .82,  .925}64.9 & \cellcolor[rgb]{ .694,  .816,  .925}63.3  & \cellcolor[rgb]{ .71,  .824,  .929}66.6  & \cellcolor[rgb]{ .749,  .847,  .941}38.2  & \cellcolor[rgb]{ .765,  .875,  .694}46.8  & \cellcolor[rgb]{ .753,  .867,  .678}75.8  \\  
         \multicolumn{1}{c|}{} & & V-JEPA 2.1~\cite{vjepa2_1}      & ViT-B/16$^\ast$ & VMix163M & Distill.   & \cellcolor[rgb]{ .694,  .816,  .925}\underline{65.0} & \cellcolor[rgb]{ .722,  .831,  .933}62.7  & \cellcolor[rgb]{ .671,  .8,  .922}\underline{67.4}  & \cellcolor[rgb]{ .639,  .78,  .91}\underline{40.1} & \cellcolor[rgb]{ .745,  .859,  .663}\underline{47.0}  & \cellcolor[rgb]{ .663,  .816,  .557}\textbf{76.5}  \\
         \multicolumn{1}{c|}{} & & \methodnameb{} \textbf{(Ours)}   & ViT-B/16 & K400 & 200   & \cellcolor[rgb]{ .608,  .761,  .902}\textbf{67.0} & \cellcolor[rgb]{ .608,  .761,  .902}\textbf{65.2} & \cellcolor[rgb]{ .608,  .761,  .902}\textbf{68.7} & \cellcolor[rgb]{ .608,  .761,  .902}\textbf{40.6} & \cellcolor[rgb]{ .663,  .816,  .557}\textbf{47.7} & \cellcolor[rgb]{ .69,  .831,  .592}\underline{76.3}  \\
    \bottomrule
    \end{tabular}%
    }
  \label{tab:results_pixel_level}%
  \vspace{-8pt}
\end{table*}%

\begin{figure*}
  \centering
    \includegraphics[width=0.94\linewidth]{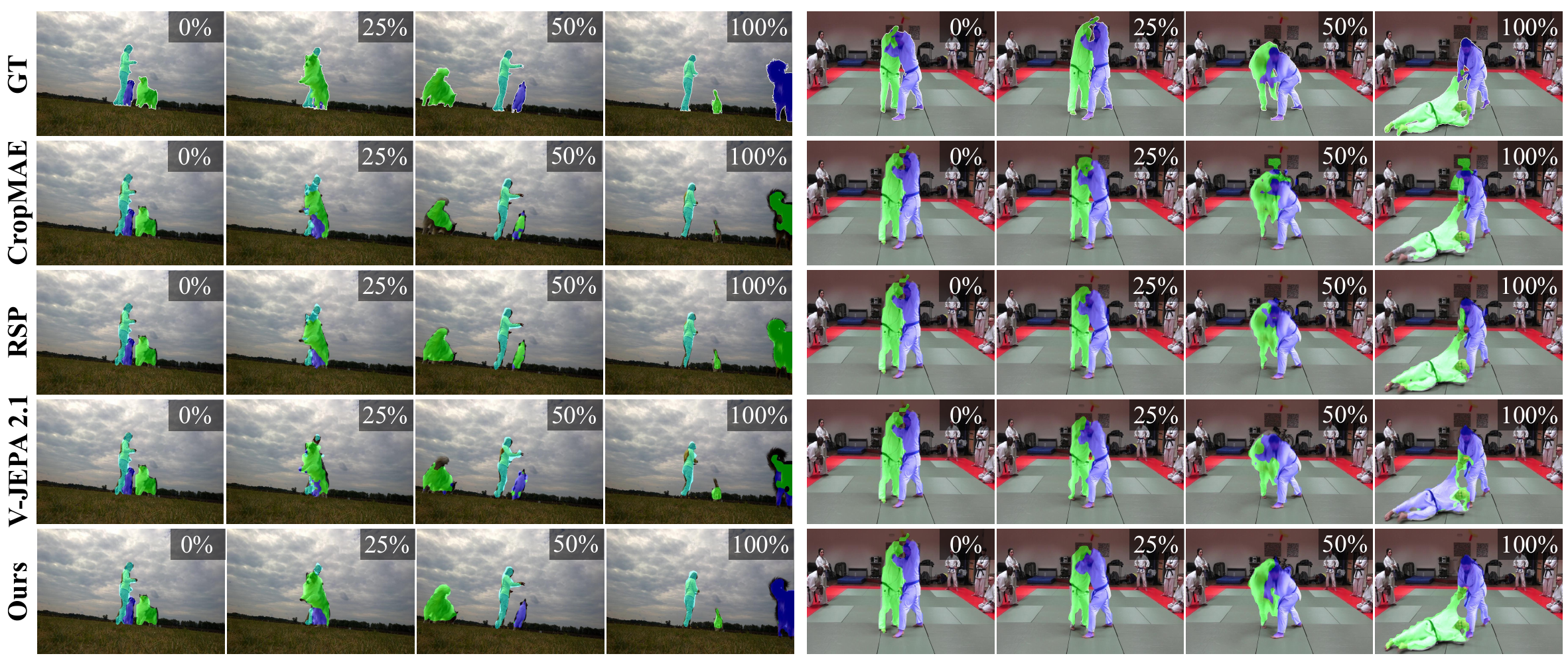}
    \vspace{-3pt}
    \caption{Qualitative visualization of object segmentation on DAVIS-2017. Our method produces cleaner segmentation masks with clearer object boundaries and better separation between different semantic regions. Best viewed in color.}
    \label{fig:vis_pixel_level}
    \vspace{-5pt}
\end{figure*}

\subsubsection{Evaluation on Pixel-level Tasks}
\label{subsec:evaluation_pixel}

\textbf{Evaluation protocol.} 
We first evaluate the quality of dense representations on three pixel-level video tasks, including video object segmentation on DAVIS-2017~\cite{DAVIS17}, human part segmentation on VIP~\cite{VIP}, and pose propagation on JHMDB~\cite{JHMDB}. 
Following recent works~\cite{SiamMAE,CropMAE,RSP}, all evaluations are conducted under a semi-supervised protocol without post-training or fine-tuning. 
Specifically, only the ground-truth annotation of the first frame is provided, and the model propagates pixel-wise predictions to subsequent frames using frozen pre-trained representations. 
This setting directly evaluates whether the learned dense representations preserve reliable correspondence across frames.

\textbf{Quantitative results.}
The evaluation results on the three pixel-level tasks are reported in \Tbref{tab:results_pixel_level}, from which we make the following observations. 
\textbf{1)} \methodname{} consistently achieves competitive or superior performance across tasks, metrics, and backbone architectures. These results suggest that modeling patch-level temporal correspondence and temporal variation benefits dense video representation learning.
\textbf{2)} Compared with pixel-space masked modeling methods, \methodname{} improves the results by at least $5.0\%$, $4.9\%$, and $1.2\%$ on $\mathcal{J}\&\mathcal{F}_{\mathrm{m}}$, mIoU, and PCK@0.1 with ViT-S/16, respectively. These gains indicate that latent-space temporal modeling provides an effective learning signal for dense video prediction, compared with reconstructing low-level pixel signals.
\textbf{3)} Compared with latent-space self-distillation methods on ViT-B/16, \methodname{} further improves $\mathcal{J}\&\mathcal{F}_{\mathrm{m}}$ by $2.0\%\sim3.9\%$. This improvement benefits from patch-level registration in the auxiliary branch and explicit modeling of temporal variation among corresponding representations. As a result, the learned representations encode more precise spatiotemporal structures for pixel-level video tasks.
\textbf{4)} When scaled up to the ViT-B/16 backbone, \methodname{} continues to bring consistent improvements over the corresponding baselines. Moreover, \methodname{} achieves results comparable to or better than V-JEPA 2.1 with the distilled ViT-B backbone. This further supports the effectiveness of variation-aware temporal modeling for video representation learning.

\textbf{Qualitative results.}
We further provide qualitative comparisons in \Fgref{fig:vis_pixel_level}. 
Compared with representative baselines, \methodname{} produces more coherent predictions with clearer object boundaries and better temporal consistency. 
The improvements are mainly reflected in regions with continuous motion and appearance changes, where the predictions remain more stable across frames. 
These results suggest that the proposed latent temporal modeling improves cross-frame correspondence while preserving local semantic discrimination for dense video prediction.

\subsubsection{Evaluation on Frame-level Tasks}
\label{subsec:evaluation_frame}
\textbf{Evaluation protocol.} 
We further evaluate the learned representations on frame-level video tasks, where temporally resolved features are required for recognizing action states and temporal boundaries. 
Frame-level representations are obtained by spatially averaging dense patch representations from each frame. 
For action segmentation, we conduct experiments on the third-person Breakfast~\cite{Breakfast} benchmark and the egocentric EgoProceL~\cite{EgoProceL} benchmark. The extracted features are fed into FACT~\cite{FACT}, which is trained for 150 epochs on Breakfast and 350 epochs on EgoProceL. 
For temporal action localization, we evaluate on the untrimmed THUMOS14~\cite{THUMOS14} dataset. Following the setting of AdaTAD~\cite{AdaTAD}, we use frame-level representations extracted from the Kinetics-400 fine-tuned model described in \cref{subsec:evaluation_video} and train AdaTAD for 60 epochs before evaluation.
For frame-level downstream tasks requiring per-frame representations, we adapt VideoMAE by using repeated-frame tubelets, allowing its original tubelet embedding to produce frame-wise representations without modifying the pretrained architecture.

\begin{table*}[htbp]
  \centering
  \caption{Evaluation results on frame-level tasks. The best and second-best results are highlighted in \textbf{bold} and \underline{underlined}. The metric mAP@[0.3:0.1:0.7] denotes the average mAP over IoU thresholds from 0.3 to 0.7 with a step size of 0.1.}
  \vspace{-5pt}
  \renewcommand{\arraystretch}{1.15}
  \large
  \resizebox{0.98\textwidth}{!}{%
    \begin{tabular}{ccc|ccc|ccc|cccc}
    \toprule
      \multirow{3}[2]{*}{Model}  & \multirow{3}[2]{*}{Backbone} & \multirow{3}[2]{*}{\multirow{2}{*}{\centering \shortstack{Training \\ Dataset}}} & \multicolumn{6}{c|}{\rule[-1.5ex]{0pt}{2ex}\textbf{Action Segmentation}} & \multicolumn{4}{c}{\textbf{Temporal Localization}} \\
      \cline{4-13}
      & & & \multicolumn{3}{c|}{\rule[-1ex]{0pt}{2ex}Breakfast} & \multicolumn{3}{c|}{\rule{0pt}{3ex}EgoProceL} & \multicolumn{4}{c}{THUMOS14} \\
      \cline{4-13}
      & & & \rule[-1.0ex]{0pt}{3.5ex}Edit & Acc & F1@0.10 & Edit & Acc & F1@0.10 & mAP@0.3 & mAP@0.5 & mAP@0.7 & mAP@[0.3:0.1:0.7] \\
    \midrule
      VideoMAE~\cite{VideoMAE}    & ViT-B/16  & K400  & \cellcolor[rgb]{ .651,  .788,  .914}58.1 & \cellcolor[rgb]{ .678,  .808,  .922}42.4 & \cellcolor[rgb]{ .682,  .808,  .922}55.9 & \cellcolor[rgb]{ .918,  .953,  .98}62.1 & \cellcolor[rgb]{ .89,  .937,  .976}70.0 & \cellcolor[rgb]{ .839,  .906,  .961}53.9 & \cellcolor[rgb]{ .71,  .843,  .616}\underline{78.1} & \cellcolor[rgb]{ .694,  .831,  .596}\underline{65.6} & \cellcolor[rgb]{ .694,  .831,  .596}\underline{40.1} & \cellcolor[rgb]{ .718,  .847,  .627}\underline{62.0}  \\
      SiamMAE~\cite{SiamMAE} & ViT-B/16 & K400  & \cellcolor[rgb]{ .875,  .925,  .973}53.1 & \cellcolor[rgb]{ .875,  .925,  .973}35.4 & \cellcolor[rgb]{ .855,  .918,  .965}51.4 & \cellcolor[rgb]{ .776,  .867,  .945}66.2 & \cellcolor[rgb]{ .816,  .89,  .957}70.5 & \cellcolor[rgb]{ .651,  .788,  .914}\underline{60.0} & \cellcolor[rgb]{ .906,  .949,  .878}66.8 & \cellcolor[rgb]{ .894,  .945,  .867}55.0 & \cellcolor[rgb]{ .89,  .941,  .859}30.8 & \cellcolor[rgb]{ .898,  .945,  .871}51.6  \\
      CropMAE~\cite{CropMAE}  & ViT-B/16  & K400  & \cellcolor[rgb]{ .961,  .98,  .992}46.0 & \cellcolor[rgb]{ .961,  .98,  .992}27.4 & \cellcolor[rgb]{ .961,  .98,  .992}44.1 & \cellcolor[rgb]{ .867,  .922,  .969}63.6 & \cellcolor[rgb]{ .655,  .792,  .914}\underline{71.6} & \cellcolor[rgb]{ .675,  .804,  .922}59.3 & \cellcolor[rgb]{ .984,  .992,  .976}54.6 & \cellcolor[rgb]{ .984,  .992,  .976}43.0 & \cellcolor[rgb]{ .984,  .992,  .976}22.4 & \cellcolor[rgb]{ .984,  .992,  .976}40.7 \\
      RSP$^\dagger$~\cite{RSP} & ViT-B/16  & K400   & \cellcolor[rgb]{ .918,  .953,  .98}52.1 & \cellcolor[rgb]{ .918,  .953,  .98}33.8 & \cellcolor[rgb]{ .918,  .953,  .98}49.8 & \cellcolor[rgb]{ .878,  .929,  .973}\underline{63.3} & \cellcolor[rgb]{ .918,  .953,  .98}69.8 & \cellcolor[rgb]{ .918,  .953,  .98}51.3 & \cellcolor[rgb]{ .886,  .937,  .855}69.6 & \cellcolor[rgb]{ .886,  .937,  .855}56.0 & \cellcolor[rgb]{ .886,  .937,  .855}30.9 & \cellcolor[rgb]{ .886,  .937,  .855}52.9   \\
      Ada-iBOT~\cite{co_settle} & ViT-B/16   &  INet + K400  & \cellcolor[rgb]{ .694,  .816,  .925}57.1 & \cellcolor[rgb]{ .671,  .8,  .918}42.7 & \cellcolor[rgb]{ .749,  .851,  .941}54.1 & \cellcolor[rgb]{ .694,  .816,  .925}\underline{68.5} & \cellcolor[rgb]{ .89,  .937,  .976}70.0 & \cellcolor[rgb]{ .682,  .808,  .922}59.0 & \cellcolor[rgb]{ .749,  .867,  .671}77.8 & \cellcolor[rgb]{ .722,  .851,  .635}65.0 & \cellcolor[rgb]{ .776,  .878,  .706}37.7 & \cellcolor[rgb]{ .776,  .878,  .706}61.2 \\
      Ada-DINO v2~\cite{co_settle} & ViT-B/16   &  INet + K400  & \cellcolor[rgb]{ .627,  .773,  .91}\underline{58.6} & \cellcolor[rgb]{ .608,  .761,  .902}\textbf{44.9} & \cellcolor[rgb]{ .608,  .761,  .902}\textbf{57.7} & \cellcolor[rgb]{ .737,  .843,  .937}67.3 & \cellcolor[rgb]{ .773,  .863,  .945}70.8 & \cellcolor[rgb]{ .749,  .847,  .937}56.9 & \cellcolor[rgb]{ .776,  .878,  .706}77.6 & \cellcolor[rgb]{ .776,  .878,  .706}63.9 & \cellcolor[rgb]{ .71,  .843,  .616}39.6 & \cellcolor[rgb]{ .769,  .875,  .694}61.3 \\ 
      \methodnameb{} \textbf{(Ours)}   & ViT-B/16  & K400  & \cellcolor[rgb]{ .608,  .761,  .902}\textbf{59.0} & \cellcolor[rgb]{ .667,  .8,  .918}\underline{42.8} & \cellcolor[rgb]{ .62,  .769,  .906}\underline{57.5} & \cellcolor[rgb]{ .608,  .761,  .902}\textbf{70.9} & \cellcolor[rgb]{ .608,  .761,  .902}\textbf{71.9} & \cellcolor[rgb]{ .608,  .761,  .902}\textbf{61.4} & \cellcolor[rgb]{ .663,  .816,  .557}\textbf{78.4} & \cellcolor[rgb]{ .663,  .816,  .557}\textbf{66.2} & \cellcolor[rgb]{ .663,  .816,  .557}\textbf{40.8} & \cellcolor[rgb]{ .663,  .816,  .557}\textbf{62.7} \\
    \bottomrule
    \end{tabular}%
    }
  \label{tab:results_frame_level}%
  \vspace{-3pt}
\end{table*}

\begin{table*}[htbp]
  \centering
  \caption{Evaluation results on video-level tasks. The best and second-best results are highlighted in \textbf{bold} and \underline{underlined}.}
  \vspace{-5pt}
  \renewcommand{\arraystretch}{1.15}
  \large
  \resizebox{0.98\textwidth}{!}{%
    \begin{tabular}{ccc|cc|cc|cc|cc|cc}
    \toprule
      \multirow{3}[2]{*}{Model}  & \multirow{3}[2]{*}{Backbone} & \multirow{3}[2]{*}{\multirow{2}{*}{\centering \shortstack{Training \\ Dataset}}} & \multicolumn{4}{c|}{\rule[-1.5ex]{0pt}{2ex}\textbf{Action Classification}} & \multicolumn{4}{c|}{\textbf{Video Retrieval}} & \multicolumn{2}{c}{\textbf{Chiral Recognition}} \\
      \cline{4-13}
      & & & \multicolumn{2}{c|}{\rule[-1.5ex]{0pt}{3ex}Kinetics-400} & \multicolumn{2}{c|}{\rule{0pt}{3ex}HMDB51} & \multicolumn{2}{c|}{UCF101} & \multicolumn{2}{c|}{HMDB51}  & \multicolumn{1}{c}{Charades}  & \multicolumn{1}{c}{SSV2} \\
      \cline{4-13}
      & & & \rule[-1.5ex]{0pt}{3.0ex} Acc@1 & Acc@5 & Acc@1 & Acc@5 & R@1 & R@10 & R@1 & R@10 & Acc & Acc  \\
    \midrule
      VideoMAE~\cite{VideoMAE}    & ViT-B/16  & K400  & \cellcolor[rgb]{ .878,  .929,  .973}65.8 & \cellcolor[rgb]{ .878,  .929,  .973}86.2 & \cellcolor[rgb]{ .918,  .953,  .98}49.5 & \cellcolor[rgb]{ .914,  .949,  .98}79.4 & \cellcolor[rgb]{ .894,  .941,  .863}94.3 & \cellcolor[rgb]{ .894,  .945,  .867}97.6 & \cellcolor[rgb]{ .886,  .937,  .855}65.9 & \cellcolor[rgb]{ .882,  .937,  .847}84.1 & \cellcolor[rgb]{ .902,  .839,  .949}64.0 & \cellcolor[rgb]{ .902,  .839,  .949}77.8 \\
      SiamMAE~\cite{SiamMAE} & ViT-B/16 & K400  & \cellcolor[rgb]{ .918,  .953,  .98}57.5 & \cellcolor[rgb]{ .918,  .953,  .98}79.9 & \cellcolor[rgb]{ .91,  .949,  .98}51.0 & \cellcolor[rgb]{ .918,  .953,  .98}78.9 & \cellcolor[rgb]{ .984,  .992,  .976}89.8 & \cellcolor[rgb]{ .984,  .992,  .976}94.1 & \cellcolor[rgb]{ .984,  .992,  .976}58.7 & \cellcolor[rgb]{ .984,  .992,  .976}75.7 & \cellcolor[rgb]{ .961,  .937,  .98}62.0 & \cellcolor[rgb]{ .941,  .902,  .969}76.2 \\
      CropMAE~\cite{CropMAE}  & ViT-B/16  & K400  & \cellcolor[rgb]{ .961,  .976,  .992}48.2 & \cellcolor[rgb]{ .961,  .976,  .992}73.1 & \cellcolor[rgb]{ .961,  .976,  .992}45.0 & \cellcolor[rgb]{ .961,  .976,  .992}76.6 & \cellcolor[rgb]{ .957,  .976,  .941}91.3 & \cellcolor[rgb]{ .953,  .973,  .933}95.5 & \cellcolor[rgb]{ .937,  .965,  .918}62.4 & \cellcolor[rgb]{ .949,  .973,  .929}78.9 & \cellcolor[rgb]{ .965,  .941,  .98}61.9 & \cellcolor[rgb]{ .965,  .941,  .98}75.1 \\
      RSP$^\dagger$~\cite{RSP} & ViT-B/16  & K400   & \cellcolor[rgb]{ .898,  .941,  .976}61.7 & \cellcolor[rgb]{ .898,  .941,  .976}83.1 & \cellcolor[rgb]{ .875,  .925,  .973}55.4 & \cellcolor[rgb]{ .89,  .937,  .976}81.3 & \cellcolor[rgb]{ .8,  .89,  .737}\underline{95.6} & \cellcolor[rgb]{ .804,  .894,  .745}98.3 & \cellcolor[rgb]{ .714,  .847,  .627}\underline{70.1} & \cellcolor[rgb]{ .804,  .894,  .745}\underline{86.0} & \cellcolor[rgb]{ .863,  .773,  .929}66.8  & \cellcolor[rgb]{ .863,  .773,  .929}81.2 \\
      Ada-iBOT~\cite{co_settle} & ViT-B/16   &  INet + K400  & \cellcolor[rgb]{ .655,  .792,  .918}\underline{71.1} & \cellcolor[rgb]{ .678,  .804,  .922}\underline{88.8} & \cellcolor[rgb]{ .714,  .827,  .929}\underline{56.9} & \cellcolor[rgb]{ .678,  .808,  .922}\underline{85.6} & \cellcolor[rgb]{ .855,  .922,  .816}94.9 & \cellcolor[rgb]{ .796,  .886,  .733}98.4 & \cellcolor[rgb]{ .918,  .953,  .89}63.9 & \cellcolor[rgb]{ .89,  .941,  .859}83.8 & \cellcolor[rgb]{ .796,  .663,  .898}67.7 & \cellcolor[rgb]{ .855,  .761,  .925}81.4 \\
      Ada-DINO v2~\cite{co_settle} & ViT-B/16   &  INet + K400  & \cellcolor[rgb]{ .733,  .839,  .937}70.0 & \cellcolor[rgb]{ .725,  .835,  .933}88.5 & \cellcolor[rgb]{ .725,  .835,  .933}56.8 & \cellcolor[rgb]{ .706,  .824,  .929}85.3 & \cellcolor[rgb]{ .831,  .91,  .78}95.2 & \cellcolor[rgb]{ .718,  .847,  .631}\underline{98.7} & \cellcolor[rgb]{ .843,  .914,  .796}67.0 & \cellcolor[rgb]{ .835,  .91,  .788}85.2 & \cellcolor[rgb]{ .784,  .643,  .894}\underline{67.9} & \cellcolor[rgb]{ .773,  .624,  .886}\underline{83.7} \\
      \methodnameb{} \textbf{(Ours)}   & ViT-B/16  & K400  & \cellcolor[rgb]{ .608,  .761,  .902}\textbf{71.8} & \cellcolor[rgb]{ .608,  .761,  .902}\textbf{89.4} & \cellcolor[rgb]{ .608,  .761,  .902}\textbf{57.4} & \cellcolor[rgb]{ .608,  .761,  .902}\textbf{86.4} & \cellcolor[rgb]{ .663,  .816,  .557}\textbf{97.2} & \cellcolor[rgb]{ .663,  .816,  .557}\textbf{99.0} & \cellcolor[rgb]{ .663,  .816,  .557}\textbf{71.4} & \cellcolor[rgb]{ .663,  .816,  .557}\textbf{89.6} & \cellcolor[rgb]{ .741,  .573,  .871}\textbf{68.4} & \cellcolor[rgb]{ .741,  .573,  .871}\textbf{84.5} \\
    \bottomrule
    \end{tabular}%
    }  
  \label{tab:results_video_level}%
  \vspace{-8pt}
\end{table*}

\textbf{Evaluation results.}
The results are reported in \Tbref{tab:results_frame_level}. 
Overall, \methodname{} brings consistent gains on frame-level tasks, showing that the learned dense representations can be converted into useful frame-wise features for temporal understanding. 
For action segmentation, \methodname{} improves the Edit scores on Breakfast and EgoProceL to $59.0\%$ and $70.9\%$, with gains of $0.4\%$ and $2.4\%$, respectively.
By reconstructing and aligning representations in the latent space, the training objective is imposed on semantic representations rather than low-level appearance signals, improving the temporal separability of different action states.
For temporal localization on THUMOS14, \methodname{} obtains a $0.7\%$ improvement in average mAP over IoU thresholds from 0.3 to 0.7, indicating that the learned frame representations provide clearer cues for delimiting action intervals across different localization criteria.
These frame-level results show that the proposed temporal variation modeling is effective for temporal parsing tasks, suggesting that \methodname{} learns variation-aware representations that better reflect action transitions and boundaries.

\subsubsection{Evaluation on Video-level Tasks}
\label{subsec:evaluation_video}

\textbf{Evaluation protocol.} 
To evaluate the global semantic modeling ability of the learned representations, we consider three video-level tasks: action classification, video retrieval, and chiral recognition. For action classification and video retrieval, video-level representations are obtained by average pooling dense features over the temporal and spatial dimensions. For chiral recognition, we preserve temporal ordering by concatenating frame-level representations along the temporal dimension to form the final representation. 
Action classification is evaluated on Kinetics-400~\cite{Kinetics} and HMDB51~\cite{HMDB51} under the full fine-tuning setting with 50 and 100 fine-tuning epochs, respectively. Video retrieval follows the standard zero-shot protocol as in previous work~\cite{s2vs,liu2024not} on UCF101~\cite{UCF101} and HMDB51~\cite{HMDB51}, where pre-trained features are directly used without task-specific fine-tuning. Chiral recognition is evaluated by linear probing on Chiral Charades and Chiral SSv2~\cite{Chirality}, which tests whether the representation can distinguish temporally ordered visual patterns.
The VideoMAE method follows the same frame-independent adaptation as in \cref{subsec:evaluation_frame}.

\textbf{Evaluation results.} 
The results are reported in \Tbref{tab:results_video_level}. 
Overall, \methodname{} achieves consistent improvements on video-level tasks, showing that it improves global semantic discrimination while retaining stable temporal cues for video-level understanding.
For action classification, \methodname{} outperforms video pre-training baselines by $6.0\%$ Acc@1 on Kinetics-400 and $2.0\%$ Acc@1 on HMDB51, and also surpasses adapted image-pretrained baselines with additional pre-training data. 
For zero-shot video retrieval, \methodname{} consistently improves retrieval performance on UCF101 and HMDB51, with gains of at least $1.6\%$ and $1.3\%$ in R@1, respectively. 
This result suggests that the learned representations provide better semantic consistency across videos and transfer more directly to video-level semantic matching. 
For chiral recognition, \methodname{} achieves the best performance on both Chiral Charades and Chiral SSv2, suggesting that temporal variation modeling helps encode the temporal progression of visual patterns.

\begin{table}[t]
  \centering
  \caption{Evaluation results on vision-based robotic learning tasks on CortexBench benchmark. The best and second-best results are marked in \textbf{bold} and \underline{underlined}, respectively. $^\dagger$ denotes results reported from prior work~\cite{RSP}. $^\ddagger$ indicates that VideoMAE uses averaged patch-token representations as its standard encoder does not employ a \texttt{[CLS]} token, while the other methods use the \texttt{[CLS]} token.}
  \vspace{-3pt}
  \renewcommand{\arraystretch}{1.2}
  \large
  \resizebox{0.49\textwidth}{!}{%
    \begin{tabular}{cc|cccc|c}
    \toprule
    \multirow{2}[2]{*}{Method} & \multirow{2}[2]{*}{Backbone} & \multicolumn{5}{c}{\textbf{Vision-based Robotic Learning}} \\
    \cline{3-7}
          &       & \rule[-1.0ex]{0pt}{3.5ex}Adroit & \rule[-1.0ex]{0pt}{3.5ex}MetaWorld & \rule[-1.0ex]{0pt}{3.5ex}DMC   & \rule[-1.0ex]{0pt}{3.5ex}Trifinger & \rule[-1.0ex]{0pt}{3.5ex}Avg. \\
    \midrule
    SimCLR$^\dagger$~\cite{SimCLRv1} & ViT-S/16 & \cellcolor[rgb]{ .996,  .957,  .91}40.4  & \cellcolor[rgb]{ .996,  .941,  .878}78.4  & \cellcolor[rgb]{ .996,  .957,  .91}39.7  & \cellcolor[rgb]{ .992,  .898,  .792}63.3 & \cellcolor[rgb]{ .992,  .914,  .824}55.5 \\
    MoCo v3$^\dagger$~\cite{MoCov3} & ViT-S/16 & \cellcolor[rgb]{ .992,  .937,  .875}39.6  & \cellcolor[rgb]{ .992,  .937,  .875}65.4  & \cellcolor[rgb]{ .992,  .925,  .847}43.7  & \cellcolor[rgb]{ .996,  .957,  .91}53.3 & \cellcolor[rgb]{ .996,  .957,  .91}50.5 \\
    DINO$^\dagger$~\cite{DINO} & ViT-S/16 & \cellcolor[rgb]{ .988,  .89,  .773}45.6 & \cellcolor[rgb]{ .988,  .875,  .749}\underline{82.4}  & \cellcolor[rgb]{ .988,  .831,  .678}50.9  & \cellcolor[rgb]{ .988,  .89,  .773}64.2 & \cellcolor[rgb]{ .988,  .843,  .694}60.8 \\
    MAE$^\dagger$~\cite{MAE} & ViT-S/16 & \cellcolor[rgb]{ .992,  .898,  .792}44.8  & \cellcolor[rgb]{ .988,  .894,  .78}81.4  & \cellcolor[rgb]{ .988,  .812,  .643}\underline{52.1}  & \cellcolor[rgb]{ .992,  .906,  .804}62.2 & \cellcolor[rgb]{ .988,  .851,  .714}60.1 \\
    \midrule
    VideoMAE$^\ddagger$~\cite{VideoMAE}  & ViT-S/16 & \cellcolor[rgb]{ .988,  .855,  .714}\underline{48.0} & \cellcolor[rgb]{ .988,  .894,  .78}81.3 & \cellcolor[rgb]{ .992,  .937,  .875}28.8 & \cellcolor[rgb]{ .984,  .757,  .553}\textbf{71.9} & \cellcolor[rgb]{ .988,  .894,  .784}57.5 \\ 
    SiamMAE~\cite{SiamMAE} & ViT-S/16 & \cellcolor[rgb]{ .988,  .894,  .78}45.3  & \cellcolor[rgb]{ .996,  .957,  .91}77.3  & \cellcolor[rgb]{ .992,  .91,  .816}45.7  & \cellcolor[rgb]{ .992,  .922,  .839}59.2 & \cellcolor[rgb]{ .992,  .902,  .796}56.9 \\
    CropMAE~\cite{CropMAE} & ViT-S/16 & \cellcolor[rgb]{ .992,  .91,  .816}44.0 & \cellcolor[rgb]{ .992,  .914,  .824}80.0 & \cellcolor[rgb]{ .992,  .922,  .839}44.1 & \cellcolor[rgb]{ .992,  .937,  .875}49.8 & \cellcolor[rgb]{ .992,  .922,  .839}54.5 \\
    RSP~\cite{RSP} & ViT-S/16 & \cellcolor[rgb]{ .992,  .918,  .831}43.3 & \cellcolor[rgb]{ .992,  .898,  .792}81.1   & \cellcolor[rgb]{ .984,  .757,  .553}\textbf{54.9}  & \cellcolor[rgb]{ .988,  .804,  .631}69.2 & \cellcolor[rgb]{ .988,  .82,  .659}\underline{62.1} \\
    \methodnameb{} \textbf{(Ours)} & ViT-S/16 & \cellcolor[rgb]{ .984,  .757,  .553}\textbf{54.7}       & \cellcolor[rgb]{ .984,  .757,  .553}\textbf{88.8}       & \cellcolor[rgb]{ .988,  .863,  .733}49.2    & \cellcolor[rgb]{ .988,  .776,  .588}\underline{70.7} & \cellcolor[rgb]{ .984,  .757,  .553}\textbf{65.8} \\
    \bottomrule
    \end{tabular}%
  }%
  \label{tab:results_robot_learning}%
  \vspace{-10pt}
\end{table}

\subsubsection{Evaluation on Robotic Learning Tasks}
\label{subsec:evaluation_robotic}
To examine whether the learned representations capture temporally coherent visual cues for vision-based robotic learning, we further evaluate them on the CortexBench benchmark~\cite{CortexBench_1,CortexBench_2}.

\textbf{Benchmarks.} 
We select four domains from the CortexBench benchmark, covering diverse robotic learning scenarios as follows.
1) Adroit~\cite{Adroit} contains dexterous manipulation tasks with a high degree-of-freedom anthropomorphic hand, where we evaluate on \texttt{Relocate} and \texttt{Reorient-Pen}.
2) MetaWorld~\cite{MetaWorld} focuses on tabletop object manipulation with a robotic arm, and we consider five representative tasks: \texttt{Assembly}, \texttt{Bin-Picking}, \texttt{Button-Press}, \texttt{Drawer-Open}, and \texttt{Hammer}.
3) DeepMind Control (DMC)~\cite{DMC} provides image-based continuous control tasks, including \texttt{Finger-Spin}, \texttt{Reacher-Easy}, \texttt{Cheetah-Run}, \texttt{Walker-Stand}, and \texttt{Walker-Walk}.
4) TriFinger~\cite{TriFinger} involves precise object manipulation with a three-finger robotic system, and we evaluate on the \texttt{Reach-Cube} and \texttt{Move-Cube}.

\textbf{Evaluation protocol.} 
Following prior work~\cite{RSP,CortexBench_1}, all tasks are evaluated under an imitation learning setting in simulated environments such as MuJoCo~\cite{Mujoco}. Policies are trained by behavior cloning from expert demonstrations and evaluated through closed-loop execution based on visual observations. We use 25 demonstrations for MetaWorld and 100 demonstrations for Adroit, DMC, and TriFinger, with all policies trained for 100 epochs. 
Visual representations are extracted from frozen backbones and used as policy inputs through either the \texttt{[CLS]} token or average-pooled patch tokens. Each task is evaluated with three random seeds, and the domain-level score is averaged over tasks and seeds. We report normalized scores for DMC, success rates for Adroit and MetaWorld, and task rewards for TriFinger. Following CortexBench, we report all metrics on a $[0,100]$ scale, where higher values indicate better performance.

\begin{figure}[t]
  \centering
     \hfill
     \includegraphics[width=0.46\linewidth]{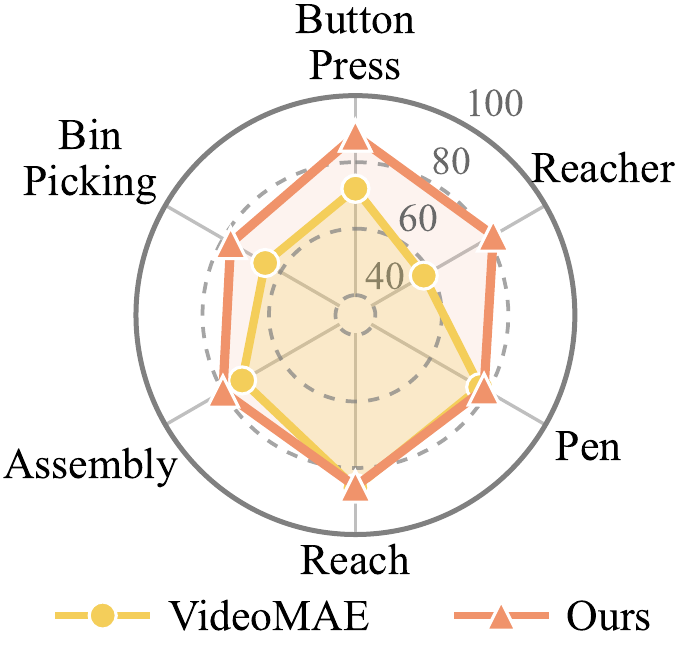}
     \hfill
     \includegraphics[width=0.46\linewidth]{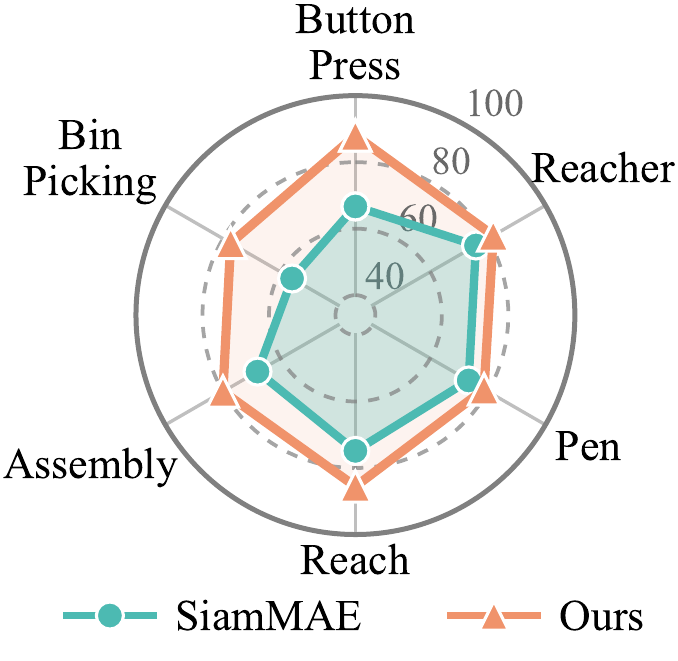}
     \hfill
     \\
     \hfill
     \includegraphics[width=0.46\linewidth]{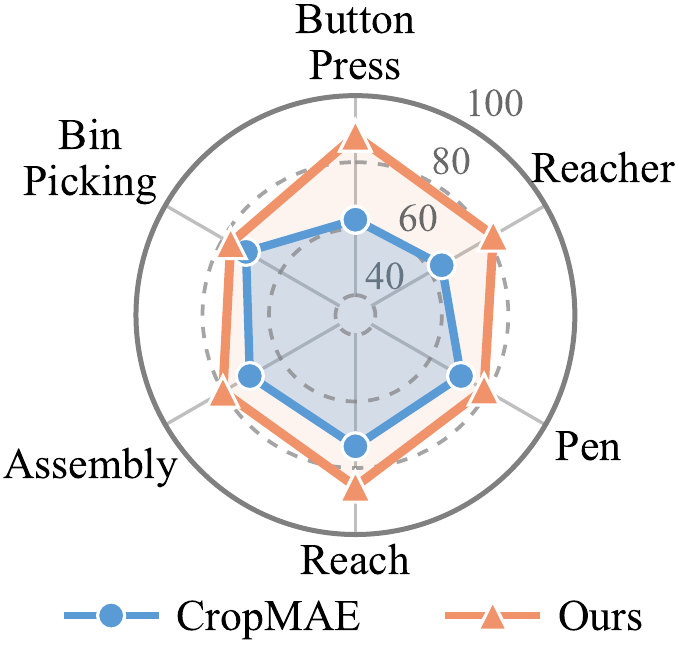}
     \hfill
     \includegraphics[width=0.46\linewidth]{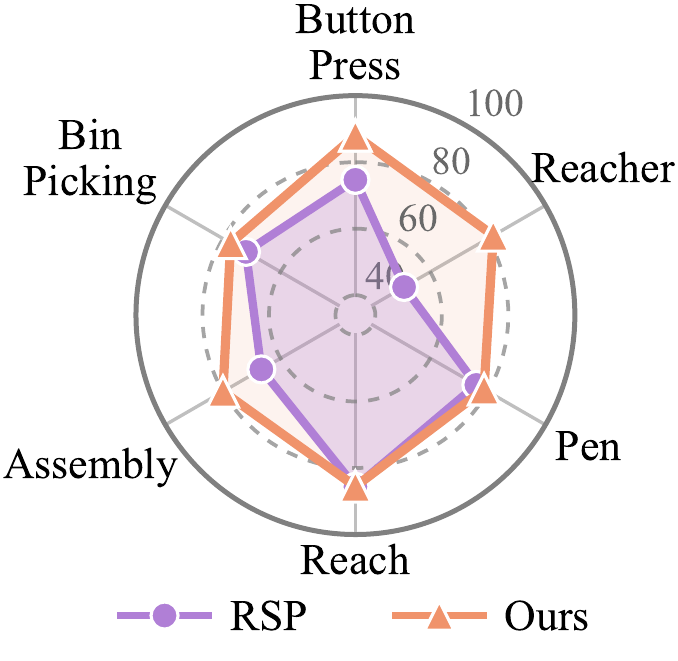}
     \hfill
    \caption{Task-wise comparison on selected CortexBench tasks.}
    \label{fig:vis_robot_learning_results}
    \vspace{-10pt}
\end{figure}

\textbf{Evaluation results.}
The results are reported in \Tbref{tab:results_robot_learning}. 
Overall, \methodname{} achieves the highest average score of $65.8\%$ across the four domains, outperforming the compared methods by at least $3.7\%$. 
\Fgref{fig:vis_robot_learning_results} further presents task-wise comparisons on representative tasks from these domains. 
These tasks require the policy to infer object states, end-effector positions, and task progress from visual observations, where the relevant cues are often reflected by continuous changes across time. 
By constraining temporal variation in the latent space, \methodname{} encourages representation changes to better reflect the underlying state evolution, which can reduce the influence of static appearance bias and frame-level noise.
Such representations provide more informative visual inputs for behavior cloning, suggesting the potential of variation-aware temporal modeling for vision-based robotic learning with continuous visual state changes.

\subsection{Pre-training Efficiency Analysis}
\label{subsec:efficiency_analysis}



We compare the pre-training efficiency of different methods in Tab.~{\color{blue}6}.
Throughput is measured on NVIDIA RTX 4090 GPUs with a global batch size of 128 and FP16 mixed precision, averaging over 200 updates after stabilization.
Each method retains its native pre-training inputs.
The reported training FLOPs count forward and loss computation across all executed branches, while inference FLOPs are measured at $480\times880$ on DAVIS.
Compared with VideoMAE, our method substantially reduces both training time and memory cost.
SiamMAE, CropMAE, and RSP use sparsely sampled conditional frames, whereas our method introduces structured two-sided temporal references for latent-space representation reconstruction.
Although these designs increase pre-training FLOPs, they provide explicit temporal constraints and enable effective pre-training with fewer epochs.
The shorter pre-training schedule helps offset the additional computation per video, allowing our method to achieve stronger performance on dense video downstream tasks at a manageable training cost.
Since PRM and the auxiliary branches are used only during pre-training, the backbone inference cost remains comparable to the other methods at 194.6 GFLOPs/frame.

We further analyze the efficiency of individual components in \Tbref{tab:efficiency_decomposition}.
The self-distillation architecture contains the student and teacher branches and accounts for 97.53\% of the parameters and 86.75\% of active GPU memory, making it the main source of model storage.
The auxiliary branch shares the student backbone, avoiding separate parameter, gradient, and optimizer-state storage.
PRM provides attention-based patch alignment with limited additional capacity and training cost, accounting for 2.38\% of the parameters, 8.09\% of active GPU memory, and 7.06\% of training time.
Dynamic modeling uses a simple projection layer and learnable trajectory coefficients, keeping its parameter, active GPU memory, and training-time shares at 0.09\%, 2.55\%, and 1.97\%, respectively.
Overall, these results indicate that backbone sharing and the limited resource requirements of PRM and dynamic modeling allow the additional temporal constraints to be incorporated at a moderate additional pre-training cost.


\begin{table}[!t]
    \centering
    \caption{End-to-end pre-training efficiency comparison. All comparisons are conducted on the same device.}
    \label{tab:results_efficiency}
    \vspace{-5pt}
    \setlength{\tabcolsep}{1.5pt}
    \renewcommand{\arraystretch}{1.2}
    \resizebox{\linewidth}{!}{%
    \begin{tabular}{c|ccc|ccc|c}
        \toprule
        \multirow{2}{*}{Method}
        & \multirow{2}{*}{Epoch $\downarrow$}
        & \multirow{2}{*}{\makecell{GPU Mem. \\ (GB) $\downarrow$}}
        & \multirow{2}{*}{\makecell{Pretrain \\ Time (h) $\downarrow$}}
        & \multirow{2}{*}{\makecell{Throughput \\ (videos/s) $\uparrow$}}
        & \multirow{2}{*}{\makecell{Train FLOPs \\ (G/video) $\downarrow$}}
        & \multirow{2}{*}{\makecell{Infer. FLOPs \\ (G/frame) $\downarrow$}}
        & DAVIS17 \\
        & & & & & & & $\mathcal{J}\&\mathcal{F}_{\mathrm{m}}\uparrow$ \\
        \midrule
        VideoMAE       & 800 & 8$\times$23.42 & $\sim160$ & 297.6  & 36.0  & 194.8 & 55.5 \\
        SiamMAE        & 400 & 4$\times$5.04  & $\sim21$  & 1269.2 & 20.2  & 194.3 & \underline{60.9} \\
        CropMAE        & 400 & 4$\times$5.02  & $\sim22$  & 1129.1 & 19.6  & 194.3 & 57.8 \\
        RSP            & 400 & 4$\times$12.19 & $\sim113$ & 417.6  & 62.4  & 194.3 & 60.5 \\
        \methodnameb{} & 200 & 4$\times$15.98 & $\sim31$  & 447.8  & 168.1 & 194.6 & \textbf{67.0} \\
        \bottomrule
    \end{tabular}%
    }
    \vspace{-3pt}
\end{table}

\begin{table}[!t]
    \centering
    \caption{Module-level pre-training efficiency of \methodname{}.
    The percentages denote shares of full-model totals.
    $^{\dagger}$: joint timing of self-distillation and the auxiliary branch.}
    \label{tab:efficiency_decomposition}
    \vspace{-5pt}
    \setlength{\tabcolsep}{1.5pt}
    \renewcommand{\arraystretch}{1.2}
    \resizebox{\linewidth}{!}{%
    \begin{tabular}{l|ccc}
        \toprule
        Component & Param. (M) & Active GPU Mem. (GB) & Training Time (ms/iter) \\
        \midrule
        self-distillation & 217.92 (97.53\%) & 10.92 (86.75\%) & \multirow{2}{*}{309.42 (89.15\%)$^{\dagger}$} \\
        auxiliary branch  & 0.00 (0.00\%)   & 0.33 (2.61\%)   & \\
        PRM               & 5.32 (2.38\%)   & 1.02 (8.09\%)   & 24.51 (7.06\%) \\
        dynamic modeling  & 0.20 (0.09\%)   & 0.32 (2.55\%)   & 6.82 (1.97\%) \\
        module interaction & -             & -              & 6.32 (1.82\%) \\
        \midrule
        \textbf{\methodname{}} & \textbf{223.44 (100\%)} & \textbf{12.58 (100\%)} & \textbf{347.07 (100\%)} \\
        \bottomrule
    \end{tabular}%
    }
    \vspace{-10pt}
\end{table}

\begin{table}[t]
    \centering
    \caption{Ablation study results based on ViT-B/16 backbone. The default settings are highlighted in \colorbox[rgb]{ .867,  .922,  .969}{blue}.}
    \label{tab:ablation}
    \setlength{\tabcolsep}{3.2pt}
    \renewcommand{\arraystretch}{1.05}
    \resizebox{\columnwidth}{!}{
    \begin{tabular}{c|ccc|ccc}
        \toprule
        \multirow{2}{*}{No.} 
        & \multirow{2}{*}{\makecell{Param. \\ Group $K$}}
        & \multirow{2}{*}{\makecell{Retained Patch \\Ratio $(\rho)$}}
        & \multirow{2}{*}{\makecell{Projection\\Dim. $d'$}}
        & DAVIS17 & VIP & JHMDB \\
        & & & 
        & $\mathcal{J}\&\mathcal{F}_{\mathrm{m}}$ 
        & mIoU 
        & PCK@0.1 \\
        \midrule
        1  & \ding{55} & \ding{55}  & \ding{55}  & 65.3 & 39.4 & 46.8 \\
        2  & 3  & \ding{55}  & \ding{55}  & 65.7 & 39.6 & 47.3 \\
        3  & 3  & 0.6 & \ding{55}  & 66.1 & 40.0 & 47.3 \\
        \cellcolor[rgb]{ .867,  .922,  .969}4  & \cellcolor[rgb]{ .867,  .922,  .969}3
           & \cellcolor[rgb]{ .867,  .922,  .969}0.6
           & \cellcolor[rgb]{ .867,  .922,  .969}256
           & \cellcolor[rgb]{ .867,  .922,  .969}\textbf{67.0}
           & \cellcolor[rgb]{ .867,  .922,  .969}\textbf{40.6}
           & \cellcolor[rgb]{ .867,  .922,  .969}\textbf{47.7} \\
        \midrule
        5  & 1  & 0.6 & 256 & 64.5 & 39.7 & 47.3 \\
        6  & 10 & 0.6 & 256 & 66.2 & 39.8 & 46.9 \\
        7  & 30 & 0.6 & 256 & 65.6 & 39.6 & 46.7 \\
        \midrule
        8  & 3 & 0.2 & 256 & 65.7 & 39.4 & 46.6 \\
        9  & 3 & 0.4 & 256 & 65.8 & 39.5 & 46.9 \\
        10 & 3 & 0.8 & 256 & 65.7 & 40.4 & 46.8 \\
        \midrule
        11 & 3 & 0.6 & 64  & 65.6 & 39.8 & 47.2 \\
        12 & 3 & 0.6 & 128 & 66.0 & 40.2 & 47.2 \\
        13 & 3 & 0.6 & 512 & 64.3 & 39.1 & 45.6 \\
        \bottomrule
    \end{tabular}
    }
    \vspace{-3pt}
\end{table}

\begin{table}[h]
    \centering
    \caption{Ablation study on training settings based on ViT-B/16 backbone. The default settings are highlighted in \colorbox[rgb]{ .867,  .922,  .969}{blue}.}
    \vspace{-3pt}
    \label{tab:ablation_training}
    \setlength{\tabcolsep}{3.2pt}
    \renewcommand{\arraystretch}{1.05}
    \resizebox{\columnwidth}{!}{%
    \begin{tabular}{c|cccc|ccc}
        \toprule
        \multirow{2}{*}{No.}
        & \multirow{2}{*}{\makecell{$\mathcal{L}_{\mathrm{var}}$ Start\\$(\xi_s)$}}
        & \multirow{2}{*}{\makecell{Mask\\Prob.}}
        & \multirow{2}{*}{\makecell{Mask Ratio\\$[\mathrm{Min},\mathrm{Max}]$}}
        & \multirow{2}{*}{\makecell{Sampling\\FPS}}
        & DAVIS17 & VIP & JHMDB \\
        & & & & 
        & $\mathcal{J}\&\mathcal{F}_{\mathrm{m}}$
        & mIoU
        & PCK@0.1 \\
        \midrule
        1 & 0.0 & 0.5 & $[0.1,0.5]$ & 2 & 64.9 & 38.7 & 46.1 \\
        2 & 0.2 & 0.5 & $[0.1,0.5]$ & 2 & 65.5 & 39.3 & 47.4 \\
        3 & 0.4 & 0.5 & $[0.1,0.5]$ & 2 & 66.6 & 39.8 & 47.5 \\
        \cellcolor[rgb]{.867,.922,.969}4
          & \cellcolor[rgb]{.867,.922,.969}0.6
          & \cellcolor[rgb]{.867,.922,.969}0.5
          & \cellcolor[rgb]{.867,.922,.969}$[0.1,0.5]$
          & \cellcolor[rgb]{.867,.922,.969}2
          & \cellcolor[rgb]{ .867,  .922,  .969}\textbf{67.0}
          & \cellcolor[rgb]{ .867,  .922,  .969}\textbf{40.6}
          & \cellcolor[rgb]{ .867,  .922,  .969}\textbf{47.7} \\
        5 & 0.8 & 0.5 & $[0.1,0.5]$ & 2 & 66.2 & 40.2 & 47.3 \\
        \midrule
        6 & 0.6 & 0.3 & $[0.1,0.5]$ & 2 & 65.9 & 39.3 & 47.3 \\
        7 & 0.6 & 0.7 & $[0.1,0.5]$ & 2 & 65.9 & 40.3 & 47.2 \\
        8 & 0.6 & 0.5 & $[0.2,0.6]$ & 2 & 66.1 & 39.4 & 46.9 \\
        9 & 0.6 & 0.5 & $[0.3,0.7]$ & 2 & 64.6 & 39.8 & 46.9  \\
        \midrule
        10 & 0.6 & 0.5 & $[0.1,0.5]$ & 1 & 65.2 & 39.2 & 47.2 \\
        11 & 0.6 & 0.5 & $[0.1,0.5]$ & 4 & 65.7 & 39.7 & 47.2 \\
        \bottomrule
    \end{tabular}%
    }
\vspace{-10pt}
\end{table}

\subsection{Ablation Study}
\label{subsec:ablation_study}

We conduct ablation studies to analyze the key design choices and pre-training hyperparameters of \methodname{}. All experiments use a ViT-B/16 backbone pre-trained on Kinetics-400 for 200 epochs and evaluate performance on three benchmarks: DAVIS, VIP, and JHMDB. The results are reported in \Tbref{tab:ablation} and \Tbref{tab:ablation_training} with default settings in blue.

\textbf{Effect of temporal modeling components.} 
\textit{Lines 1-4} in \Tbref{tab:ablation} progressively introduce the main components of \methodname{}. 
Compared with the baseline in \textit{line 1}, introducing temporal variation modeling with multiple trajectory parameter groups in \textit{line 2} improves the results by explicitly constraining representation evolution across frames.
High-confidence patch registration in \textit{line 3} brings further gains by applying this constraint to more reliable cross-frame correspondences.
The full setting in \textit{line 4}, which further introduces a dedicated projection space for temporal variation modeling, achieves overall improvements of $1.7\%$, $1.2\%$, and $0.9\%$ on DAVIS, VIP, and JHMDB, respectively.
These results show that temporal variation modeling is most effective when applied to reliable cross-frame correspondences and optimized in a space tailored for latent-space constraints.

\textbf{Effect of trajectory parameter groups.} 
\textit{Lines 4-7} in \Tbref{tab:ablation} study the number of trajectory parameter groups $K$. Using only one group in \textit{line 5} forces different videos to share a single temporal evolution pattern, which can be restrictive for diverse video dynamics.
Increasing $K$ provides more flexibility, but overly large values do not bring further improvement, as shown in \textit{lines 6-7}. 
The default setting $K=3$ improves the results by at least $0.8\%$, $0.8\%$, and $0.4\%$ over the other settings in this group on DAVIS, VIP, and JHMDB. 
This suggests that a small number of groups is sufficient to model common temporal variation patterns, while excessive grouping may weaken the regularization effect and reduce optimization stability.


\textbf{Effect of retained patch ratio.}
\textit{Lines 4 and 8-10} in \Tbref{tab:ablation} evaluate the retained patch ratio $\rho$.
A small $\rho$ limits the number of patch trajectories used for temporal variation modeling, while a large $\rho$ includes more uncertain matches and may introduce noise into trajectory estimation.
The default setting $\rho=0.6$ improves DAVIS performance by at least $1.2$ percentage points over the other ratio settings.
These results support retaining a moderate proportion of patch positions to balance the coverage and reliability of temporal supervision.

\textbf{Effect of projection dimension.} 
\textit{Lines 4 and 11-13} in \Tbref{tab:ablation} analyze the dimension $d'$ of the temporal projection space. 
Temporal variation is relatively compact compared with the original dense representation and thus benefits from appropriate information compression. 
When the projection dimension is too small, the temporal space may lose useful semantic information. 
However, increasing the dimension to 512 in \textit{line 13} clearly degrades performance, indicating that insufficient compression may introduce noise and task-irrelevant variations from dense representations into the temporal space, thereby weakening the auxiliary constraint.
The default setting $d'=256$ improves the results by at least $1.0\%$ over the other projection dimensions on DAVIS. 
This suggests that a compact yet expressive projection space is more suitable for modeling temporal variation.

\textbf{Effect of pre-training hyperparameters.}
\textit{Lines 1-5} in \Tbref{tab:ablation_training} show that introducing $\mathcal{L}_{\mathrm{var}}$ at $\xi_s=0.6$ improves performance over activation from the start by $2.1$, $1.9$, and $1.6$ percentage points on DAVIS, VIP, and JHMDB, respectively.
Early activation may constrain representations and correspondences that are not yet sufficiently reliable, while overly late activation may leave insufficient training time for adaptation to temporal variation.
These results support introducing the constraint after initial representation learning while retaining sufficient training time for temporal variation modeling.
\textit{Lines 4 and 6-9} support moderate masking to balance reconstruction difficulty and reliable correspondence learning.
Insufficient masking encourages reliance on visible regions within the current frame, while excessive masking reduces contextual cues for PRM and may increase mask-induced perturbations in trajectory estimation.
\textit{Lines 4 and 10-11} show that both lower and higher sampling rates underperform the default 2 FPS.
This suggests that moderate sampling better preserves useful temporal cues while reducing redundant observations, providing effective two-sided context for temporal modeling.

\subsection{Analysis of Video Representations}
\label{subsec:representation_analysis}

We further analyze the learned video representations from both quantitative and qualitative perspectives.
We first report quantitative metrics to evaluate temporal correspondence, representation diversity, and semantic separability.
We then visualize the cross-frame attention heatmaps in PRM to examine whether reliable patch-level temporal correspondences are established.
Finally, we present representation-space trajectories to inspect the temporal organization of frame-level representations.

\subsubsection{Quantitative Representation Analysis}
\label{subsubsection:quantitative_analysis}

We first provide a compact quantitative analysis of the learned video representations.
Detailed metric definitions and implementation protocols are provided in the \cref{app:additional_analysis}.

\noindent\textbf{Experiment setup.}
To provide an interpretable assessment of representation quality, we analyze dense video features from static and dynamic perspectives, with all metrics computed directly in the original representation space.
For all methods, we use the same sampled videos and frame indices and extract patch tokens after removing the \texttt{[CLS]} token and register tokens.
\textbf{1) Static perspective.}
Cycle consistency accuracy measures the proportion of patches returning to their original positions along a three-frame palindrome trajectory, reflecting correspondence reliability.
Effective rank~\cite{erank} measures the normalized effective dimensionality of token features from their singular-value distribution, reflecting representation diversity.
Separation ratio compares inter-video and intra-video feature distances, reflecting video-level separability.
\textbf{2) Dynamic perspective.}
Temporal proximity measures the distance gap between nonadjacent and adjacent frame representations, reflecting temporal locality.
Local coherence measures the temporal correlation between consecutive frame representations, reflecting the persistence of local feature states.
Trajectory smoothness measures trajectory curvature through turning angles across temporal intervals.

\begin{figure}[t]
    \centering
    \begin{subfigure}{1.0\linewidth}
        \centering
        \includegraphics[width=0.98\linewidth]{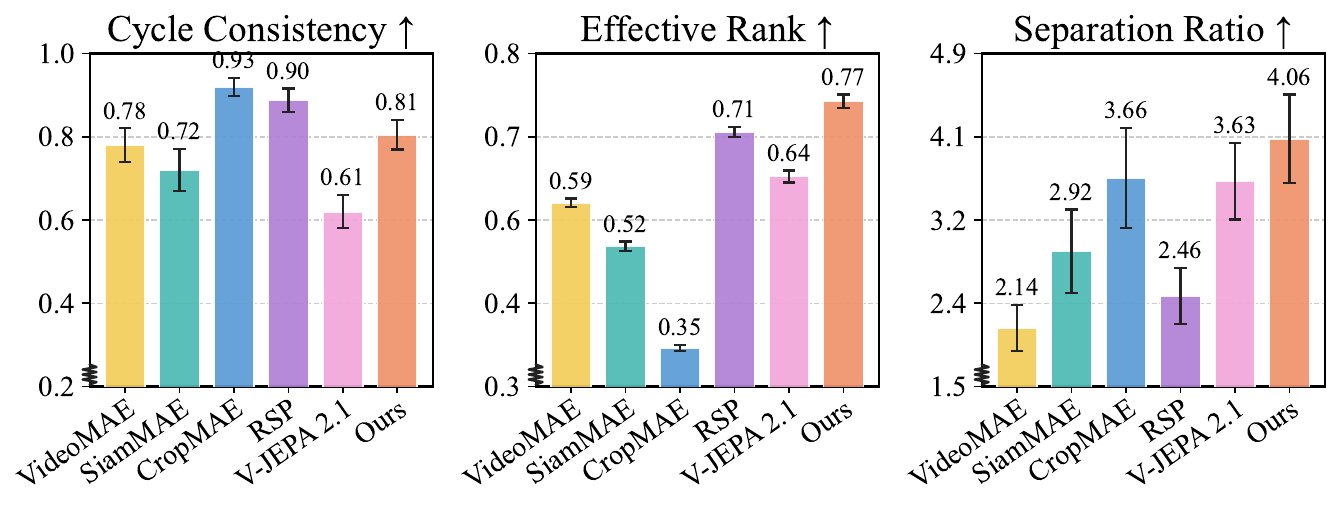}
        \label{fig:paper_bars_davis}
    \end{subfigure}
    \par
    \begin{subfigure}{1.0\linewidth}
        \centering
        \includegraphics[width=0.98\linewidth]{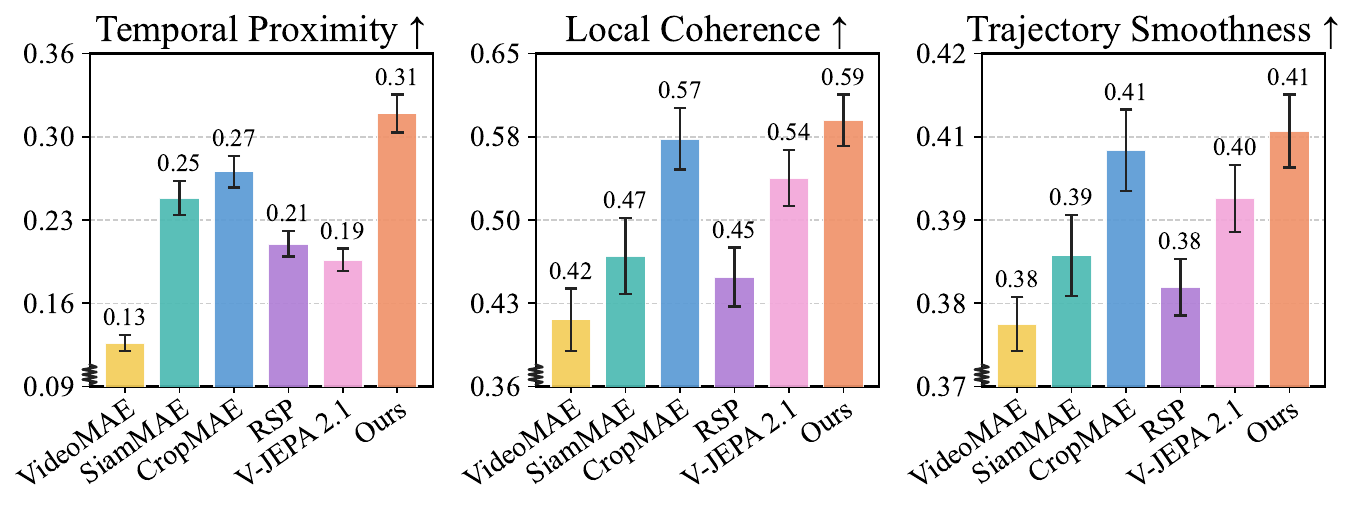}
        \label{fig:paper_bars_davis_dynamics}
    \end{subfigure}
    \caption{Quantitative comparison of representation quality on DAVIS from static (top) and dynamic (bottom) perspectives. Higher values are preferred for all metrics.}
    \label{fig:metrics}
    \vspace{-8pt}
\end{figure}

\noindent\textbf{Evaluation results.}
\Fgref{fig:metrics} presents the quantitative comparison on DAVIS.
\textbf{1) Static perspective.}
\methodname{} achieves competitive cycle-consistency accuracy, indicating reliable short-range temporal correspondence consistent with the explicit patch-level alignment provided by PRM.
Its highest effective rank indicates less concentration of dense tokens in a few dominant dimensions, suggesting that diversity is retained under the temporal variation constraint.
The best separation ratio indicates stronger inter-video separation relative to intra-video variation, suggesting that modeling frame-wise changes preserves video-level separability.
\textbf{2) Dynamic perspective.}
\methodname{} achieves the best results on all three measurements.
Higher temporal proximity indicates that adjacent-frame representations are closer relative to nonadjacent ones, while higher local coherence indicates stronger temporal correlation between consecutive feature states.
Higher trajectory smoothness further indicates smaller average turning angles along the representation trajectories.
The explicit second-order constraint links aligned feature states with their temporal changes, encouraging coherent local evolution while retaining temporal variation.
These results are consistent with this regularization improving temporal organization, complementing the reliable patch correspondence provided by PRM.

Overall, \methodname{} maintains a well-organized static representation structure while learning more coherent and smoother dynamic trajectories.


\begin{figure*}[t]
  \centering
    \includegraphics[width=0.96\linewidth]{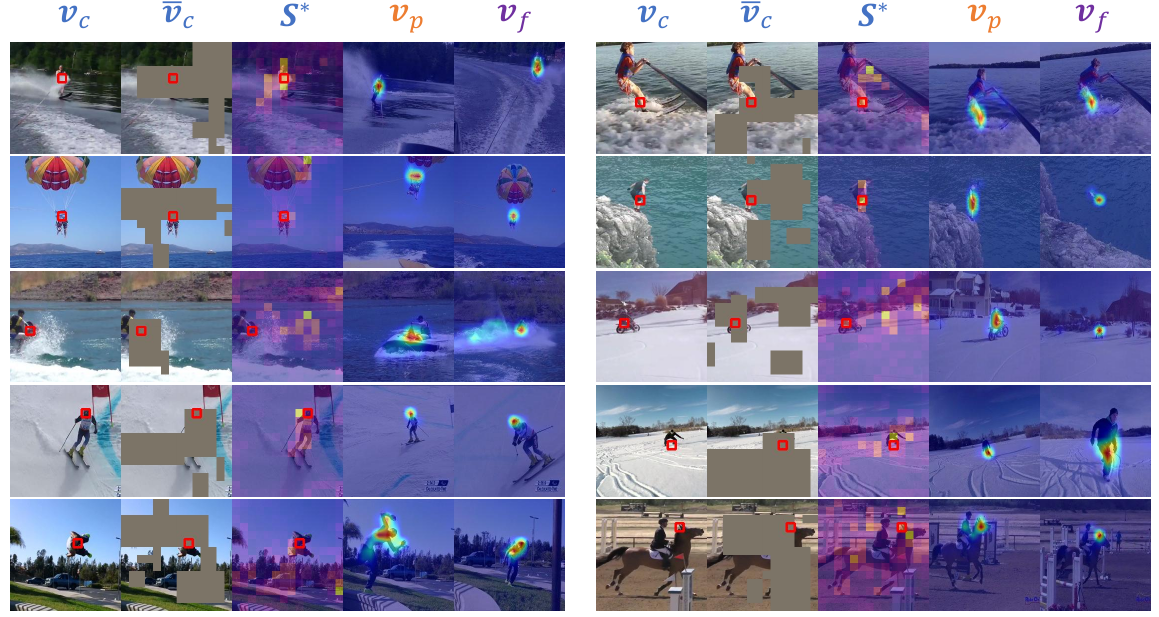}
    \vspace{-3pt}
    \caption{Cross-attention and confidence visualization in PRM. We visualize the current frame $\bm v_c$, the masked current frame $\bar{\bm v}_c$, the joint confidence map $\bm S^{\ast}$, and the cross-attention maps from masked patches to the past and future frames $\bm v_p$ and $\bm v_f$.}
    \label{fig:vis_heatmaps}
    \vspace{-5pt}
\end{figure*}

\begin{figure*}[t]
  \centering
    \includegraphics[width=0.96\linewidth]{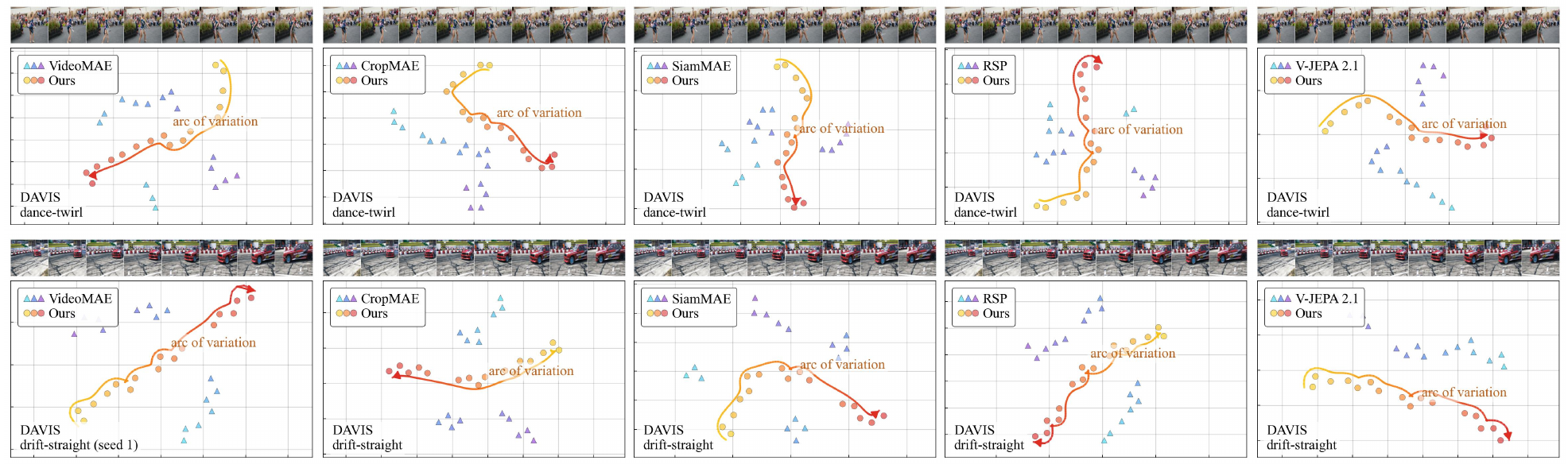}
    \vspace{-3pt}
    \caption{t-SNE visualization of representation trajectories for videos. Each row shows the same video across different methods, with temporal order indicated by color gradients and arrows. Best viewed in color.}
    \label{fig:vis_arc}
    \vspace{-10pt}
\end{figure*}

\subsubsection{Visualization of Cross-attention Maps in PRM}
\label{subsubsection:map_PRM}
In \Fgref{fig:vis_heatmaps}, we visualize the cross-attention maps between the masked patches in the current frame and the auxiliary frames in PRM to examine whether reliable temporal correspondence is established. 
As shown in the second, fourth, and fifth columns, even when the queried patches are masked in the current frame, PRM can still attend to semantically corresponding regions in the auxiliary frames. 
The joint confidence map $\bm{S}^{\ast}$ in the third column further shows that high-confidence responses are mainly concentrated on foreground objects and motion-related regions, where cross-frame correspondences are more reliable. 
These results indicate that the auxiliary branch establishes reliable patch-level temporal correspondence for latent-space reconstruction.


\subsubsection{Representation Trajectories Comparison}
\label{subsubsection:trajectories}
In \Fgref{fig:vis_arc}, we use t-SNE to provide an intuitive low-dimensional visualization of the distribution and temporal organization of frame-level representations.
Compared with previous methods, the visualizations show smoother and more continuous trajectories for \methodname{}.
These visual observations complement the quantitative measurements of temporal proximity, local coherence, and trajectory smoothness in the original representation space, suggesting that \methodname{} learns more ordered visual-state trajectories.

\vspace{-6pt}

%% file: sec/5_conclusion.tex
\section{Conclusion and Future Work}
\label{sec:conclusion}

In this work, we propose \methodname{}, a self-supervised framework for video representation learning with variation-aware temporal correspondence.
The framework introduces a sandwich sampling strategy to provide two-sided temporal references and proposes the Patch Registration Module (PRM) to establish reliable dense correspondence for masked representation reconstruction.
To model semantic consistency and temporal changes, \methodname{} performs latent-space self-distillation on registered representations, derives a finite-difference residual from a second-order Taylor expansion, and regularizes the trajectories of corresponding representations in a separate projected space.
Experiments on pixel-level, frame-level, video-level, and vision-based robotic learning tasks demonstrate the effectiveness and applicability of \methodname{} across diverse video understanding scenarios.
Representation analysis further shows that \methodname{} learns a more structured representation space with reliable temporal correspondence, stronger token-level diversity, and favorable semantic separability.

Future work may extend the current local three-frame constraint to multi-point temporal constraints for more accurate trajectory approximation and further investigate object-level representation evolution through semantically consistent region aggregation.



%% file: sec/bio.tex
\begin{IEEEbiography}
	[{\includegraphics[width=1in,height=1.25in,clip,keepaspectratio]{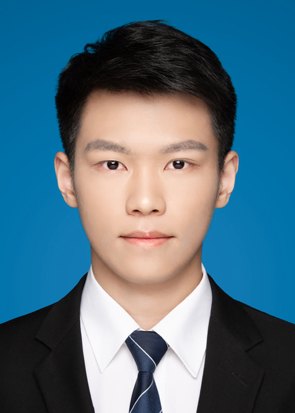}}]
  {Yang Liu} received the B.S. degree from Harbin Institute of Technology at Weihai in 2023. He is currently pursuing the Ph.D. degree in the School of Computer Science and Technology, University of Chinese Academy of Sciences. His current research interests include self-supervised learning and video understanding. He has authored or coauthored 6 academic papers in major international journals and conferences (including NeurIPS, CVPR, ACM Multimedia, etc).
\end{IEEEbiography}

\begin{IEEEbiography}
	[{\includegraphics[width=1in,height=1.25in,clip,keepaspectratio]{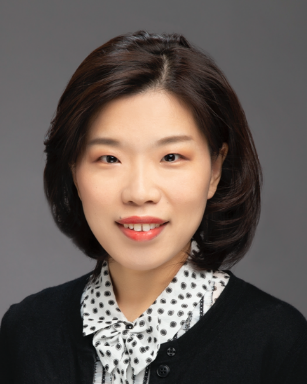}}]{Qianqian Xu} received the B.S. degree in computer science from China University of Mining and Technology in 2007 and the Ph.D. degree in computer science from University of Chinese Academy of Sciences in 2013. She is currently a Professor with the Institute of Computing Technology, Chinese Academy of Sciences, Beijing, China. Her research interests include statistical machine learning, with applications in multimedia and computer vision. She has authored or coauthored 100+ academic papers in prestigious international journals and conferences (including T-PAMI, IJCV, T-IP, NeurIPS, ICML, CVPR, AAAI, etc). Moreover, she serves as an associate editor of IEEE Transactions on Circuits and Systems for Video Technology, IEEE Transactions on Multimedia, and ACM Transactions on Multimedia Computing, Communications, and Applications.
\end{IEEEbiography}

\begin{IEEEbiography}
	[{\includegraphics[width=1in,height=1.25in,clip,keepaspectratio]{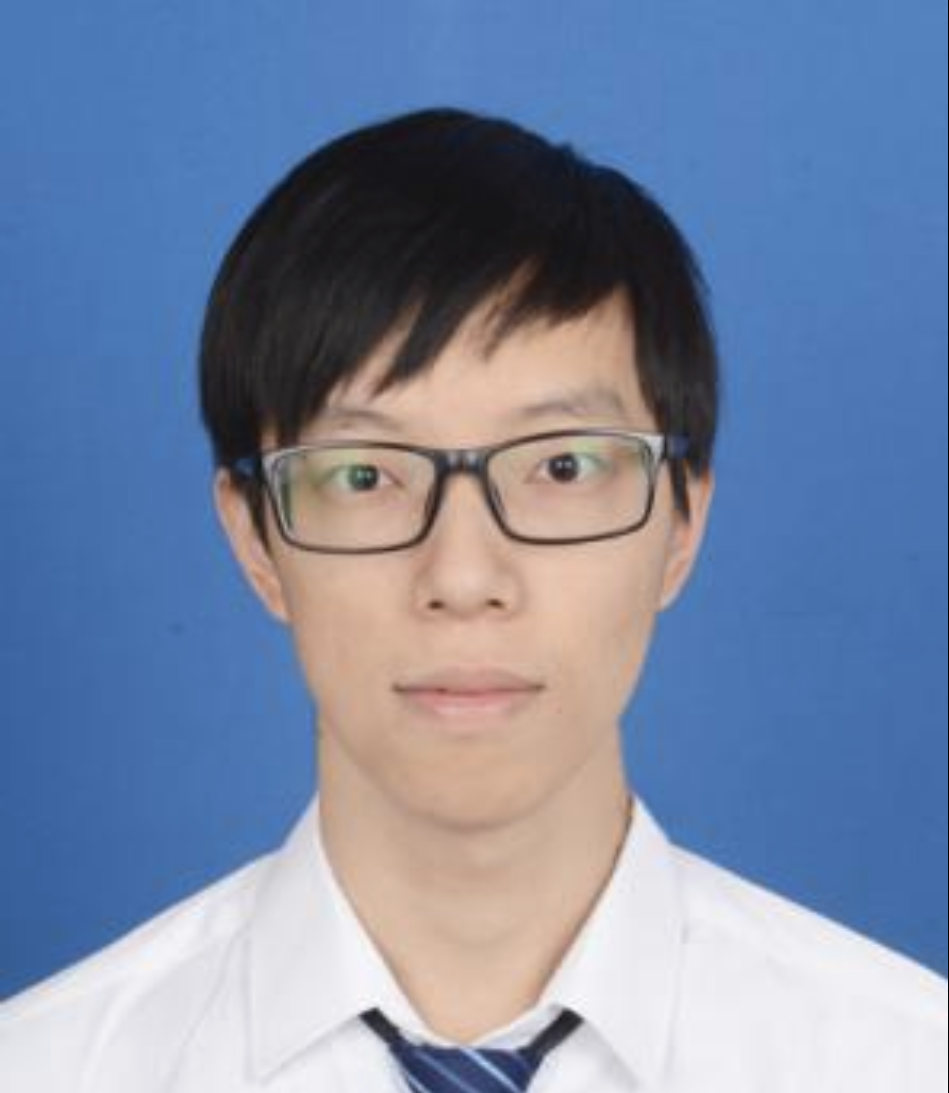}}]
  {Peisong Wen} received the B.S. degree in intelligent science and technology from Nankai University in 2020 and the Ph.D. degree in computer science from the University of Chinese Academy of Sciences in 2025. He is currently a postdoctoral research fellow with the University of Chinese Academy of Sciences. His research interests lie in machine learning and computer vision. He has authored or coauthored 20 academic papers in top-tier international journals and conferences (including T-PAMI, IJCV, NeurIPS, ICML, etc).
\end{IEEEbiography}

\begin{IEEEbiography}
    [{\includegraphics[width=1in,height=1.25in,clip,keepaspectratio]{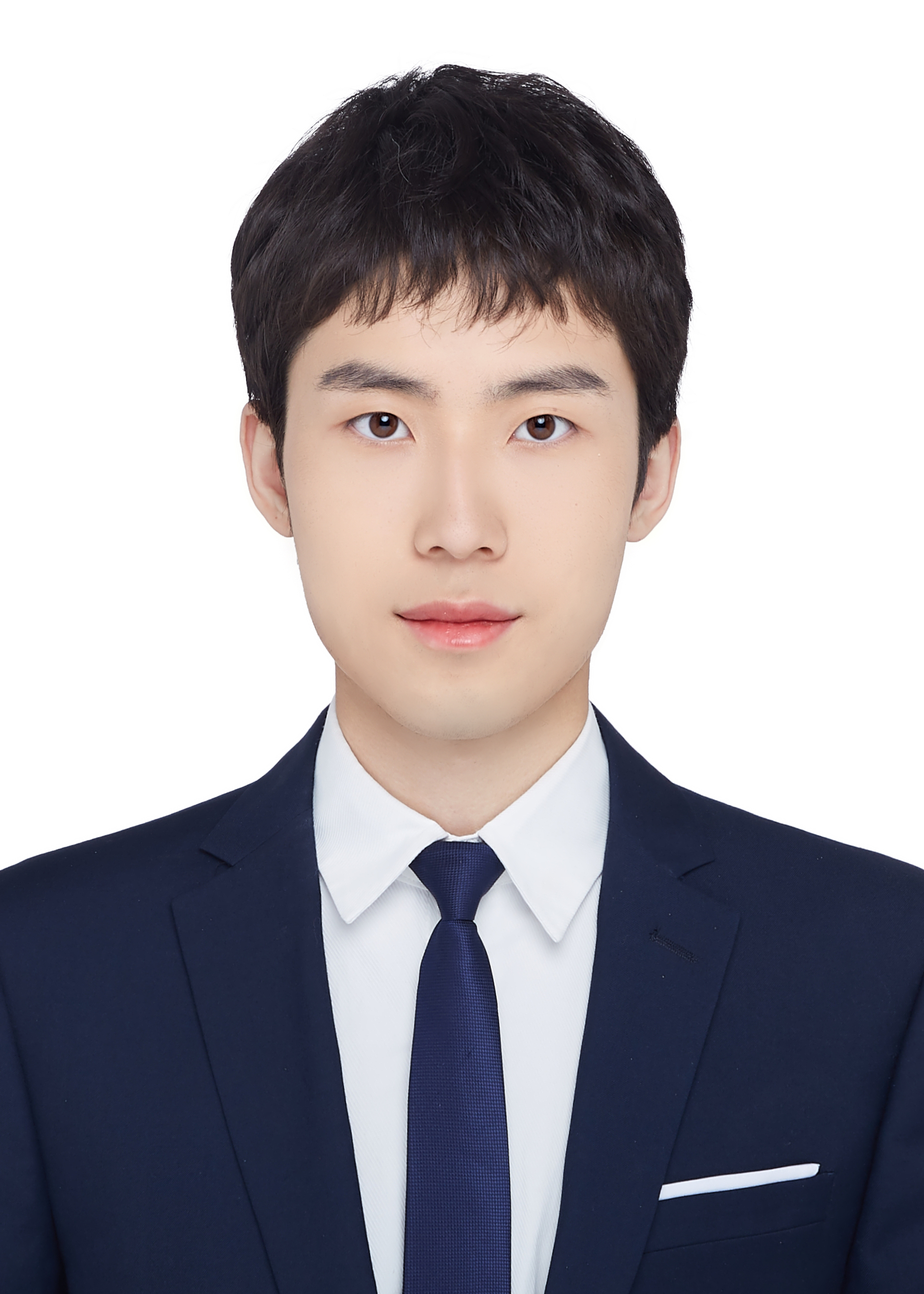}}]
  {Siran Dai} received the B.S. degree in robotic engineering from Beijing University of Technology in 2022. He is currently a Ph.D. student at the University of Chinese Academy of Sciences. His research interests include machine learning, self-supervised learning, and information theory. He has authored or coauthored 9 academic papers in major international journals and conferences (including T-PAMI, NeurIPS, CVPR, AAAI, ACM Multimedia, etc).
\end{IEEEbiography}

\begin{IEEEbiography}
	[{\includegraphics[width=1in,height=1.25in,clip,keepaspectratio]{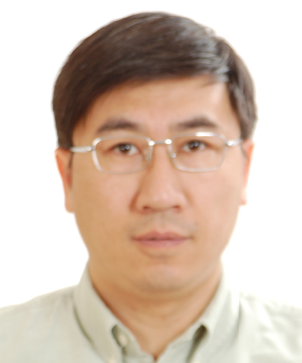}}]{Qingming Huang} is a chair professor in the University of Chinese Academy of Sciences and an adjunct research professor in the Institute of Computing Technology, Chinese Academy of Sciences. He graduated with a Bachelor degree in Computer Science in 1988 and Ph.D. degree in Computer Engineering in 1994, both from Harbin Institute of Technology, China. His research areas include multimedia computing, image processing, computer vision and pattern recognition. He has authored or coauthored more than 400 academic papers in prestigious international journals and top-level international conferences. He was the associate editor of IEEE Trans. on CSVT and Acta Automatica Sinica, and the reviewer of various international journals including IEEE Trans. on PAMI, IEEE Trans. on Image Processing, IEEE Trans. on Multimedia, etc. He is a Fellow of IEEE and has served as general chair, program chair, area chair and TPC member for various conferences, including ACM Multimedia, CVPR, ICCV, ICME, ICMR, PCM, BigMM, PSIVT, etc.
\end{IEEEbiography}

%% file: sec/X_appendix.tex
\onecolumn
\appendices

\section*{\textcolor{blue}{\Large Contents}}
\startcontents[appendices]

\begingroup
\renewcommand{\baselinestretch}{1.25}\selectfont
\setlength{\parskip}{2pt}
\printcontents[appendices]{l}{1}{\setcounter{tocdepth}{3}}
\endgroup

\newpage

\section{Supplementary Explanation of Method}
\label{sec:method_app}

\subsection{Comparison with Previous Masked Modeling Structures}
\label{subsec:diff_previous}

This subsection focuses on the structural comparison of the masked-modeling component in \methodname{} with previous MVM designs.

\begin{figure*}[htbp]
  \centering
  \begin{subfigure}{0.33\linewidth}
    \includegraphics[width=1.0\linewidth]{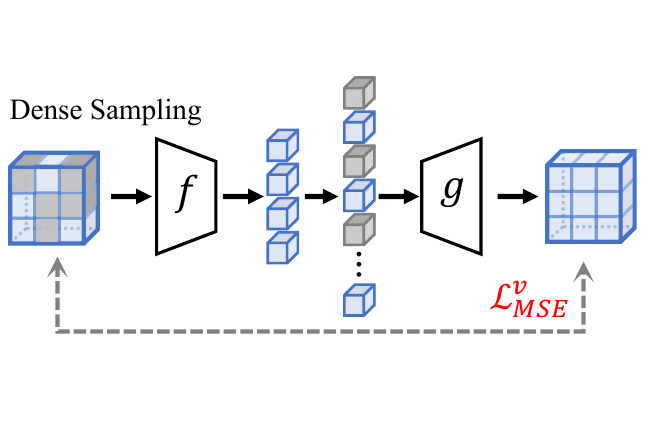}
    \caption{Structure of early MVM methods.}
    \label{fig:compare_methods_1}
    \end{subfigure}
  \begin{subfigure}{0.33\linewidth}
    \includegraphics[width=1.0\linewidth]{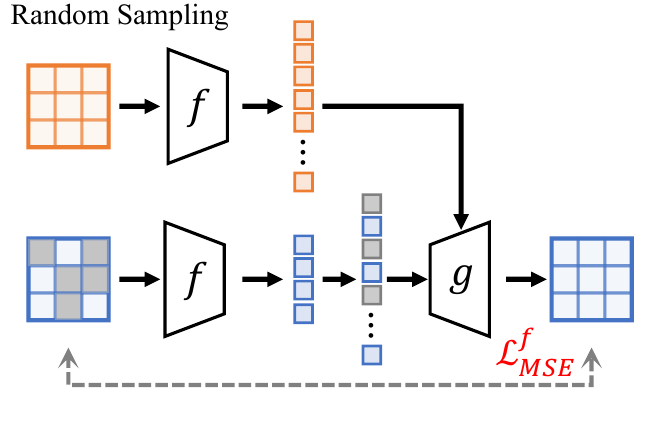}
    \caption{Structure of recent MVM methods.}
    \label{fig:compare_methods_2}
  \end{subfigure}
  \begin{subfigure}{0.33\linewidth}
    \includegraphics[width=1.0\linewidth]{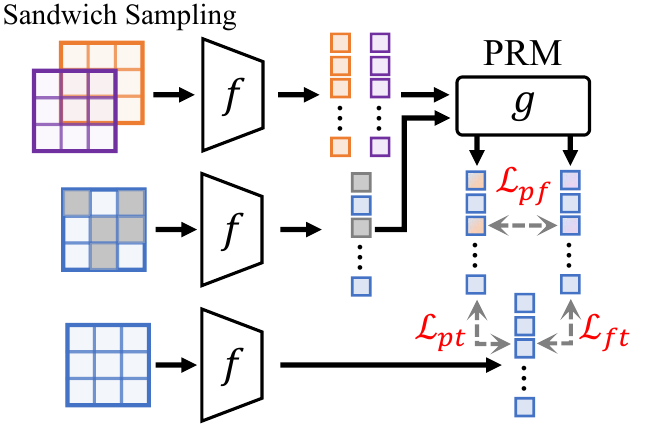}
    \caption{Structure of MVM in \methodname{} (Ours).}
    \label{fig:compare_methods_3}
  \end{subfigure}
    \caption{Comparison of structures between our framework and previous MVM methods.}
    \label{fig:compare_methods}
\end{figure*}

The success of Masked Image Modeling (MIM) has inspired its extension to the video domain as Masked Video Modeling (MVM), which has demonstrated impressive performance across multiple video tasks~\cite{DAVIS17,Kinetics,SSV2,HMDB51,UCF101} and provided effective representations for visual generation~\cite{ad3,ad4}.
Given a video $\bm{V} = \{\bm{v}_j \in \mathbb{R}^{H \times W \times C}\}_{j=1}^{T}$, early MVM methods~\cite{MAE_ST,VideoMAE,VideoMAEv2,DropMAE,OmniMAE} aim to train the video encoder $f$ by densely sampling video clips and forcing the model to recover masked pixels. Specifically, as shown in \Fgref{fig:compare_methods_1}, for a sampled video clip $\bm{V}' \subset \bm{V}$, $\bm{V}'$ is divided into spatiotemporal tubes, a large proportion of which are masked and reconstructed. Next, the masked clip is passed through an encoder-decoder structure to predict the original clip $\widehat{\bm{V}'}$ by recovering the masked pixels. Finally, the mean squared error (MSE) loss is utilized to minimize the difference between $\widehat{\bm{V}'}$ and $\bm{V}'$:
\begin{equation}
    \mathcal{L}_{MSE}^v = \frac{1}{|\mathcal{M}|}\sum_{i \in \mathcal{M}} \| (\bm{V}')^i - (\widehat{\bm{V}'})^i \|_2^2 ,
\label{eq:mse_loss_1}
\end{equation}
where $\mathcal{M}$ is the set of \texttt{[MASK]} tokens.

Recent MVM methods~\cite{SiamMAE,CropMAE,RSP} have shifted to a random sampling strategy to reduce computational cost. Typically, as shown in \Fgref{fig:compare_methods_2}, these methods first sample a current frame $\bm{v}_c \in \bm{V}$ with several masked patches, then sample an unmasked past frame $\bm{v}_p \in \bm{V}$. The masked current frame $\bm{v}_c$ is restored to $\widehat{\bm{v}}_c$ using $\bm{v}_p$ as prior information based on a conditional decoder. Similarly, the mean squared error (MSE) loss is used to minimize the difference between $\widehat{\bm{v}}_c$ and $\bm{v}_c$:
\begin{equation}
    \mathcal{L}_{MSE}^f = \frac{1}{|\mathcal{M}|}\sum_{i \in \mathcal{M}} \| (\bm{v}_c)^i - (\widehat{\bm{v}}_c)^i \|_2^2 .
\label{eq:mse_loss_2}
\end{equation}

As discussed in \cref{sec:introduction}, the random sampling strategy introduces uncertainty in reconstruction since a single conditional frame can lead to multiple potential predictions. Besides, previous MVM methods primarily restore the masked regions at the pixel level, making the model retain excessive low-level information. To address these issues, as shown in \Fgref{fig:compare_methods_3}, we propose \methodname{} with two key properties for MVM: \textbf{1)} a sandwich sampling strategy to establish temporal correspondence from auxiliary frames, thereby reducing the reconstruction uncertainty; \textbf{2)} an auxiliary branch on top of a self-distillation structure to reconstruct the masked patches in the latent space, enabling the model to capture higher-level semantic representations.
Following the formulation in Sec.~3, we employ CE loss for representation reconstruction and MSE loss for temporal squeezing:

\begin{equation}
    \begin{aligned}
        \lpt &= -\frac{1}{|\mathcal M(\bm v_c)|}\sum_{i \in \mathcal{M}(\bm v_c)} (\hat {\bm q}_c)^i \log(\bm q_c^p)^i  , \\
        \lft &= -\frac{1}{|\mathcal M(\bm v_c)|}\sum_{i \in \mathcal{M}(\bm v_c)} (\hat {\bm q}_c)^i \log (\bm q_c^f)^i ,
    \label{eq:loss1}
    \end{aligned}
\end{equation}
\begin{equation}
    \lpf = \frac{1}{|\mathcal M(\bm v_c)|}\sum_{i \in \mathcal M(\bm v_c)} \| (\bm q_c^p)^i - (\bm q_c^f)^i \|_2^2 ,
\label{eq:loss2}
\end{equation}
where $\bm q_c^p$ and $\bm q_c^f$ are the restored representations of the current frame guided by the past and future frames, respectively, and $\hat{\bm q}_c$ is the representation from the teacher branch.

\subsection{Detailed Derivation of Variation-aware Temporal Modeling}
\label{subsec:var_derivation_supp}

This section provides the detailed derivation of the variation-aware temporal modeling objective.
After patch registration, the past and future representations are aligned to the current-frame coordinate, forming registered triplets with the masked current representation.
These registered triplets provide reliable local temporal correspondences for modeling how corresponding representations evolve over time.
Since visual targets usually change continuously within short temporal intervals, their corresponding latent representations are expected to form locally smooth trajectories rather than independent observations.
Based on this motivation, we derive a second-order local trajectory relation from Taylor expansion and instantiate it as a differentiable self-supervised objective in a projected temporal space.

\subsubsection{Representation Trajectory Modeling}
\label{subsubsec:trajectories_estimation_supp}

Let ${\zcsp}$, ${\zcsf}$, and $\bar{\bm z}_c$ denote the registered past representation, registered future representation, and masked current representation, respectively.
Following the main paper, we first map these representations into a separate space to separate semantic alignment and temporal variation modeling:
\begin{equation}
    \bm p_c^{p} = \phi({\zcsp}), \quad
    \bar{\bm p}_c = \phi(\bar{\bm z}_c), \quad
    \bm p_c^{f} = \phi({\zcsf}),
    \label{eq:temporal_projection_supp}
\end{equation}
where $\phi:\mathbb{R}^{d}\rightarrow\mathbb{R}^{d'}$ is implemented as a linear projection followed by layer normalization.
Here, $\bm p_c^{p},\bar{\bm p}_c,\bm p_c^{f}\in\mathbb{R}^{N\times d'}$, where $N$ denotes the number of patch tokens and $d'$ denotes the dimension of the projected temporal space.

For a locally smooth temporal process, Taylor expansion provides a standard approximation of the state around a reference time.
Given a reference state $\bm P(t_0)$, a nearby state $\bm P(t)$ can be approximated as
\begin{equation}
    \bm P(t)
    =
    \bm P(t_0)
    +
    (t-t_0)\frac{\partial \bm P}{\partial t}(t_0)
    +
    \frac{(t-t_0)^2}{2}
    \frac{\partial^2 \bm P}{\partial t^2}(t_0)
    +
    \bm R_3(t),
    \label{eq:taylor_general_supp}
\end{equation}
where $\bm R_3(t)$ denotes the third-order remainder terms.
In our setting, each registered patch position provides three ordered observations of local latent evolution.
Under local temporal sampling, we neglect the higher-order term by setting $\bm R_3(t)\approx 0$, and approximate the first- and second-order temporal variations at the current timestamp by finite differences.

Let $t_p=t_c-\Delta_p$ and $t_f=t_c+\Delta_f$ denote the past and future timestamps, where $\Delta_p>0$ and $\Delta_f>0$ are the temporal intervals.
The backward and forward finite differences around $t_c$ are defined as
\begin{equation}
    \bm v_p
    =
    \frac{\bar{\bm p}_c-\bm p_c^p}{\Delta_p},
    \quad
    \bm v_f
    =
    \frac{\bm p_c^f-\bar{\bm p}_c}{\Delta_f}.
    \label{eq:backward_forward_difference_supp}
\end{equation}
Since the two temporal intervals can be different, the first-order temporal derivative at $t_c$ is estimated by a weighted combination of the two finite differences:
\begin{equation}
    \widehat{\frac{\partial \bm p}{\partial t}}(t_c)
    =
    \frac{\Delta_p}{\Delta_p+\Delta_f}\bm v_f
    +
    \frac{\Delta_f}{\Delta_p+\Delta_f}\bm v_p .
    \label{eq:first_order_weighted_supp}
\end{equation}
Substituting \Eqref{eq:backward_forward_difference_supp} into \Eqref{eq:first_order_weighted_supp} gives
\begin{equation}
    \begin{aligned}
    \widehat{\frac{\partial \bm p}{\partial t}}(t_c)
    &=
    \frac{\Delta_p}{\Delta_p+\Delta_f}
    \frac{\bm p_c^f-\bar{\bm p}_c}{\Delta_f}
    +
    \frac{\Delta_f}{\Delta_p+\Delta_f}
    \frac{\bar{\bm p}_c-\bm p_c^p}{\Delta_p}
    \\
    &=
    \frac{
    \Delta_p^2 \bm p_c^f
    -
    \Delta_f^2 \bm p_c^p
    +
    (\Delta_f^2-\Delta_p^2)\bar{\bm p}_c
    }
    {\Delta_p\Delta_f(\Delta_p+\Delta_f)} .
    \end{aligned}
    \label{eq:first_order_difference_supp}
\end{equation}
The second-order temporal derivative is estimated by the difference between the future-side and past-side finite differences, normalized by the distance between their temporal centers:
\begin{equation}
    \begin{aligned}
    \widehat{\frac{\partial^2 \bm p}{\partial t^2}}(t_c)
    &=
    \frac{2}{\Delta_p+\Delta_f}
    \left(
    \frac{\bm p_c^f-\bar{\bm p}_c}{\Delta_f}
    -
    \frac{\bar{\bm p}_c-\bm p_c^p}{\Delta_p}
    \right)
    \\
    &=
    \frac{
    2\Delta_f \bm p_c^p
    -
    2(\Delta_p+\Delta_f)\bar{\bm p}_c
    +
    2\Delta_p \bm p_c^f
    }
    {\Delta_p\Delta_f(\Delta_p+\Delta_f)} .
    \end{aligned}
    \label{eq:second_order_difference_supp}
\end{equation}
When $\Delta_p=\Delta_f=\Delta$, these estimates reduce to the standard centered finite-difference forms:
\begin{equation}
    \widehat{\frac{\partial \bm p}{\partial t}}(t_c)
    =
    \frac{\bm p_c^f-\bm p_c^p}{2\Delta},
    \quad
    \widehat{\frac{\partial^2 \bm p}{\partial t^2}}(t_c)
    =
    \frac{\bm p_c^p-2\bar{\bm p}_c+\bm p_c^f}{\Delta^2}.
    \label{eq:uniform_difference_supp}
\end{equation}

On top of these finite-difference estimates, we further approximate the registered past and future states with the second-order Taylor expansion around $t_c$:
\begin{equation}
    \bm p_c^{p}
    =
    \bar{\bm p}_c
    -
    \Delta_p
    \widehat{\frac{\partial \bm p}{\partial t}}(t_c)
    +
    \frac{\Delta_p^2}{2}
    \widehat{\frac{\partial^2 \bm p}{\partial t^2}}(t_c),
    \label{eq:taylor_past_supp}
\end{equation}
\begin{equation}
    \bm p_c^{f}
    =
    \bar{\bm p}_c
    +
    \Delta_f
    \widehat{\frac{\partial \bm p}{\partial t}}(t_c)
    +
    \frac{\Delta_f^2}{2}
    \widehat{\frac{\partial^2 \bm p}{\partial t^2}}(t_c).
    \label{eq:taylor_future_supp}
\end{equation}
These two approximations motivate a second-order local relation among the registered triplet:
\begin{equation}
    \widehat{\frac{\partial^2 \bm p}{\partial t^2}}(t_c)
    +
    \gamma_1
    \widehat{\frac{\partial \bm p}{\partial t}}(t_c)
    +
    \gamma_0
    \bm p^{\delta}
    =
    \bm 0,
    \label{eq:taylor_ode_relation_supp}
\end{equation}
where
\begin{equation}
    \bm p^{\delta}
    =
    \bar{\bm p}_c
    -
    \frac{\bm p_c^p+\bm p_c^f}{2}
    \label{eq:pdelta_supp}
\end{equation}
denotes the deviation of the current state from the two-sided average and serves as the zeroth-order term in this relation.
The coefficients $\gamma_1$ and $\gamma_0$ characterize the local trajectory relation and are related to the temporal intervals under the Taylor approximation.
However, due to approximation errors, a single registered triplet provides limited evidence for learning a common temporal-evolution pattern.

To obtain a more reliable constraint, we jointly model local trajectory relations across multiple observations.
Given $B$ observations with similar local variation patterns,
$\{(\widehat{\frac{\partial^2 \bm p^{(i)}}{\partial t^2}}, \widehat{\frac{\partial \bm p^{(i)}}{\partial t}}, \bm p^{\delta,(i)})\}_{i=1}^B$,
we assume that they share the same parameters $(\bar{\gamma}_0,\bar{\gamma}_1)$.
For each patch position and feature coordinate, the corresponding entries of the zeroth-order term $\bm p^{\delta,(i)}$, the first-order term $\widehat{\frac{\partial \bm p^{(i)}}{\partial t}}(t_c^{(i)})$, and the second-order term $\widehat{\frac{\partial^2 \bm p^{(i)}}{\partial t^2}}(t_c^{(i)})$ form the three coordinates of a point.
The shared coefficients constrain these points from different observations to remain close to a common plane with normal vector $(\bar{\gamma}_0,\bar{\gamma}_1,1)^\top$.
The shared relation characterizes a common local evolution pattern, which can be interpreted as a latent temporal-evolution prototype.

To further interpret this relation, consider a fixed patch position and feature coordinate.
Let $x^{(i)}(\tau)$ denote the continuous local representation coordinate centered at the two-sided reference of triplet $i$, where $\tau=t-t_c^{(i)}$ and the reference is held fixed within this local model.
At $\tau=0$, $x^{(i)}(0)$ corresponds to the selected entry of $\bm p^{\delta,(i)}$.
Centering leaves the first- and second-order derivatives unchanged, and their values at $\tau=0$ are approximated by the corresponding finite-difference estimates.
For a given shared coefficient pair, the associated local constant-coefficient model is
\begin{equation}
    \frac{\partial^2 x^{(i)}}{\partial \tau^2}
    +
    \bar{\gamma}_1
    \frac{\partial x^{(i)}}{\partial \tau}
    +
    \bar{\gamma}_0 x^{(i)}
    =
    0.
\end{equation}
Substituting an exponential mode $e^{\lambda\tau}$ gives the characteristic equation and its roots:
\begin{equation}
\begin{aligned}
    \lambda^2+\bar{\gamma}_1\lambda+\bar{\gamma}_0&=0,\\
    \lambda_{1,2}&=
    \frac{-\bar{\gamma}_1
    \pm\sqrt{\bar{\gamma}_1^2-4\bar{\gamma}_0}}{2}.
\end{aligned}
\end{equation}
For distinct roots $\lambda_1$ and $\lambda_2$, the general solution is
\begin{equation}
    x^{(i)}(\tau)
    =
    a_i e^{\lambda_1\tau}
    +
    b_i e^{\lambda_2\tau},
\end{equation}
where $a_i$ and $b_i$ are determined by $x^{(i)}(0)$ and $\frac{\partial x^{(i)}}{\partial \tau}(0)$.
For a repeated root $\lambda_1=\lambda_2=\lambda$, the solution takes the form $x^{(i)}(\tau)=(a_i+b_i\tau)e^{\lambda\tau}$.
The common plane therefore has a local dynamical interpretation: the shared coefficients determine the same characteristic roots and temporal modes, while different initial conditions generate distinct trajectories within the same solution family.
This explains how different registered triplets can share a temporal-evolution prototype.

The temporal intervals enter the finite-difference estimates for each triplet. The shared coefficients are learned jointly from the state and derivative information across observations under this temporal sampling scheme, rather than prescribed solely from the intervals of an individual triplet.
The shared trajectory fitting problem is formulated as
\begin{equation}
\begin{aligned}
\text{(OP0)} \quad
&\min_{\bar{\gamma}_0,\bar{\gamma}_1}
\quad
\mathcal{R}(t_c)
=
\frac{1}{B}
\sum_{i=1}^{B}
\left\|
\bm \epsilon^{(i)}
\right\|_F^2
\\
&~~~~~=
\frac{1}{B}
\sum_{i=1}^{B}
\left\|
    \widehat{\frac{\partial^2 \bm p^{(i)}}{\partial t^2}}(t_c^{(i)})
    +
    \bar{\gamma}_1
    \widehat{\frac{\partial \bm p^{(i)}}{\partial t}}(t_c^{(i)})
    +
    \bar{\gamma}_0
    \bm p^{\delta,(i)}
\right\|_F^2 .
\end{aligned}
\label{eq:op0_batch_trajectory_supp}
\end{equation}

Here, $\bm \epsilon^{(i)}$ collects the residuals of the common plane relation across feature coordinates for the $i$-th registered triplet.
For fixed representations, \Eqref{eq:op0_batch_trajectory_supp} estimates the shared coefficients by minimizing the average squared residual over all $B$ observations.
The shared prototype is therefore learned by fitting the same plane to points from multiple observations, each constraining the coefficients that determine the common local solution family.
Pooling observations with similar local variation patterns can reduce the influence of individual approximation errors and improve the accuracy of prototype estimation.
During representation learning, penalizing these residuals also encourages the corresponding points to remain close to the shared plane, thereby regularizing the representation trajectories according to the learned prototype.

\subsubsection{Learning with Adaptive Trajectory Constraints}
\label{subsubsec:adaptive_constraints_supp}

The formulation in \cref{subsubsec:trajectories_estimation_supp} turns temporal variation modeling into fitting a unified second-order trajectory constraint over registered triplets.
This subsection provides the detailed implementation of this constraint in the self-supervised learning framework.

Directly solving \Eqref{eq:op0_batch_trajectory_supp} is impractical during training, since it is difficult to sample observations with similar local variation patterns from the entire dataset.
We therefore replace the coefficients with learnable channel-wise parameters and optimize them jointly with the representation learning objective.
To further accommodate diverse temporal variation patterns, we maintain $K$ groups of learnable parameter vectors
\begin{equation}
    \bm\Gamma
    =
    \{(\bm{\gamma}_0^{k},\bm{\gamma}_1^{k})\}_{k=1}^{K},
    \quad
    \bm{\gamma}_0^{k}, \bm{\gamma}_1^{k} \in \mathbb{R}^{d'},
    \label{eq:learnable_trajectory_coefficients_supp}
\end{equation}
where $\bm{\gamma}_0^{k}$ and $\bm{\gamma}_1^{k}$ are defined in the projected temporal space.

For the $i$-th observation and the $k$-th parameter group, the learnable parameters and finite-difference estimates yield the following trajectory residual:
\begin{equation}
    \bm r^{k,(i)}(t_c)
    =
    \widehat{\frac{\partial^2 \bm p^{(i)}}{\partial t^2}}(t_c)
    +
    \bm{\gamma}_1^{k} \odot
    \widehat{\frac{\partial \bm p^{(i)}}{\partial t}}(t_c)
    +
    \bm{\gamma}_0^{k} \odot \bm p^{\delta,(i)} .
    \label{eq:ode_residual_matrix_supp}
\end{equation}
Here, $\bm r^{k,(i)}(t_c)\in\mathbb{R}^{N\times d'}$ measures the mismatch between the finite-difference local evolution of the $i$-th registered triplet and the $k$-th learnable second-order trajectory constraint.
The operator $\odot$ denotes element-wise multiplication, and $\bm{\gamma}_0^{k},\bm{\gamma}_1^{k}\in\mathbb{R}^{d'}$ are broadcast along the token dimension.

Since the trajectory residual depends on reliable temporal correspondence, we compute it only over high-confidence registered patches.
For the $i$-th observation, let $\Omega_\rho^{(i)}$ denote the high-confidence patch set defined in \Eqref{eq:high_confidence_set}.
We construct a binary patch map $\bm M_\rho^{(i)}\in\{0,1\}^{N\times 1}$ as
\begin{equation}
    \bm M_\rho^{(i)}(n)
    =
    \begin{cases}
    1, & n\in\Omega_\rho^{(i)}, \\
    0, & n\notin\Omega_\rho^{(i)},
    \end{cases}
    \quad n=1,\ldots,N .
    \label{eq:high_confidence_map_supp}
\end{equation}
During residual computation, $\bm M_\rho^{(i)}$ is broadcast along the channel dimension.
For the $i$-th sample and the $k$-th parameter group, the mean squared residual over high-confidence patches is computed as
\begin{equation}
    \begin{aligned}
    \mathcal{R}^{k,(i)}(t_c)
    &=
    \frac{1}{d'|\Omega_\rho^{(i)}|}
    \left\|
    \bm M_\rho^{(i)} \odot \bm r^{k,(i)}(t_c)
    \right\|_F^2
    \\
    &=
    \frac{1}{d'|\Omega_\rho^{(i)}|}
    \sum_{n\in\Omega_\rho^{(i)}}
    \left\|
    \bm r_{n}^{k,(i)}(t_c)
    \right\|_2^2 .
    \end{aligned}
    \label{eq:op1_residual_supp}
\end{equation}
Here, the Frobenius norm aggregates the channel-wise deviations in the projected temporal space, and the binary patch map restricts the residual computation to reliable registered patches.

With multiple trajectory parameter groups, the adaptive trajectory fitting objective can be written as
\begin{equation}
    \begin{aligned}
    \text{(OP1)} \quad
    \min_{\bm\Gamma}
    \quad
    \frac{1}{B}
    \sum_{i=1}^{B}
    \min_{k\in\{1,\ldots,K\}}
    \mathcal{R}^{k,(i)}(t_c).
    \end{aligned}
    \label{eq:op1_adaptive_trajectory_supp}
\end{equation}
The inner minimization performs a hard selection of the parameter group that best fits the selected patch trajectories of each sample.
To make this selection differentiable and stable during training, we use a soft assignment over group-wise residuals for each sample:
\begin{equation}
    \omega_k^{(i)}
    =
    \frac{
    \exp(-\tau_o \mathcal{R}^{k,(i)}(t_c))
    }
    {
    \sum_{q=1}^{K}
    \exp(-\tau_o \mathcal{R}^{q,(i)}(t_c))
    },
    \label{eq:ode_soft_assignment_supp}
\end{equation}
where $\tau_o$ is a sharpness hyperparameter for the assignment.
The weights are computed separately for each sampled frame triplet and shared by its selected patches, while the parameter groups $\bm\Gamma$ are shared across samples.

The temporal variation loss is then obtained by averaging the sample-wise weighted residuals over the batch:
\begin{equation}
    \begin{aligned}
    \mathcal L_{\mathrm{var}}(t_c)
    =
    \frac{1}{B}
    \sum_{i=1}^{B}
    \sum_{k=1}^{K}
    \omega_k^{(i)} \mathcal{R}^{k,(i)}(t_c).
    \end{aligned}
    \label{eq:variation_loss_supp}
\end{equation}
This soft assignment gives larger weights to the parameter groups that better fit the selected patch trajectories of each sample, enabling differentiable selection of suitable variation patterns.
This formulation does not require samples within a mini-batch to share similar temporal variation patterns.
Together with the local consistency objective, $\mathcal L_{\mathrm{var}}$ encourages temporally corresponding representations to preserve local semantics while following coherent changes across ordered frames.

\section{Detailed Description of Experiments}
\label{sec:method_experiments}

\subsection{Training Datasets}

\textbf{Kinetics-400}~\cite{Kinetics} is a widely used large-scale benchmark for video representation learning and action understanding. It contains 400 categories of human actions collected from YouTube, with 239,789 available training videos. Each video clip is trimmed to approximately 10 seconds, providing diverse temporal patterns and motion cues for self-supervised pre-training. In our experiments, we use the Kinetics-400 training set as the pre-training dataset and extract frames at $2$ FPS. This sampling strategy reduces temporal redundancy while preserving sufficient motion cues for learning.

\subsection{Evaluation Benchmarks and Metrics}
\label{subsec:evaluation_ben_met}

\subsubsection{Benchmarks and Metrics on Pixel-level Tasks}
\label{subsubsec:pixel_level_benchmarks}

We first evaluate the quality of dense representations on three pixel-level video tasks, including video object segmentation on DAVIS-2017~\cite{DAVIS17}, human part segmentation on VIP~\cite{VIP}, and pose propagation on JHMDB~\cite{JHMDB}. 
Following recent works~\cite{SiamMAE,CropMAE,RSP}, all evaluations are conducted under a semi-supervised protocol without additional training or fine-tuning. Specifically, only the ground-truth annotation of the first frame is provided, and the model propagates pixel-wise predictions to subsequent frames using frozen pre-trained representations. This setting directly evaluates whether the learned dense representations preserve reliable spatiotemporal correspondence across frames.

\textbf{DAVIS-2017}~\cite{DAVIS17} is a standard benchmark for video object segmentation. It is used to evaluate the quality of dense object-level correspondence across video frames. We report three commonly used metrics for measuring the overall segmentation performance.

\textbf{1)} $\mathcal{J}_{\mathrm{m}}$ measures the average region similarity by computing the overlap between the predicted segmentation mask $P_i$ and the ground-truth mask $G_i$ for each video $\bm{V}_i$:
\begin{equation}
    \mathcal{J}_{\mathrm{m}} =
    \frac{1}{n}
    \sum_{i=1}^{n}
    \frac{|P_i \cap G_i|}{|P_i \cup G_i|}.
\label{eq:J_mean}
\end{equation}

\textbf{2)} $\mathcal{F}_{\mathrm{m}}$ measures the average contour accuracy by computing the harmonic mean of boundary precision $Pre_i$ and boundary recall $Rec_i$ for each video $\bm{V}_i$:
\begin{equation}
    \mathcal{F}_{\mathrm{m}} =
    \frac{1}{n}
    \sum_{i=1}^{n}
    \frac{2 \cdot Pre_i \cdot Rec_i}{Pre_i + Rec_i}.
\label{eq:F_mean}
\end{equation}

\textbf{3)} $\mathcal{J}\&\mathcal{F}_{\mathrm{m}}$ provides a comprehensive measure by averaging the region similarity and contour accuracy:
\begin{equation}
    \mathcal{J}\&\mathcal{F}_{\mathrm{m}} =
    \frac{\mathcal{J}_{\mathrm{m}} + \mathcal{F}_{\mathrm{m}}}{2}.
\label{eq:JF_mean}
\end{equation}

\textbf{VIP}~\cite{VIP} is a fine-grained video human parsing benchmark, which is adopted for evaluating semantic part propagation. The primary evaluation metric is mean Intersection over Union (mIoU). Given the class set $C$, the predicted segmentation mask $P_{i,j}$, and the ground-truth mask $G_{i,j}$ for class $C_j$ in video $\bm{V}_i$, mIoU is defined as
\begin{equation}
    \mathrm{mIoU} =
    \frac{1}{|C|}
    \sum_{j=1}^{|C|}
    \frac{1}{n}
    \sum_{i=1}^{n}
    \frac{|P_{i,j} \cap G_{i,j}|}{|P_{i,j} \cup G_{i,j}|}.
\label{eq:mIoU}
\end{equation}

\textbf{JHMDB}~\cite{JHMDB} is a benchmark dataset for human action understanding and pose estimation. In our experiments, it is adopted for human pose propagation and evaluated by Percentage of Correct Keypoints (PCK), which measures the accuracy of propagated keypoints under a normalized distance threshold:
\begin{equation}
    \mathrm{PCK}@k =
    \frac{1}{n}
    \sum_{i=1}^{n}
    \frac{1}{|S_i|}
    \sum_{j=1}^{|S_i|}
    \mathbbm{1}
    \left[
    D(\hat{p}_{i,j}, p_{i,j}) < k \cdot d_i
    \right],
\label{eq:PCK}
\end{equation}
where $S_i$ denotes the set of keypoints in video $\bm{V}_i$, $d_i$ is the human body scale for normalization, and $D(\hat{p}_{i,j}, p_{i,j})$ denotes the Euclidean distance between the predicted keypoint $\hat{p}_{i,j}$ and the ground-truth keypoint $p_{i,j}$. The parameter $k$ specifies the maximum allowable normalized localization error. Following previous works~\cite{SiamMAE,CropMAE,RSP,tcore}, we report PCK@0.1 and PCK@0.2.

\subsubsection{Benchmarks and Metrics on Frame-level Tasks}

We further evaluate the learned representations on frame-level video tasks, where temporally resolved features are required for recognizing action states and temporal boundaries. Frame-level representations are obtained by spatially averaging dense patch representations from each frame. 
For action segmentation, we conduct experiments on the third-person Breakfast~\cite{Breakfast} benchmark and the egocentric EgoProceL~\cite{EgoProceL} benchmark. The extracted features are fed into FACT~\cite{FACT}, which is trained for 150 epochs on Breakfast and 350 epochs on EgoProceL. 
For temporal action localization, we evaluate on the untrimmed THUMOS14~\cite{THUMOS14} dataset.

\textbf{Breakfast}~\cite{Breakfast} is a widely used third-person benchmark for fine-grained action segmentation. It contains 1,712 untrimmed videos of breakfast preparation activities. The dataset covers 10 breakfast-related activities and 48 fine-grained action units, with frame-level temporal annotations. Following the standard evaluation protocol, FACT~\cite{FACT} is trained on \textit{split2-4} and evaluated on \textit{split1}.

\textbf{EgoProceL}~\cite{EgoProceL} is an egocentric procedural activity benchmark designed for procedure learning from first-person videos. It consists of 62 hours of egocentric videos recorded by 130 subjects performing 16 procedural tasks, with temporally annotated key steps. Compared with third-person action segmentation datasets, EgoProceL emphasizes fine-grained procedural transitions under egocentric viewpoints, where action boundaries are closely related to hand-object interactions, camera motion, and changes in manipulated objects. The same action segmentation metrics are adopted for Breakfast and EgoProceL.

\textbf{1) $\mathrm{Edit}$} measures the sequence-level similarity between the predicted and ground-truth label sequences after collapsing consecutive duplicate labels. Let $\mathbf{y}=(y_{1},\ldots,y_{T})$ and $\hat{\mathbf{y}}=(\hat{y}_{1},\ldots,\hat{y}_{T})$ denote the ground-truth and predicted frame-wise label sequences, respectively. Let $\mathcal{C}(\cdot)$ denote the operation that removes consecutive repeated labels, $\mathrm{Lev}(\cdot,\cdot)$ denote the Levenshtein distance, and $|\cdot|$ denote the sequence length. The normalized edit score is defined as
\begin{equation}
    \mathrm{Edit}
    =
    1 -
    \frac{
    \mathrm{Lev}\!\left(\mathcal{C}(\hat{\mathbf{y}}),\,\mathcal{C}(\mathbf{y})\right)
    }{
    \max\!\left\{|\mathcal{C}(\hat{\mathbf{y}})|,\,|\mathcal{C}(\mathbf{y})|\right\}
    }.
\label{eq:Edit}
\end{equation}

\textbf{2) $\mathrm{Acc}$} denotes the frame-wise accuracy, measuring the percentage of frames assigned to the correct action label:
\begin{equation}
    \mathrm{Acc}
    =
    \frac{1}{T}
    \sum_{t=1}^{T}
    \mathbbm{1}\!\left\{\hat{y}_{t}=y_{t}\right\},
\label{eq:Acc}
\end{equation}
where $\mathbbm{1}\{\cdot\}$ is the indicator function.

\textbf{3) $\mathrm{F1@}k$} is the segmental $\mathrm{F1}$ score at temporal $\mathrm{IoU}$ threshold $k$. Let the ground-truth segment set be $\mathcal{S}=\{(s^{g}_{j},e^{g}_{j},c^{g}_{j})\}$ and the predicted segment set be $\hat{\mathcal{S}}=\{(s^{p}_{i},e^{p}_{i},c^{p}_{i})\}$, where $s$, $e$, and $c$ denote the start frame, end frame, and action class, respectively. For a predicted segment and a ground-truth segment of the same class, the temporal Intersection-over-Union is computed as
\begin{equation}
    \mathrm{IoU}
    \left(
    (s^{p}_{i},e^{p}_{i}),
    (s^{g}_{j},e^{g}_{j})
    \right)
    =
    \frac{
    \left\{
    \min(e^{p}_{i},e^{g}_{j})
    -
    \max(s^{p}_{i},s^{g}_{j})
    \right\}_{+}
    }{
    \max(e^{p}_{i},e^{g}_{j})
    -
    \min(s^{p}_{i},s^{g}_{j})
    }.
\label{eq:IoU}
\end{equation}
A predicted segment is counted as a true positive (\textit{TP}) if it uniquely matches a ground-truth segment of the same class with $\mathrm{IoU}\ge k$. Unmatched predictions are counted as false positives (\textit{FP}), and unmatched ground-truth segments are counted as false negatives (\textit{FN}). With precision $\mathit{P}=\frac{\mathit{TP}}{\mathit{TP}+\mathit{FP}}$ and recall $\mathit{R}=\frac{\mathit{TP}}{\mathit{TP}+\mathit{FN}}$, the segmental $\mathrm{F1}$ score is computed as
\begin{equation}
    \mathrm{F1@}k
    =
    \frac{2\,\mathit{P}\cdot\mathit{R}}{\mathit{P}+\mathit{R}},
\label{eq:F1_at_k}
\end{equation}
where $k\in\{0.10,0.25,0.50\}$ follows the standard thresholds for action segmentation.

\textbf{THUMOS14}~\cite{THUMOS14} is a standard benchmark for temporal action localization in untrimmed videos. It contains 413 temporally annotated untrimmed videos from 20 action categories, with 200 validation videos and 213 test videos for the localization task. In our experiments, the frame-level representations are extracted from the Kinetics-400 fine-tuned model described in \cref{subsec:evaluation_video}. Following AdaTAD~\cite{AdaTAD}, the temporal localization model is trained on the 200 validation videos for 60 epochs and evaluated on the 213 test videos. The task requires predicting both the action category and the temporal boundary of each action instance. The evaluation metric is mean Average Precision (mAP) under different temporal $\mathrm{IoU}$ thresholds.

For a temporal $\mathrm{IoU}$ threshold $k$, a predicted action instance is regarded as correct if it has the same action class as a ground-truth instance and its temporal overlap with that instance is no smaller than $k$. The average precision for class $c$ at threshold $k$ is denoted as $\mathrm{AP}_{c}(k)$. The mAP at threshold $k$ is computed as
\begin{equation}
    \mathrm{mAP}@k
    =
    \frac{1}{|C|}
    \sum_{c=1}^{|C|}
    \mathrm{AP}_{c}(k),
\label{eq:mAP_k}
\end{equation}
where $C$ denotes the set of action classes. Following the common protocol on THUMOS14, we report mAP at temporal $\mathrm{IoU}$ thresholds $k\in\{0.3,0.4,0.5,0.6,0.7\}$. The averaged score over these thresholds is denoted as
\begin{equation}
    \mathrm{mAP}@[0.3:0.1:0.7]
    =
    \frac{1}{5}
    \sum_{k\in\{0.3,0.4,0.5,0.6,0.7\}}
    \mathrm{mAP}@k.
\label{eq:mAP_03_07}
\end{equation}

\subsubsection{Benchmarks and Metrics on Video-level Tasks}

To evaluate the global semantic modeling ability of the learned representations, we consider three video-level tasks: action classification, video retrieval, and chiral recognition. For action classification and video retrieval, video-level representations are obtained by average pooling dense features over the temporal and spatial dimensions. For chiral recognition, we preserve temporal ordering by concatenating frame-level representations along the temporal dimension to form the final representation.

Action classification is evaluated on Kinetics-400~\cite{Kinetics} and HMDB51~\cite{HMDB51} under the full fine-tuning setting. For each dataset, the model is fine-tuned on the corresponding training split and evaluated on the evaluation split. We fine-tune the model for 50 epochs on Kinetics-400 and 100 epochs on HMDB51. During fine-tuning, each input clip contains 16 frames. During evaluation, we adopt a multi-view testing protocol with 5 temporal segments and 3 spatial crops. The standard top-$k$ accuracy is reported.

\textbf{$\mathrm{Acc@}k$} measures the percentage of evaluation videos whose ground-truth label appears in the top-$k$ predicted classes. Let $\mathbf{z}^{(i)}\in\mathbb{R}^{C}$ denote the predicted logits of the $i$-th video over $C$ classes, and let $y_i\in\{1,\ldots,C\}$ denote its ground-truth label. Denote by $\pi_k(\mathbf{z}^{(i)})$ the index set of the top-$k$ largest entries in $\mathbf{z}^{(i)}$. The top-$k$ accuracy is defined as
\begin{equation}
    \mathrm{Acc@}k
    =
    \frac{1}{N}
    \sum_{i=1}^{N}
    \mathbbm{1}
    \left\{
    y_i \in \pi_k\!\left(\mathbf{z}^{(i)}\right)
    \right\},
\label{eq:acck_video}
\end{equation}
where $N$ is the number of evaluation videos and $\mathbb{I}\{\cdot\}$ is the indicator function. We report $\mathrm{Acc@1}$ and $\mathrm{Acc@5}$.

Video retrieval is conducted on UCF101~\cite{UCF101} and HMDB51~\cite{HMDB51}. UCF101 contains 13,320 videos from 101 human action categories, and HMDB51 contains 6,766 videos from 51 human action categories. Following the standard protocol of previous work, pre-trained representations are directly used for retrieval without task-specific fine-tuning. For each video, 30 frames are uniformly sampled and encoded by the frozen backbone. The resulting dense features are averaged over temporal and spatial dimensions to obtain the final video-level representation. During evaluation, each video in the evaluation split is used once as the query, while the remaining videos form the reference set. Retrieval performance is averaged over all query rounds.

$\mathrm{Recall@}K$ measures the percentage of queries for which at least one positive sample is retrieved within the top-$K$ results. Let $\mathcal{P}_j$ denote the set of ranks of all positive samples for the $j$-th query. The recall at rank $K$ is defined as
\begin{equation}
    \mathrm{Recall@}K
    =
    \frac{1}{|\mathcal{Q}|}
    \sum_{j=1}^{|\mathcal{Q}|}
    \mathbbm{1}
    \left\{
    \min(\mathcal{P}_j) \le K
    \right\}.
\label{eq:recall_def}
\end{equation}

Chiral recognition evaluates whether a video representation is sensitive to the temporal ordering of visual changes. Following~\cite{Chirality}, we conduct linear probing on Chiral SSv2 and Chiral Charades. Chiral SSv2 is constructed from Something-Something-v2~\cite{SSV2}, a fine-grained action recognition benchmark centered on human-object interactions and motion-sensitive action categories. Chiral Charades is constructed from Charades~\cite{Charades}, a dataset of indoor daily activities with temporally localized action annotations. The chiral recognition task groups temporally opposite actions into binary classification problems and evaluates whether the representation can distinguish order-sensitive visual patterns.

For each chiral group, the backbone representation is frozen. Frame-level representations are extracted from each video and concatenated along the temporal dimension to form an order-preserving video representation. A linear classifier is then trained for binary classification within each chiral group. The final performance is reported as the average classification accuracy across all chiral groups.

\noindent\textbf{$\mathrm{Acc}$} measures the percentage of correctly classified videos. Let $\mathbf{z}^{(i)}\in\mathbb{R}^{2}$ denote the logits predicted by the linear classifier for the $i$-th video, and let $y_i\in\{0,1\}$ denote the ground-truth label within the corresponding chiral pair. The classification accuracy is defined as
\begin{equation}
    \mathrm{Acc}
    =
    \frac{1}{N}
    \sum_{i=1}^{N}
    \mathbbm{1}
    \left\{
    \arg\max_{c} z^{(i)}_{c} = y_i
    \right\},
\label{eq:acc_chiral}
\end{equation}
where $N$ is the number of evaluation videos and $\mathbbm{1}\{\cdot\}$ is the indicator function.

\subsubsection{Benchmarks and Metrics on Robotic Learning Tasks}

To examine whether the learned representations capture temporally coherent visual cues for vision-based robotic learning, we further evaluate them on CortexBench~\cite{CortexBench_1,CortexBench_2}. CortexBench evaluates frozen pre-trained visual representations by using them as perception modules for downstream embodied control tasks.

Following prior work~\cite{RSP,CortexBench_1,CortexBench_2}, all tasks are evaluated under a vision-based imitation learning setting. The visual backbone is kept frozen, and a behavior cloning model is trained from expert demonstrations using visual observations as input. Visual representations are extracted from either the \texttt{[CLS]} token or average-pooled patch tokens. All behavior cloning models are trained for 100 epochs and evaluated with three random seeds. For each domain, the final score is averaged over the selected tasks and random seeds. Following CortexBench, all reported scores are presented on a $[0,100]$ scale, where higher values indicate better performance.

\textbf{Adroit}~\cite{Adroit} is a dexterous manipulation benchmark based on an anthropomorphic robotic hand. It contains contact-rich manipulation tasks that require coordinated hand motion and object interaction. We evaluate two tasks, \texttt{Relocate} and \texttt{Reorient-Pen}. These tasks require manipulating an object toward a task-specified goal, such as moving an object to a target location or reorienting a pen-like object to a desired orientation. Following the CortexBench setting, the behavior cloning model is trained with 100 expert demonstrations for each task. We report the highest test success rate.

\textbf{MetaWorld}~\cite{MetaWorld} is a simulated benchmark for multi-task and meta-reinforcement learning in robotic manipulation. It contains a collection of tabletop manipulation environments built around a simulated Sawyer robot arm. We consider five representative tasks: \texttt{Assembly}, \texttt{Bin-Picking}, \texttt{Button-Press}, \texttt{Drawer-Open}, and \texttt{Hammer}. These tasks cover common tabletop manipulation behaviors, including insertion, picking, button pressing, drawer opening, and tool-use actions. Following the CortexBench setting, the behavior cloning model is trained with 25 expert demonstrations for each task. We report the highest test success rate.

\textbf{DeepMind Control (DMC)}~\cite{DMC} is a continuous control suite with standardized task structures and interpretable reward functions. It is commonly used as an image-based control benchmark with physics simulation. In our experiments, we evaluate five tasks: \texttt{Finger-Spin}, \texttt{Reacher-Easy}, \texttt{Cheetah-Run}, \texttt{Walker-Stand}, and \texttt{Walker-Walk}. These tasks cover both manipulation-style and locomotion-style control scenarios, where the frozen visual representation provides visual state information for action prediction. Following the CortexBench setting, the behavior cloning model is trained with 100 expert demonstrations for each task. We report the highest test scaled score.

\textbf{TriFinger}~\cite{TriFinger} is a three-finger robotic manipulation benchmark for object interaction. It focuses on controlling a compact multi-finger system to interact with a cube. We evaluate \texttt{Reach-Cube} and \texttt{Move-Cube}. These tasks require the behavior cloning model to predict actions for reaching or moving the cube according to the task goal. Following the CortexBench setting, the behavior cloning model is trained with 100 expert demonstrations for each task. We report the maximum average scaled reward on the test dataset.

\begin{table*}[t]
  \centering
  \caption{Pre-training hyperparameter settings of the proposed \methodname{} framework.}
  \label{tab:training_settings}
  \hfill
  \begin{subtable}[t]{0.51\textwidth}
    \centering
    \caption{Sampling strategy and loss weighting schedule.}
    \label{tab:training_settings_sampling_loss}
    \resizebox{\linewidth}{!}{%
      \begin{tabular}{lc|c}
        \toprule
        \multicolumn{1}{c}{Hyperparameter} & \multicolumn{1}{c|}{Notation} & Value \\
        \midrule
        \rowcolor[rgb]{ .867, .922, .969} \multicolumn{3}{c}{Sampling strategy} \\
        \midrule
        Current frame range & / & $[0.3, 0.7]$ \\
        Past frame offset range & $[\alpha, \beta]$ & $[0.10, 0.20]$ \\
        Future frame offset range & $[\alpha, \beta]$ & $[0.10, 0.20]$ \\
        Number of global views & / & 2 \\
        Number of local views & / & 8 \\
        Global crop size & / & $(224 \times 224)$ \\
        Local crop size & / & $(96 \times 96)$ \\
        Past and future frame size & / & $(224 \times 224)$ \\
        Mask probability & / & 0.5 \\
        Mask ratio & / & $[0.1, 0.5]$ \\
        \midrule
        \rowcolor[rgb]{ .867, .922, .969} \multicolumn{3}{c}{Loss function} \\
        \midrule
        Weight of reconstruction loss & $\lambda_1$ & 0.8 \\
        Weight of squeezing loss & $\lambda_2$ & 20 \\
        Weight of DINO loss & $\lambda_3$ & 1.0 \\
        Peak weight of temporal variation loss & $\lambda_4$ & 0.5 \\
        Start progress of temporal variation loss & $\xi_{\mathrm{s}}$ & 0.6 \\
        Peak progress of temporal variation loss & $\xi_{\mathrm{p}}$ & 0.8 \\
        Final relative weight of $\mathcal{L}_{var}$ & $r_{\mathrm{end}}$ & 0.2 \\
        \bottomrule
      \end{tabular}%
    }
  \end{subtable}
  \hfill
  \begin{subtable}[t]{0.46\textwidth}
    \centering
    \caption{Network architecture and optimization settings.}
    \label{tab:training_settings_optimization}
    \resizebox{\linewidth}{!}{%
      \begin{tabular}{lc|cc}
        \toprule
        \multicolumn{1}{c}{\multirow{2}[4]{*}{Hyperparameter}}
        & \multicolumn{1}{c|}{\multirow{2}[4]{*}{Notation}}
        & \multicolumn{2}{c}{Value} \\
        \cmidrule{3-4}
        & & ViT-S/16 & ViT-B/16 \\
        \midrule
        \rowcolor[rgb]{ .867, .922, .969} \multicolumn{4}{c}{Network architecture} \\
        \midrule
        Number of ViT encoder blocks & / & \multicolumn{2}{c}{12} \\
        Patch size & $p$ & \multicolumn{2}{c}{16} \\
        ViT feature dimension & $d$ & 384 & 768 \\
        Projection $\phi$ dimension & $d'$ & \multicolumn{2}{c}{256} \\
        Number of parameter groups & $K$ & \multicolumn{2}{c}{3} \\
        Retained patch ratio & $\rho$ & \multicolumn{2}{c}{0.6} \\
        Soft-assignment sharpness & $\tau_o$ & \multicolumn{2}{c}{2.0} \\
        \midrule
        \rowcolor[rgb]{ .867, .922, .969} \multicolumn{4}{c}{Optimization settings} \\
        \midrule
        Optimizer & / & \multicolumn{2}{c}{AdamW} \\
        Learning rate scheduler & / & \multicolumn{2}{c}{Cosine decay} \\
        Minimum learning rate & $lr_{\min}$ & \multicolumn{2}{c}{$1 \times 10^{-6}$} \\
        Weight decay & / & \multicolumn{2}{c}{$0.04 \rightarrow 0.4$} \\
        EMA momentum & / & \multicolumn{2}{c}{$0.992 \rightarrow 1$} \\
        Base learning rate & $blr$ & $2 \times 10^{-3}$ & $1 \times 10^{-3}$ \\
        Actual learning rate & $lr$ & \multicolumn{2}{c}{$blr\sqrt{bs/256}$} \\
        PRM learning rate & / & \multicolumn{2}{c}{$0.1 \times lr$} \\
        Epochs & / & 400 & 200/400 \\
        Warm-up epochs & / & 20 & 10/20 \\
        Batch size & $bs$ & 256 & 128 \\
        \bottomrule
      \end{tabular}%
    }
  \end{subtable}
  \hfill
  \vspace{-8pt}
\end{table*}

\subsection{Training Settings}
\label{app:app_training_settings}

\textbf{Network architecture.}
We adopt ViT-Small (ViT-S) and ViT-Base (ViT-B)~\cite{Vision_Transformer} with a patch size of 16 as the backbone models. The feature dimension is 384 for ViT-S/16 and 768 for ViT-B/16. By default, the number of trajectory parameter groups is set to $K=3$, and the projection dimension of the separate temporal modeling space is set to $d'=256$.

\textbf{Pre-training sampling strategy.}
All models using the proposed framework are pre-trained on Kinetics-400~\cite{Kinetics}. We extract frames from each video at $2$ FPS to reduce temporal redundancy while retaining sufficient motion cues. For each video, a current frame is sampled first, and its past and future reference frames are then selected according to the temporal offset range. By default, the lower and upper bounds of the offset ratio are set to $\alpha=0.10$ and $\beta=0.20$. For the current frame, we generate $2$ global views of size $224\times224$ and $8$ local views of size $96\times96$. Standard data augmentations, including color jittering, grayscale conversion, and Gaussian blur, are applied during pre-training. Following DINOv2~\cite{DINOv2}, $50\%$ of the global crops in the student branch are masked, where $10\%\sim50\%$ of patches are randomly selected as masked patches.

\textbf{Pre-training schedule.}
The ViT-S/16 model is trained for $400$ epochs with a batch size of $256$, where the first $20$ epochs are used for warm-up. The ViT-B/16 model is trained with a batch size of $128$. Specifically, ViT-B/16 is trained for $200$ epochs for pixel-level downstream tasks and for $400$ epochs for the other downstream tasks. The first $10$ epochs are used for warm-up when ViT-B/16 is trained for $200$ epochs, and the warm-up length is scaled accordingly for the $400$-epoch setting.

\textbf{Optimization strategy.}
The student branch is optimized by AdamW, and the teacher branch is updated by the exponential moving average of the student weights. The base learning rate $\textit{blr}$ is set to $2\times10^{-3}$ for ViT-S/16 and $1\times10^{-3}$ for ViT-B/16. The learning rate follows a cosine decay schedule and decays to $1\times10^{-6}$ at the end of pre-training. The actual learning rate is computed as
$\textit{lr}=\textit{blr}\sqrt{\textit{bs}/256}$, where $\textit{bs}$ denotes the batch size. For the Patch Registration Module (PRM), the learning rate is set to $0.1\times\textit{lr}$.
When varying $\xi_{\mathrm{s}}$, we set $\xi_{\mathrm{p}}=(1+\xi_{\mathrm{s}})/2$ so that the increasing and decreasing stages each occupy half of the remaining pre-training period.


\textbf{Loss weights and variation-aware scheduling.}
The loss weights are set to $\lambda_1=0.8$, $\lambda_2=20$, $\lambda_3=1.0$, and $\lambda_4=0.5$. Here, $\lambda_1$, $\lambda_2$, and $\lambda_3$ follow the default settings of~\cite{tcore}, while only $\lambda_4$ is tuned in our experiments.
The variation-aware loss $\mathcal{L}_{\mathrm{var}}$ is introduced only in the latter stage of pre-training, where $\xi\in[0,1]$ denotes the normalized pre-training progress, with $0$ and $1$ corresponding to its start and end, respectively.
The effective weight of $\mathcal{L}_{\mathrm{var}}$ is $\lambda_4 \times s(\xi)$, where
\begin{equation}
s(\xi)=
\begin{cases}
0,
& 0\leq\xi<\xi_{\mathrm{s}},\\[1mm]
\displaystyle\frac{1}{2}
\left[
1-\cos\!\left(
\frac{\xi-\xi_{\mathrm{s}}}
{\xi_{\mathrm{p}}-\xi_{\mathrm{s}}}\pi
\right)
\right],
& \xi_{\mathrm{s}}\leq\xi<\xi_{\mathrm{p}},\\[2mm]
\displaystyle r_{\mathrm{end}}
+\frac{1-r_{\mathrm{end}}}{2}
\left[
1+\cos\!\left(
\frac{\xi-\xi_{\mathrm{p}}}
{1-\xi_{\mathrm{p}}}\pi
\right)
\right],
& \xi_{\mathrm{p}}\leq\xi\leq1.
\end{cases}
\label{eq:variation_loss_schedule}
\end{equation}
Here, $\xi_{\mathrm{s}}$ and $\xi_{\mathrm{p}}$ denote the progress at which the loss is introduced and reaches its peak weight, respectively, with $0\leq\xi_{\mathrm{s}}<\xi_{\mathrm{p}}<1$, and $r_{\mathrm{end}}$ denotes the final weight relative to the peak weight $\lambda_4$.
By default, we set $\xi_{\mathrm{s}}=0.6$, $\xi_{\mathrm{p}}=0.8$, and $r_{\mathrm{end}}=0.2$.

The general training hyperparameters of our \methodname{} framework are summarized in \Tbref{tab:training_settings}. The experiments are implemented in PyTorch and mainly conducted on a Linux workstation equipped with an AMD EPYC 9654 96-Core Processor and four NVIDIA RTX 4090 GPUs.

\subsection{Competitors}
\label{subsec:competitors}

We compare our method with representative self-supervised representation learning methods. Following the organization in the main paper, these competitors are divided into two categories: \textit{Contrastive learning methods} and \textit{Masked modeling methods}.

\noindent\textbf{1) Contrastive learning methods}:
\begin{itemize}
  \item \textbf{SimCLR}~\cite{SimCLRv1} learns visual representations by maximizing the agreement between differently augmented views of the same image under a contrastive learning objective.
  
  \item \textbf{MoCo v3}~\cite{MoCov3} extends the momentum contrastive framework to Vision Transformers and studies stable self-supervised ViT training with a momentum encoder.
  
  \item \textbf{DINO}~\cite{DINO} learns self-supervised ViT representations through a teacher-student self-distillation framework without using manual annotations.
  
\end{itemize}

\noindent\textbf{2) Masked modeling methods}:
\begin{itemize}
  \item \textbf{MAE}~\cite{MAE} masks random image patches and reconstructs the missing pixels with an asymmetric encoder-decoder architecture.
  
  \item \textbf{MAE-ST}~\cite{MAE_ST} extends masked autoencoding to videos by masking spatiotemporal patches and reconstructing them in pixel space.
  
  \item \textbf{VideoMAE}~\cite{VideoMAE} applies masked autoencoding to videos with tube masking and an extremely high masking ratio for masked video reconstruction.
  
  \item \textbf{DropMAE}~\cite{DropMAE} studies video masked autoencoding for temporal matching tasks and introduces spatial-attention dropout to reduce the dependence on static spatial cues during reconstruction.

  \item \textbf{SiamMAE}~\cite{SiamMAE} learns visual correspondence from paired video frames with asymmetric masking, where a partially masked target frame is reconstructed by attending to an unmasked reference frame.

  \item \textbf{CropMAE}~\cite{CropMAE} follows the Siamese masked autoencoding design but constructs paired views from different crops of the same image, enabling pre-training without paired video frames.

  \item \textbf{RSP}~\cite{RSP} learns visual representations through stochastic future-frame prediction, where the model predicts plausible future observations while retaining an auxiliary masked modeling objective.

  \item \textbf{iBOT}~\cite{iBOT} performs masked image modeling with an online tokenizer, where the teacher network provides targets for both class-token self-distillation and masked patch-token prediction.

  \item \textbf{DINOv2}~\cite{DINOv2} scales discriminative self-supervised ViT pre-training by combining image-level and patch-level objectives with curated data and improved training recipes.

  \item \textbf{Ada-iBOT}~\cite{co_settle} and \textbf{Ada-DINOv2}~\cite{co_settle} initialize from ImageNet-pretrained iBOT and DINOv2 models, respectively, and further adapt the image representations on Kinetics-400.

  \item \textbf{V-JEPA 2.1}~\cite{vjepa2_1} extends video joint-embedding predictive learning with dense prediction objectives, producing spatially grounded and temporally consistent visual representations.
  
\end{itemize}

\section{Additional Experimental Results}

\subsection{Training Dynamics}

\begin{figure*}[htbp]
  \centering
  \begin{subfigure}{0.33\linewidth}
    \includegraphics[width=0.86\linewidth]{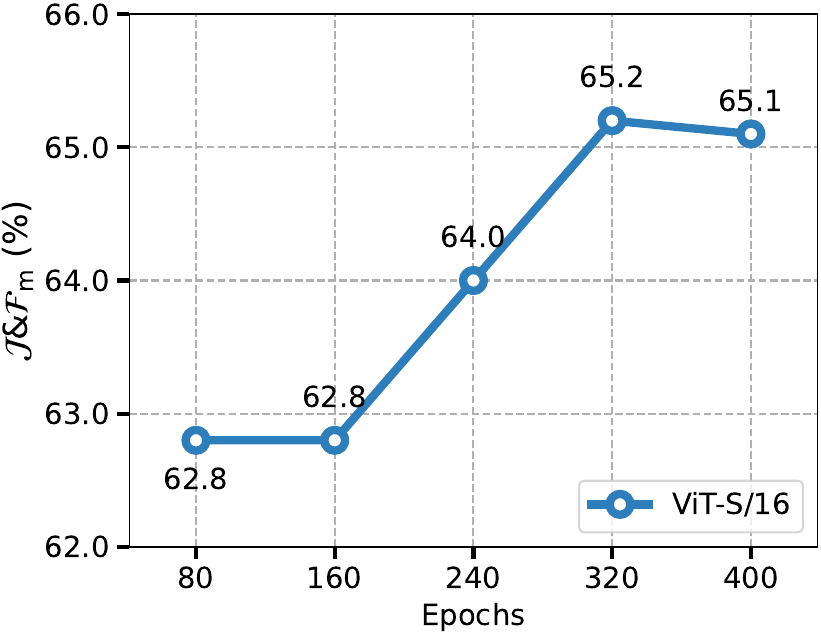}
    \caption{$\mathcal{J}\&\mathcal{F}_{\mathrm{m}}$ on DAVIS-2017 with ViT-S/16.}
    \label{fig:plot_DAVIS_S}
  \end{subfigure}
  \begin{subfigure}{0.33\linewidth}
    \includegraphics[width=0.86\linewidth]{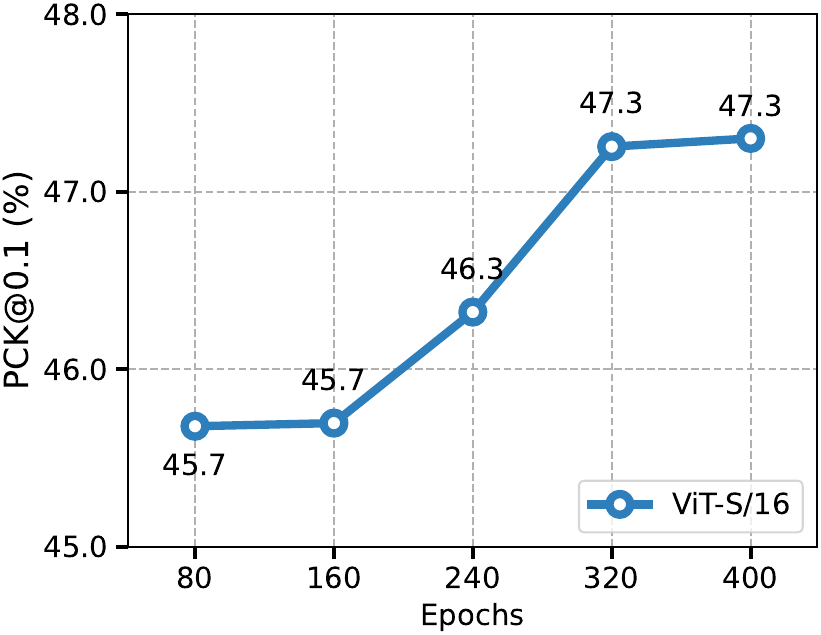}
    \caption{PCK@0.1 on JHMDB with ViT-S/16.}
    \label{fig:plot_JHMDB_S}
  \end{subfigure}
  \begin{subfigure}{0.33\linewidth}
    \includegraphics[width=0.86\linewidth]{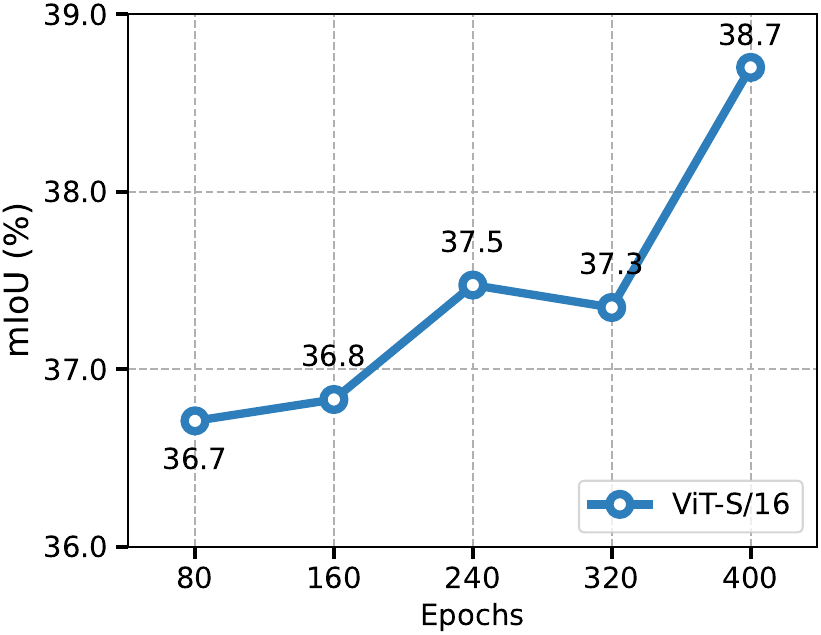}
    \caption{mIoU on VIP with ViT-S/16.}
    \label{fig:plot_VIP_S}
  \end{subfigure}

  \par

  \begin{subfigure}{0.33\linewidth}
    \includegraphics[width=0.86\linewidth]{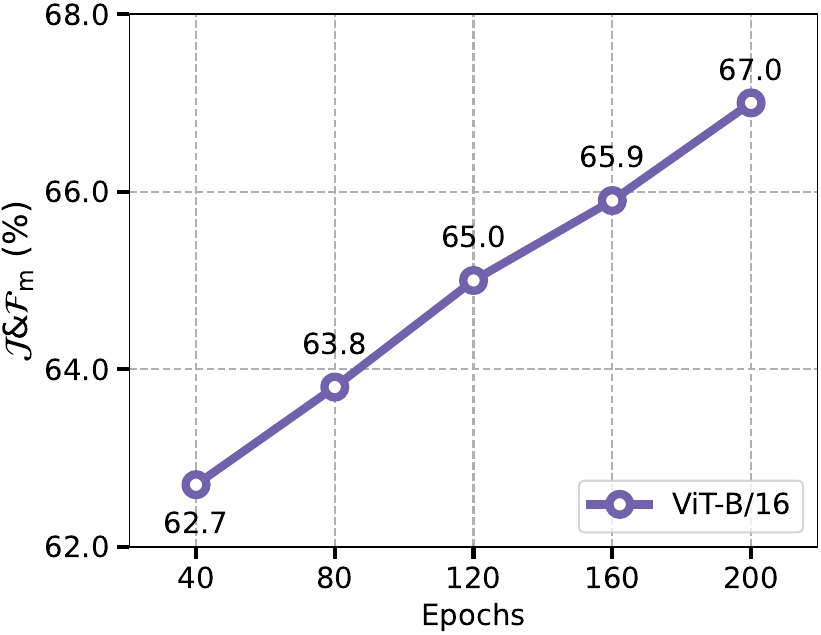}
    \caption{$\mathcal{J}\&\mathcal{F}_{\mathrm{m}}$ on DAVIS-2017 with ViT-B/16.}
    \label{fig:plot_DAVIS_B}
  \end{subfigure}
  \begin{subfigure}{0.33\linewidth}
    \includegraphics[width=0.86\linewidth]{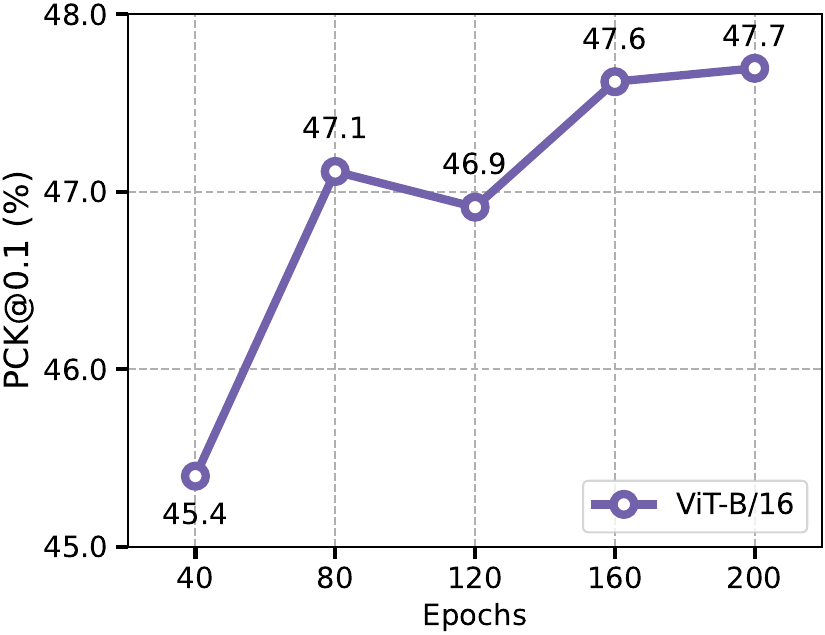}
    \caption{PCK@0.1 on JHMDB with ViT-B/16.}
    \label{fig:plot_JHMDB_B}
  \end{subfigure}
  \begin{subfigure}{0.33\linewidth}
    \includegraphics[width=0.86\linewidth]{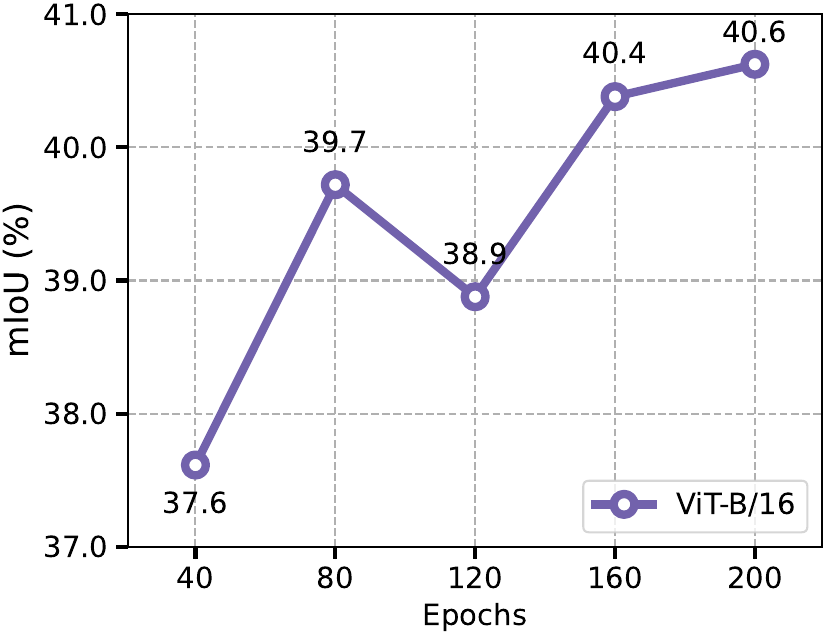}
    \caption{mIoU on VIP with ViT-B/16.}
    \label{fig:plot_VIP_B}
  \end{subfigure}

  \caption{Training dynamics on three pixel-level benchmarks with the ViT-S/16 and ViT-B/16 backbones.}
  \label{fig:dynamic_pixel}
\end{figure*}

\begin{figure*}[htbp]
  \centering
  \begin{subfigure}{0.33\linewidth}
    \includegraphics[width=0.86\linewidth]{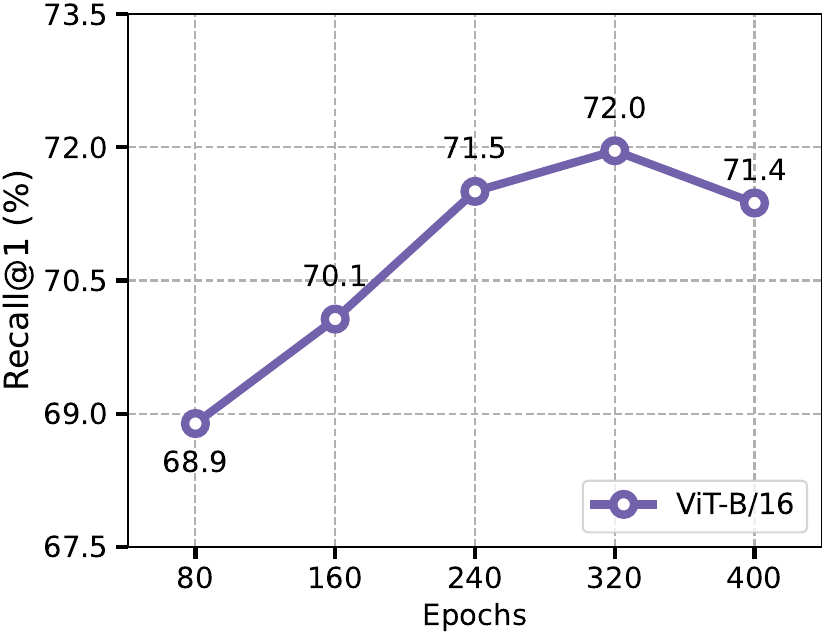}
    \caption{R@1 on HMDB51 retrieval.}
    \label{fig:plot_retrieval_R1_HMDB51}
  \end{subfigure}
  \begin{subfigure}{0.33\linewidth}
    \includegraphics[width=0.86\linewidth]{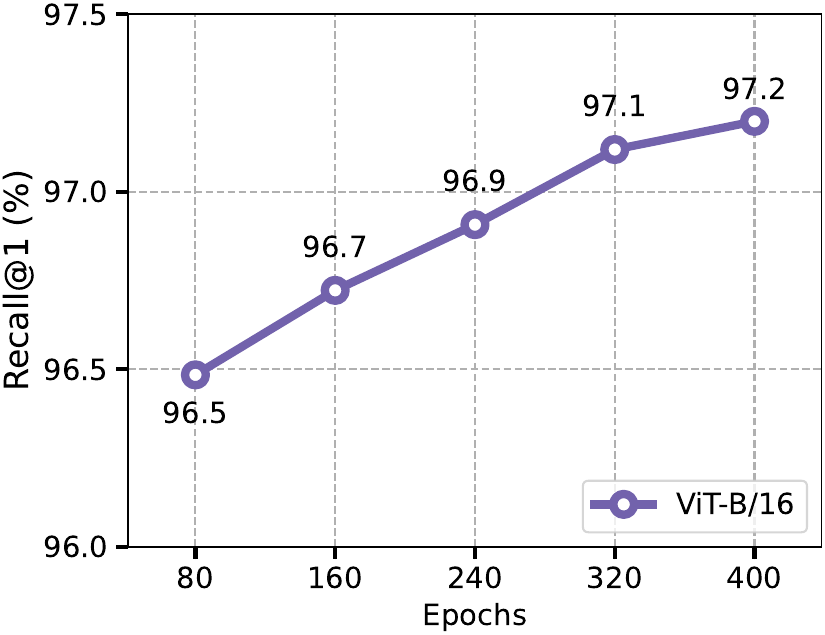}
    \caption{R@1 on UCF101 retrieval.}
    \label{fig:plot_retrieval_R1_UCF101}
  \end{subfigure}
  \begin{subfigure}{0.33\linewidth}
    \includegraphics[width=0.86\linewidth]{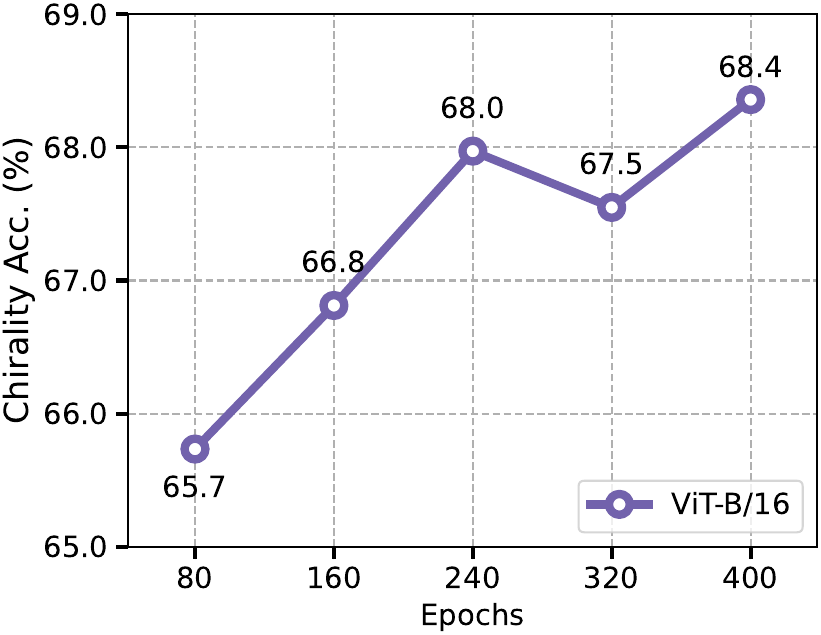}
    \caption{Chirality accuracy on Chiral Charades.}
    \label{fig:plot_chiral_charades}
  \end{subfigure}

  \caption{Training dynamics on three video-level benchmarks with the ViT-B/16 backbone.}
  \vspace{-10pt}
  \label{fig:dynamic_video}
\end{figure*}

To analyze the training dynamics of the proposed framework, we evaluate intermediate checkpoints on both pixel-level and video-level downstream tasks. \Fgref{fig:dynamic_pixel} reports the performance on DAVIS-2017, JHMDB, and VIP with ViT-S/16 and ViT-B/16 backbones, while \Fgref{fig:dynamic_video} reports the video-level performance on HMDB51 retrieval, UCF101 retrieval, and Chiral Charades with ViT-B/16. Overall, the downstream performance exhibits a generally increasing trend as pre-training progresses, and the larger ViT-B/16 backbone further improves the final performance compared with ViT-S/16 on pixel-level tasks.

The consistent gains on both pixel-level and video-level benchmarks indicate an overall improvement in representation quality throughout pre-training. The improvements on dense propagation tasks reflect stronger local correspondence and object discrimination, while the gains on retrieval and chiral recognition suggest better semantic separability and temporal sensitivity. These results show that the proposed training process improves complementary properties of video representations, rather than only benefiting a single type of downstream evaluation.

\subsection{Detailed Results on Robotic Learning Tasks}

\Tbref{tab:robot_detailed_results} reports detailed task-wise results on the four robotic learning domains. Overall, \methodname{} achieves stronger average performance across the selected CortexBench domains, indicating that the learned representations are effective for vision-based imitation learning. The improvements are more evident in object-centric manipulation domains, such as MetaWorld and Adroit, where visual observations contain continuously changing object states, hand or end-effector configurations, and task progress. This is consistent with the design of our method: by imposing a temporal variation constraint in the latent space, \methodname{} encourages corresponding representations to evolve more regularly across adjacent frames, making them more informative for downstream imitation models.

The results on DMC and TriFinger further show that the learned representations remain competitive under different types of visual dynamics, including continuous body-state changes and fine-grained cube interaction. Compared with methods that mainly emphasize frame-level appearance or local reconstruction, \methodname{} better captures structured temporal changes in the visual input. These results support the benefit of modeling latent representation trajectories for robotic learning tasks that require temporally coherent visual state estimation.

\begin{table*}[htbp]
  \centering
  \caption{Detailed evaluation results on vision-based robotic learning tasks on the CortexBench benchmark. The best and second-best results in each column are highlighted in \textbf{bold} and \underline{underlined}, respectively. $^\ddagger$ indicates that VideoMAE uses averaged patch-token representations because its standard encoder does not employ a \texttt{[CLS]} token, while the other methods use the \texttt{[CLS]} token.}
  \label{tab:robot_detailed_results}
  \setlength{\tabcolsep}{3pt}
  \renewcommand{\arraystretch}{1.35}
  \resizebox{\textwidth}{!}{%
    \begin{tabular}{ccc|cccccc}
    \toprule
    \multirow{2}[4]{*}{Models} & \multirow{2}[4]{*}{Backbone} & \multirow{2}[4]{*}{Tokens} & \multicolumn{6}{c}{MetaWorld} \\
    \cmidrule{4-9}
          &       &       & \texttt{Assembly} & \texttt{Bin-Picking} & \texttt{Button-Press} & \texttt{Drawer-Open} & \texttt{Hammer-V2} & \textbf{Average} \\
    \midrule
    VideoMAE$^\ddagger$ & ViT-S/16 & patch & \cellcolor[rgb]{ .765,  .859,  .945}\underline{73.33} & \cellcolor[rgb]{ .784,  .871,  .949}65.33 & \cellcolor[rgb]{ .804,  .882,  .953}72.00 & \cellcolor[rgb]{ .608,  .761,  .902}\textbf{100.00} & \cellcolor[rgb]{ .733,  .839,  .937}\underline{96.00} & \cellcolor[rgb]{ .812,  .886,  .957}\underline{81.33} \\
    SiamMAE & ViT-S/16 & [CLS] & \cellcolor[rgb]{ .89,  .937,  .976}68.00 & \cellcolor[rgb]{ .918,  .953,  .98}56.00 & \cellcolor[rgb]{ .871,  .925,  .969}66.67 & \cellcolor[rgb]{ .608,  .761,  .902}\textbf{100.00} & \cellcolor[rgb]{ .733,  .839,  .937}\underline{96.00} & \cellcolor[rgb]{ .918,  .953,  .98}77.33 \\
    CropMAE & ViT-S/16 & [CLS] & \cellcolor[rgb]{ .827,  .898,  .957}70.67 & \cellcolor[rgb]{ .686,  .812,  .922}\underline{72.00} & \cellcolor[rgb]{ .918,  .953,  .98}62.67 & \cellcolor[rgb]{ .608,  .761,  .902}\textbf{100.00} & \cellcolor[rgb]{ .796,  .878,  .949}94.67 & \cellcolor[rgb]{ .847,  .91,  .965}80.00 \\
    RSP & ViT-S/16 & [CLS] & \cellcolor[rgb]{ .918,  .953,  .98}66.67 & \cellcolor[rgb]{ .686,  .812,  .922}\underline{72.00} & \cellcolor[rgb]{ .773,  .863,  .945}\underline{74.67} & \cellcolor[rgb]{ .608,  .761,  .902}\textbf{100.00} & \cellcolor[rgb]{ .918,  .953,  .98}92.00 & \cellcolor[rgb]{ .82,  .894,  .957}81.07 \\
    \methodnameb{} \textbf{(Ours)}  & ViT-S/16 & [CLS] & \cellcolor[rgb]{ .608,  .761,  .902}\textbf{80.00} & \cellcolor[rgb]{ .608,  .761,  .902}\textbf{77.33} & \cellcolor[rgb]{ .608,  .761,  .902}\textbf{88.00} & \cellcolor[rgb]{ .608,  .761,  .902}\textbf{100.00} & \cellcolor[rgb]{ .608,  .761,  .902}\textbf{98.67} & \cellcolor[rgb]{ .608,  .761,  .902}\textbf{88.80} \\
    \midrule
    \midrule
    \multirow{2}[4]{*}{Models} & \multirow{2}[4]{*}{Backbone} & \multirow{2}[4]{*}{Tokens} & \multicolumn{6}{c}{DeepMind Control} \\
    \cmidrule{4-9}
          &       &       & \texttt{Walker-Stand} & \texttt{Walker-Walk} & \texttt{Reacher-Easy} & \texttt{Cheetah-Run} & \texttt{Finger-Spin} & \textbf{Average} \\
    \midrule
    VideoMAE$^\ddagger$ & ViT-S/16 & patch & \cellcolor[rgb]{ .996,  .957,  .91}50.26 & \cellcolor[rgb]{ .996,  .957,  .91}16.77 & \cellcolor[rgb]{ .996,  .914,  .831}57.74 & \cellcolor[rgb]{ .996,  .957,  .91}7.63 & \cellcolor[rgb]{ .996,  .957,  .91}11.39 & \cellcolor[rgb]{ .996,  .957,  .91}28.76 \\
    SiamMAE & ViT-S/16 & [CLS] & \cellcolor[rgb]{ .992,  .855,  .725}\underline{67.60} & \cellcolor[rgb]{ .996,  .902,  .812}28.17 & \cellcolor[rgb]{ .988,  .796,  .624}\underline{75.87} & \cellcolor[rgb]{ .996,  .949,  .898}8.05 & \cellcolor[rgb]{ .988,  .816,  .659}48.97 & \cellcolor[rgb]{ .992,  .827,  .678}45.73 \\
    CropMAE & ViT-S/16 & [CLS] & \cellcolor[rgb]{ .996,  .922,  .847}56.25 & \cellcolor[rgb]{ .996,  .929,  .855}22.99 & \cellcolor[rgb]{ .992,  .875,  .761}63.97 & \cellcolor[rgb]{ .988,  .776,  .588}\underline{17.33} & \cellcolor[rgb]{ .988,  .776,  .584}\underline{60.01} & \cellcolor[rgb]{ .992,  .843,  .702}44.11 \\
    RSP & ViT-S/16 & [CLS] & \cellcolor[rgb]{ .984,  .757,  .553}\textbf{83.73} & \cellcolor[rgb]{ .984,  .757,  .553}\textbf{57.04} & \cellcolor[rgb]{ .996,  .957,  .91}50.90 & \cellcolor[rgb]{ .984,  .757,  .553}\textbf{18.32} & \cellcolor[rgb]{ .984,  .757,  .553}\textbf{64.51} & \cellcolor[rgb]{ .984,  .757,  .553}\textbf{54.90} \\
    \methodnameb{} \textbf{(Ours)}  & ViT-S/16 & [CLS] & \cellcolor[rgb]{ .992,  .871,  .753}65.16 & \cellcolor[rgb]{ .992,  .867,  .745}\underline{35.60} & \cellcolor[rgb]{ .984,  .757,  .553}\textbf{81.88} & \cellcolor[rgb]{ .996,  .933,  .871}8.91 & \cellcolor[rgb]{ .988,  .796,  .624}54.26 & \cellcolor[rgb]{ .988,  .804,  .631}\underline{49.16} \\
    \midrule
    \midrule
    \multirow{2}[4]{*}{Models} & \multirow{2}[4]{*}{Backbone} & \multirow{2}[4]{*}{Tokens} & \multicolumn{3}{c|}{TriFinger} & \multicolumn{3}{c}{Adroit} \\
    \cmidrule{4-6}\cmidrule{7-9}
          &       &       & \texttt{Move} & \texttt{Reach} & \multicolumn{1}{c|}{\textbf{Average}} & \texttt{Pen} & \texttt{Relocate} & \textbf{Average} \\
    \midrule
    VideoMAE$^\ddagger$ & ViT-S/16 & patch & \cellcolor[rgb]{ .663,  .816,  .557}\textbf{57.73} & \cellcolor[rgb]{ .663,  .816,  .557}\textbf{85.98} & \multicolumn{1}{c|}{\cellcolor[rgb]{ .663,  .816,  .557}\textbf{71.85}} & \cellcolor[rgb]{ .8,  .663,  .902}\underline{77.33} & \cellcolor[rgb]{ .878,  .796,  .941}\underline{18.67} & \cellcolor[rgb]{ .878,  .796,  .941}\underline{48.00} \\
    SiamMAE & ViT-S/16 & [CLS] & \cellcolor[rgb]{ .8,  .89,  .737}43.62 & \cellcolor[rgb]{ .933,  .965,  .914}74.84 & \multicolumn{1}{c|}{\cellcolor[rgb]{ .839,  .914,  .788}59.23} & \cellcolor[rgb]{ .894,  .82,  .949}73.33 & \cellcolor[rgb]{ .894,  .82,  .949}17.33 & \cellcolor[rgb]{ .922,  .867,  .961}45.33 \\
    CropMAE & ViT-S/16 & [CLS] & \cellcolor[rgb]{ .965,  .98,  .957}26.22 & \cellcolor[rgb]{ .965,  .98,  .957}73.47 & \multicolumn{1}{c|}{\cellcolor[rgb]{ .965,  .98,  .957}49.85} & \cellcolor[rgb]{ .953,  .922,  .976}70.67 & \cellcolor[rgb]{ .894,  .82,  .949}17.33 & \cellcolor[rgb]{ .945,  .906,  .973}44.00 \\
    RSP & ViT-S/16 & [CLS] & \cellcolor[rgb]{ .714,  .843,  .62}52.80 & \cellcolor[rgb]{ .675,  .824,  .569}\underline{85.64} & \multicolumn{1}{c|}{\cellcolor[rgb]{ .702,  .839,  .608}69.22} & \cellcolor[rgb]{ .831,  .714,  .918}76.00 & \cellcolor[rgb]{ .953,  .922,  .976}10.67 & \cellcolor[rgb]{ .953,  .922,  .976}43.33 \\
    \methodnameb{} \textbf{(Ours)}  & ViT-S/16 & [CLS] & \cellcolor[rgb]{ .682,  .827,  .58}\underline{55.95} & \cellcolor[rgb]{ .675,  .824,  .573}85.55 & \multicolumn{1}{c|}{\cellcolor[rgb]{ .678,  .827,  .58}\underline{70.75}} & \cellcolor[rgb]{ .765,  .608,  .882}\textbf{78.67} & \cellcolor[rgb]{ .765,  .608,  .882}\textbf{30.67} & \cellcolor[rgb]{ .765,  .608,  .882}\textbf{54.67} \\
    \bottomrule
    \end{tabular}%
  }
\end{table*}%

\subsection{Additional Analysis of Video Representations}
\label{app:additional_analysis}
\subsubsection{Quantitative Representation Analysis}
\label{app:app_quantitative}


To quantitatively examine the structure of learned video representations, we report six metrics from static and dynamic perspectives, all computed directly in the original representation space.
The static perspective examines the structure of the representation space within and across videos, while the dynamic perspective assesses the continuity and regularity of representation trajectories over time.
For each dataset, we use the same sampled videos and frame indices for all compared methods to ensure a fair comparison.
Unless otherwise specified, we sample up to $M=1000$ videos from each dataset, using all videos for datasets with fewer than $M$ videos, and uniformly select $T=16$ frames from each video.
All metrics are computed directly in the original representation space from dense patch tokens after removing the \texttt{[CLS]} token and register tokens.

\noindent\textbf{Static perspective.}
We examine the structure of the representation space using cycle consistency, effective rank, and separation ratio.
For effective rank and separation ratio, we randomly select $N_p=16$ patch tokens from the center $9\times9$ region of each frame with a fixed seed.
For cycle consistency, we average-pool the full patch-token feature map into a $7\times7$ grid before computing patch-wise transition matrices.

\noindent\textbf{1) Cycle Consistency.}
This metric evaluates whether patch-level correspondences remain stable along a short temporal trajectory.
Given a video, we construct local palindrome paths of three frames,
$t_1\rightarrow t_2\rightarrow t_3\rightarrow t_2\rightarrow t_1$.
For two adjacent frames $t_a$ and $t_b$ on the path, we compute the patch-wise transition matrix by applying a row-wise softmax to the similarity matrix:
\begin{equation}
\bm A_{t_a\rightarrow t_b}(i,j)
=
\frac{
\exp\left(s(\bm z_{t_a}^{i},\bm z_{t_b}^{j})/\tau\right)
}{
\sum_{j'=1}^{P}
\exp\left(s(\bm z_{t_a}^{i},\bm z_{t_b}^{j'})/\tau\right)
},
\end{equation}
where $\bm z_{t}^{i}$ denotes the $i$-th pooled patch token at frame $t$, $P=49$ is the number of pooled patches, $s(\cdot,\cdot)$ denotes feature similarity, and $\tau$ is the temperature.
The transition matrix of the full palindrome cycle is
\begin{equation}
\bm P
=
\bm A_{t_1\rightarrow t_2}
\bm A_{t_2\rightarrow t_3}
\bm A_{t_3\rightarrow t_2}
\bm A_{t_2\rightarrow t_1}.
\end{equation}
We define the cycle-consistency accuracy as the proportion of patches that return to their original indices:
\begin{equation}
\mathrm{Cyc.\ Acc.}
=
\frac{1}{P}
\sum_{i=1}^{P}
\mathbbm{1}
\left[
\arg\max_{j}\bm P(i,j)=i
\right].
\end{equation}
The metric is first computed for each video and then averaged over all sampled videos.
A higher value indicates more reliable patch-level temporal consistency across adjacent frames.

\noindent\textbf{2) Effective Rank.}
Effective rank~\cite{erank} is used to measure whether dense token representations span a rich feature space or collapse to a few dominant dimensions.
For each method, we collect the sampled center-region patch tokens from all evaluated videos and frames, and concatenate them into a feature matrix
\begin{equation}
\bm X =
\left[
\bm z_1,\bm z_2,\ldots,\bm z_B
\right]^{\top}
\in \mathbb{R}^{B\times D},
\end{equation}
where $\bm z_b\in\mathbb{R}^{D}$ denotes a sampled patch token, $B$ is the total number of collected tokens, and $D$ is the feature dimension.
In our evaluation, $B$ is much larger than $D$, so the singular-value spectrum of $\bm X$ reflects how many feature dimensions are effectively used by the dense representations.

Let $\{\sigma_i\}_{i=1}^{r}$ be the singular values of $\bm X$, where $r=\min(B,D)$.
We normalize them as
\begin{equation}
p_i =
\frac{\sigma_i}{\sum_{j=1}^{r}\sigma_j},
\qquad i=1,\ldots,r.
\end{equation}
The normalized effective rank is then defined as
\begin{equation}
\operatorname{Erank}(\bm X)
=
\frac{
\exp\left(
-\sum_{i=1}^{r} p_i \log p_i
\right)
}{
r
}.
\end{equation}
A higher effective rank indicates that the dense token features are distributed over more active dimensions, reflecting stronger token-level representation diversity for better dense feature discriminability.

\noindent\textbf{3) Separation Ratio.}
Separation ratio measures the semantic separability of video representations in the feature space.
A well-structured representation is expected to separate videos with different semantics and cluster adjacent frames from the same video that share similar semantics.
Accordingly, the inter-video distance reflects the separation degree of different semantic representations, while the intra-video distance reflects the compactness of similar semantic representations.

For intra-video distance, we compute the distance between adjacent interior frames within each video.
Given a video $V^{(n)}$, let $\bm z_{t}^{(n)}(i)$ denote the $i$-th sampled patch token at frame $t$.
The intra-video distance is defined as the average patch-token distance over adjacent frame pairs:
\begin{equation}
d_{\mathrm{intra}}^{(n)}(t,t+1)
=
\frac{1}{N_p}
\sum_{i=1}^{N_p}
\left\|
\bm z_{t}^{(n)}(i)
-
\bm z_{t+1}^{(n)}(i)
\right\|_{2}.
\end{equation}
We collect the adjacent-frame distances from all sampled videos and obtain $D_{\mathrm{intra}}^{95}$ using a $95\%$ trimmed mean, which removes the largest $5\%$ distances to reduce the influence of noisy frame transitions.

For inter-video distance, each video is represented by averaging the sampled center-region tokens from three temporal positions, i.e., the frames at $1/4$, $1/2$, and $3/4$ of the sampled sequence.
Let $\bar{\bm z}^{(n)}(i)$ denote the resulting representative token of video $V^{(n)}$.
For two different videos $V^{(u)}$ and $V^{(v)}$, the inter-video distance is defined as the average patch-token distance between their representative tokens:
\begin{equation}
d_{\mathrm{inter}}(u,v)
=
\frac{1}{N_p}
\sum_{i=1}^{N_p}
\left\|
\bar{\bm z}^{(u)}(i)
-
\bar{\bm z}^{(v)}(i)
\right\|_{2}.
\end{equation}
We collect the pairwise distances from all unordered video pairs and obtain $D_{\mathrm{inter}}^{95}$ using the same $95\%$ trimmed-mean aggregation.

The final separation ratio is defined as
\begin{equation}
\mathrm{SepRatio}
=
\frac{
D_{\mathrm{inter}}^{95}
}{
D_{\mathrm{intra}}^{95}
}.
\end{equation}
A higher separation ratio indicates that different semantic representations are more clearly separated, while similar representations within the same video remain more compact, reflecting stronger semantic discriminability of the learned video features.

\noindent\textbf{Dynamic perspective.}
We examine the temporal evolution of frame representations using temporal proximity, local coherence, and trajectory smoothness.
Given a video, let $\bm H_t\in\mathbb{R}^{N\times D}$ denote the encoder patch-token matrix of frame $t$, where $N$ is the number of patch tokens and $D$ is the feature dimension.
We define the frame representation and pairwise distance as
\begin{equation}
    \bm x_t =
    \frac{\operatorname{vec}(\bm H_t)}
    {\|\operatorname{vec}(\bm H_t)\|_2},
    \qquad
    d_{ij}=\|\bm x_i-\bm x_j\|_2,
    \label{eq:dynamic_frame_representation}
\end{equation}
where $\operatorname{vec}(\cdot)$ stacks the patch tokens in their spatial order.
The sequence $\{\bm x_t\}_{t=1}^{T}$ forms the representation trajectory used for the following measurements.

\noindent\textbf{4) Temporal Proximity.}
This metric evaluates whether adjacent frames are relatively close in the representation space compared with nonadjacent frames.
For each frame $t$, we define the adjacent-frame and nonadjacent-frame sets as
\begin{equation}
    \mathcal{N}_t=\{j:|j-t|=1\},
    \qquad
    \mathcal{F}_t=\{j:|j-t|\geq2\},
\end{equation}
where all frame indices belong to $\{1,\ldots,T\}$.
We compute the mean distance from frame $t$ to each set and define temporal proximity as the average difference between these distances:
\begin{equation}
    \mathrm{TP}=
    \frac{1}{T}\sum_{t=1}^{T}
    \left(
    \frac{1}{|\mathcal{F}_t|}
    \sum_{j\in\mathcal{F}_t}d_{tj}
    -
    \frac{1}{|\mathcal{N}_t|}
    \sum_{j\in\mathcal{N}_t}d_{tj}
    \right).
    \label{eq:temporal_proximity}
\end{equation}
A higher value indicates a larger distance gap between nonadjacent and adjacent frames, reflecting stronger temporal locality in the learned representation trajectory.

\noindent\textbf{5) Local Coherence.}
Local coherence measures the temporal correlation between consecutive frame representations, reflecting the persistence of local feature states.
For each video, we first center the frame representations by their video-wise mean:
\begin{equation}
    \bar{\bm x}=\frac{1}{T}\sum_{t=1}^{T}\bm x_t,
    \qquad
    \bm y_t=\bm x_t-\bar{\bm x}.
\end{equation}
We then compute the average inner product between consecutive centered representations and normalize it by their average squared magnitude:
\begin{equation}
    \mathrm{LC}=
    \frac{
    \displaystyle\frac{1}{T-1}
    \sum_{t=1}^{T-1}\bm y_t^{\top}\bm y_{t+1}
    }{
    \displaystyle\frac{1}{T}
    \sum_{t=1}^{T}\|\bm y_t\|_2^2
    }.
    \label{eq:local_coherence}
\end{equation}
The metric is evaluated for videos with numerically nonzero centered variation.
A higher value indicates stronger correlation between consecutive representations relative to their variation within the video, reflecting more coherent local evolution.

\noindent\textbf{6) Trajectory Smoothness.}
Trajectory smoothness characterizes trajectory curvature through the turning angles between successive trajectory segments.
For a temporal interval $h$, we consider the three representation states at $t-h$, $t$, and $t+h$, and define the incoming and outgoing displacements as
\begin{equation}
    \bm a_{t,h}=\bm x_t-\bm x_{t-h},
    \qquad
    \bm b_{t,h}=\bm x_{t+h}-\bm x_t.
\end{equation}
The turning angle at frame $t$ is computed as
\begin{equation}
    \theta_{t,h}=
    \arccos\left(
    \frac{\bm a_{t,h}^{\top}\bm b_{t,h}}
    {\|\bm a_{t,h}\|_2\|\bm b_{t,h}\|_2}
    \right).
    \label{eq:trajectory_turning_angle}
\end{equation}
We consider temporal intervals $h=1,\ldots,\lfloor(T-1)/2\rfloor$.
For each interval, let $\mathcal{I}_h\subseteq\{h+1,\ldots,T-h\}$ contain the centers where both displacement vectors are numerically nonzero for all compared methods.
Let $\mathcal{H}$ denote the set of intervals with at least one valid center.
The trajectory smoothness score is defined as
\begin{equation}
    \mathrm{TS}=
    \frac{1}{|\mathcal{H}|}
    \sum_{h\in\mathcal{H}}
    \left(
    1-\frac{1}{\pi|\mathcal{I}_h|}
    \sum_{t\in\mathcal{I}_h}\theta_{t,h}
    \right).
    \label{eq:trajectory_smoothness}
\end{equation}
The score averages equally over valid centers within each temporal interval and then over valid intervals.
A higher value indicates smaller average turning angles, reflecting smoother representation trajectories across temporal intervals.

Each trajectory measurement is first computed for each video and then averaged over the videos for which it is defined for all compared methods.

\begin{figure*}[htbp]
    \centering
    \begin{subfigure}{\linewidth}
        \centering
        \includegraphics[width=0.49\linewidth]{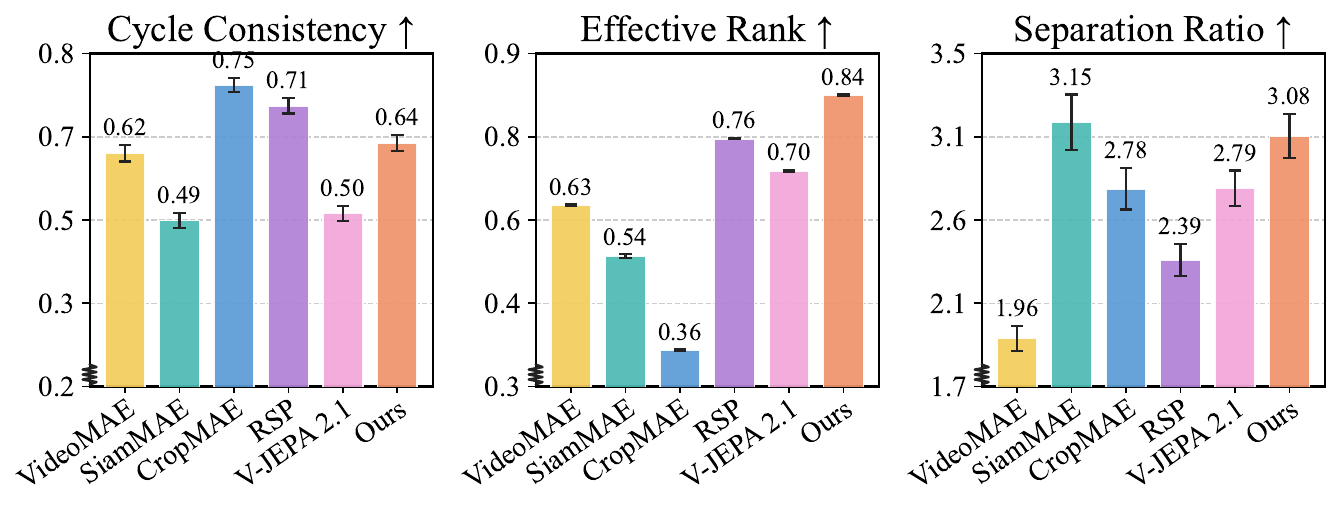}
        \hfill
        \includegraphics[width=0.49\linewidth]{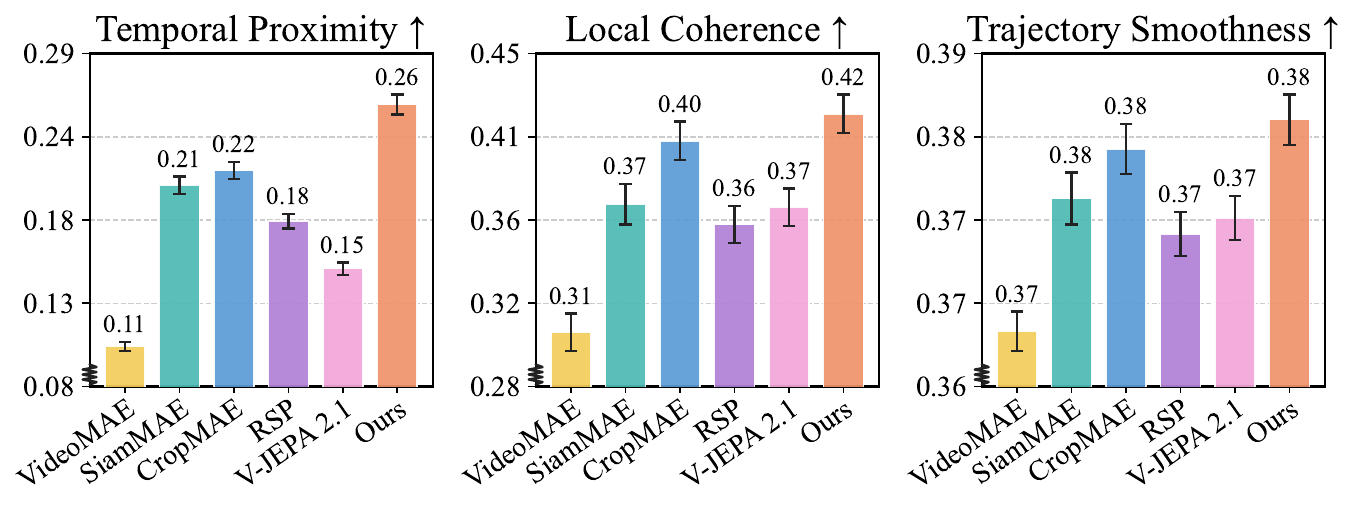}
        \caption{Quantitative comparison of representation quality on Kinetics-400 from static (left) and dynamic (right) perspectives.}
        \label{fig:paper_bars_k400}
    \end{subfigure}
    \par\vspace{6pt}
    \begin{subfigure}{\linewidth}
        \centering
        \includegraphics[width=0.49\linewidth]{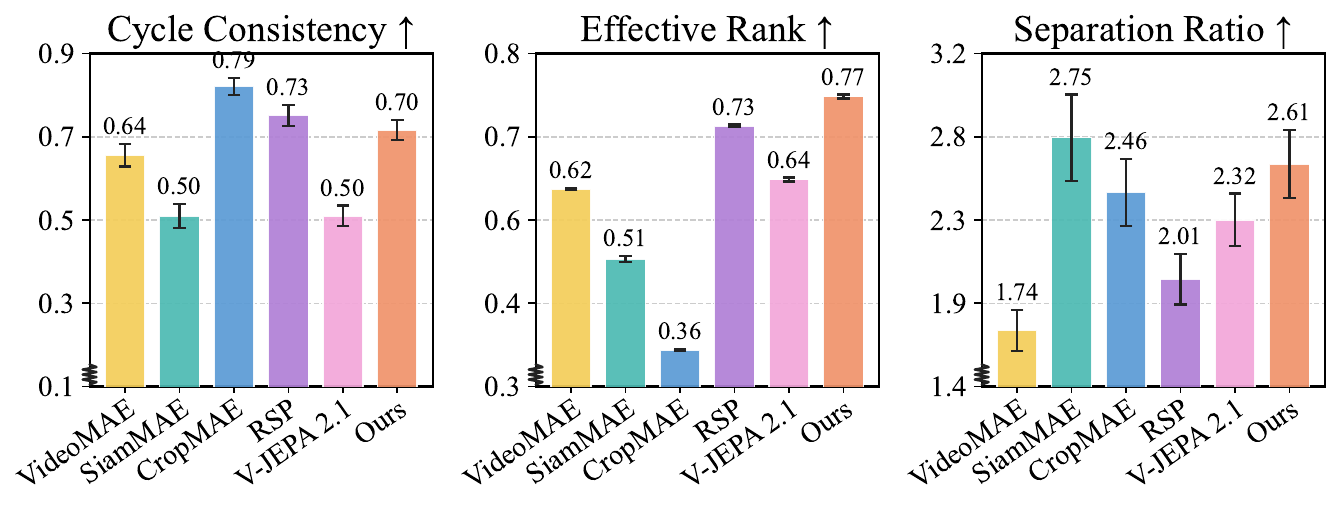}
        \hfill
        \includegraphics[width=0.49\linewidth]{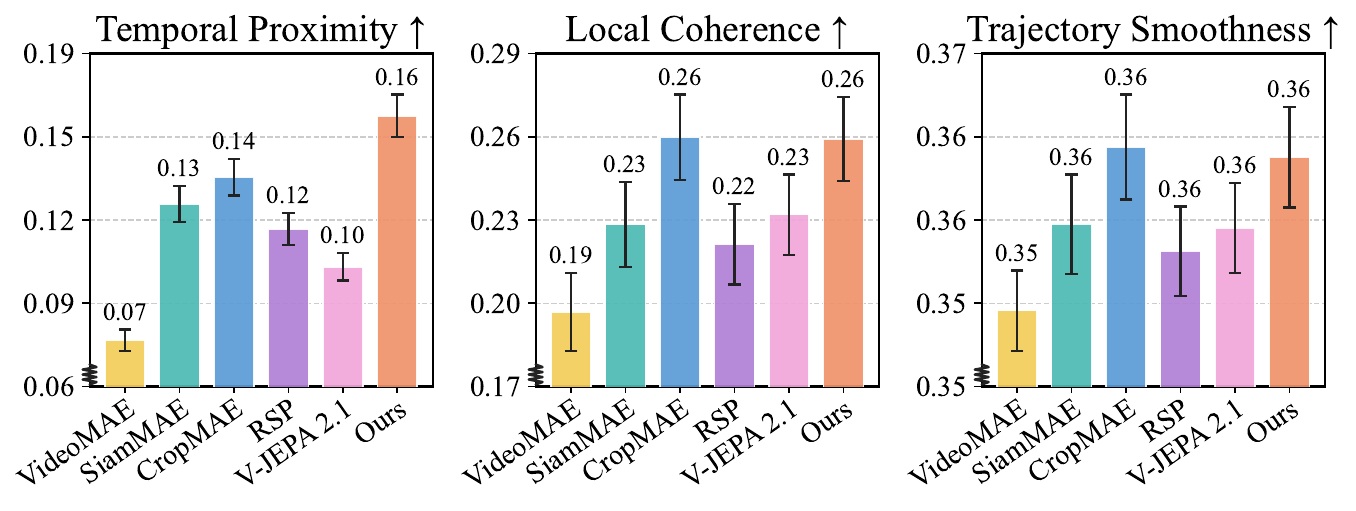}
        \caption{Quantitative comparison of representation quality on VIP from static (left) and dynamic (right) perspectives.}
        \label{fig:paper_bars_vip}
    \end{subfigure}
    \par\vspace{6pt}
    \begin{subfigure}{\linewidth}
        \centering
        \includegraphics[width=0.49\linewidth]{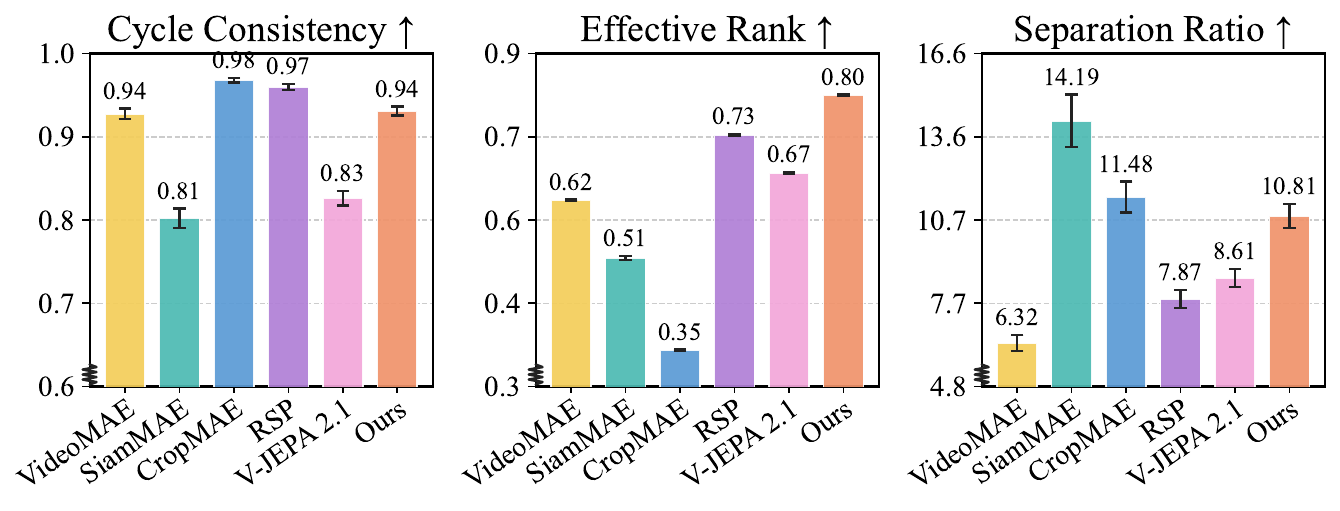}
        \hfill
        \includegraphics[width=0.49\linewidth]{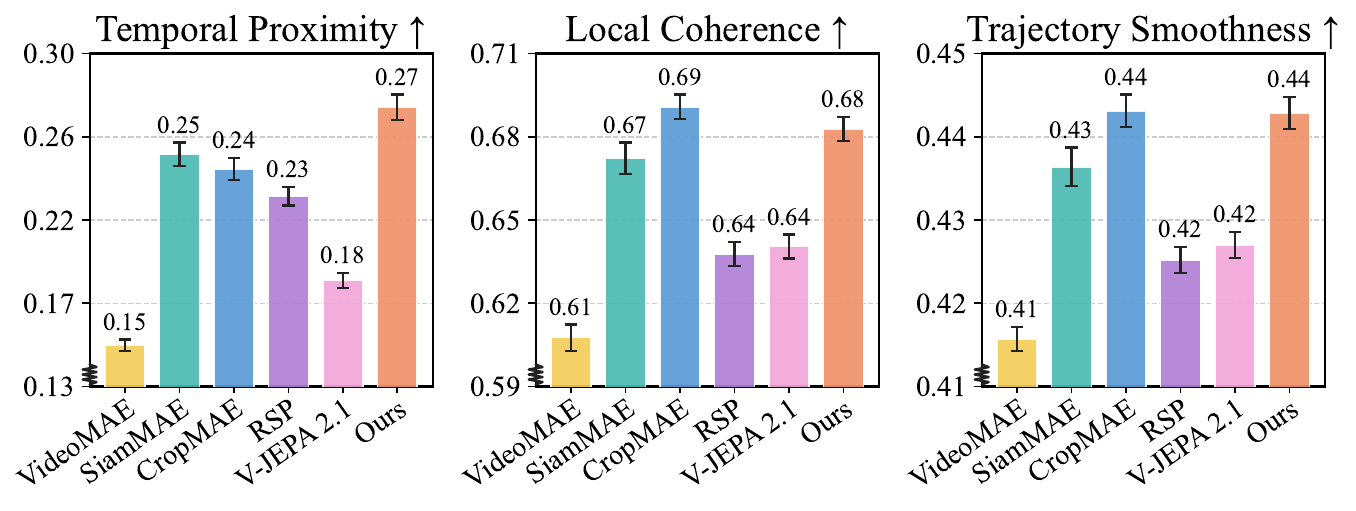}
        \caption{Quantitative comparison of representation quality on JHMDB from static (left) and dynamic (right) perspectives.}
        \label{fig:paper_bars_jhmdb}
    \end{subfigure}
    \par\vspace{6pt}
    \begin{subfigure}{\linewidth}
        \centering
        \includegraphics[width=0.49\linewidth]{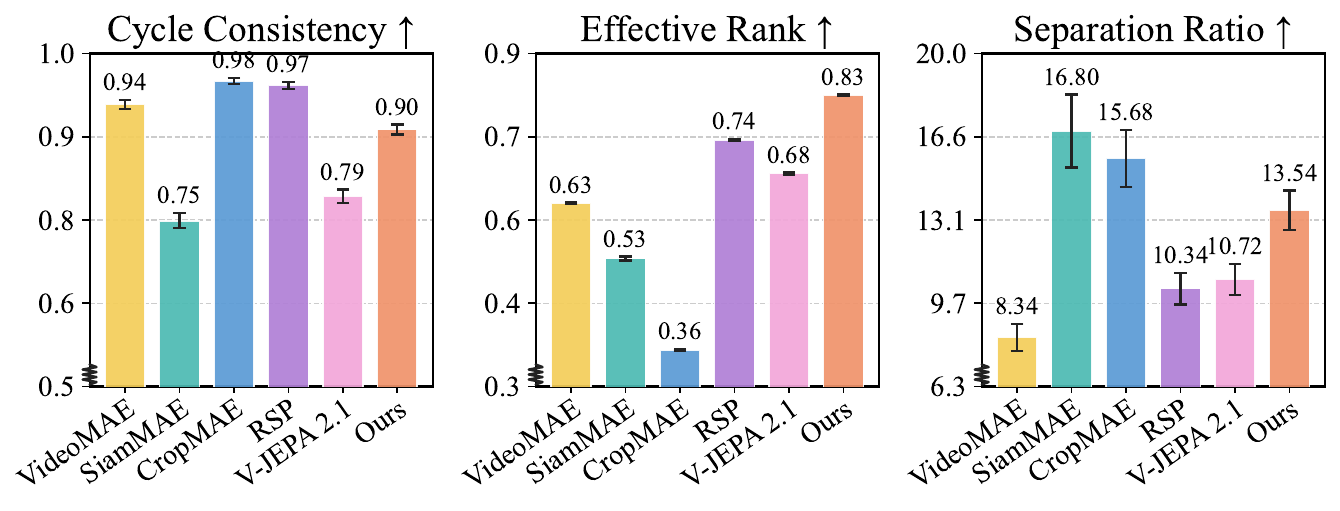}
        \hfill
        \includegraphics[width=0.49\linewidth]{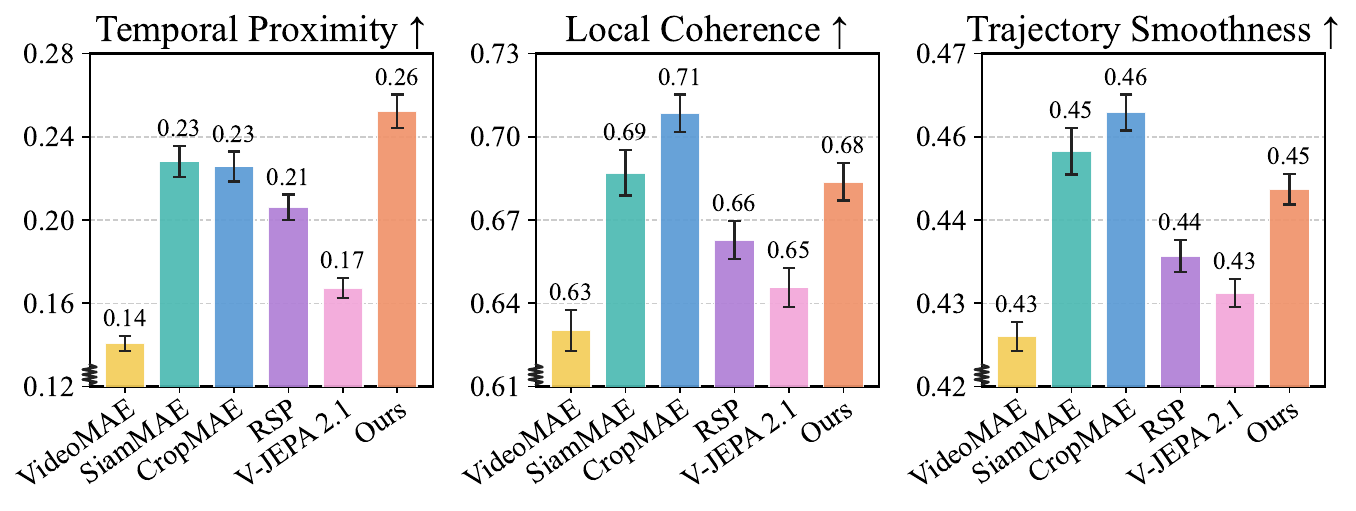}
        \caption{Quantitative comparison of representation quality on THUMOS14 from static (left) and dynamic (right) perspectives.}
        \label{fig:paper_bars_thumos14}
    \end{subfigure}
    \par\vspace{6pt}
    \begin{subfigure}{\linewidth}
        \centering
        \includegraphics[width=0.49\linewidth]{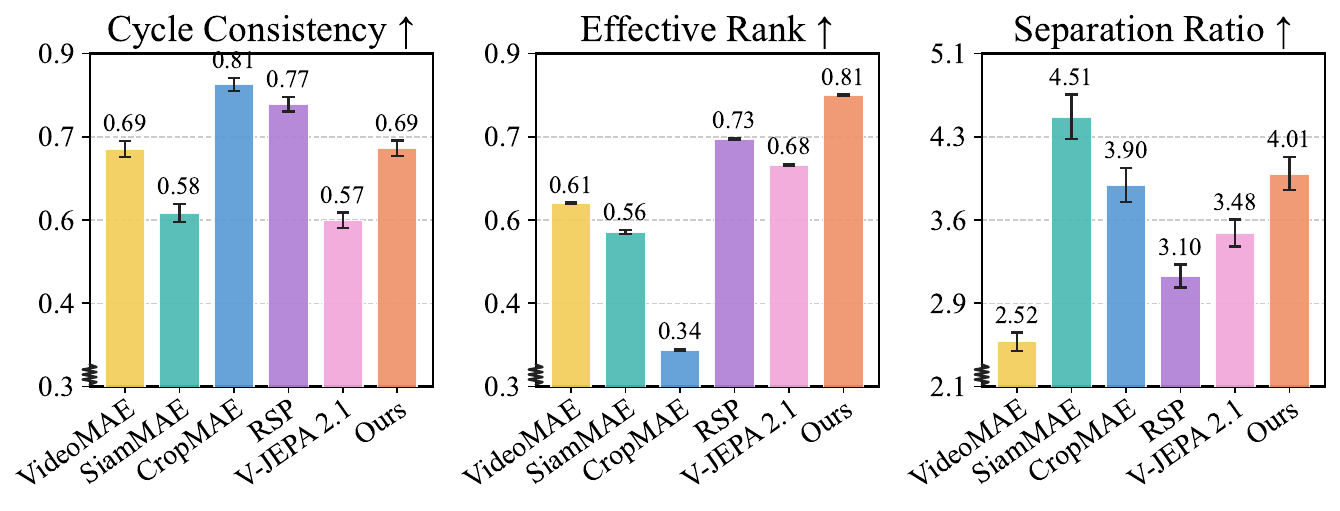}
        \hfill
        \includegraphics[width=0.49\linewidth]{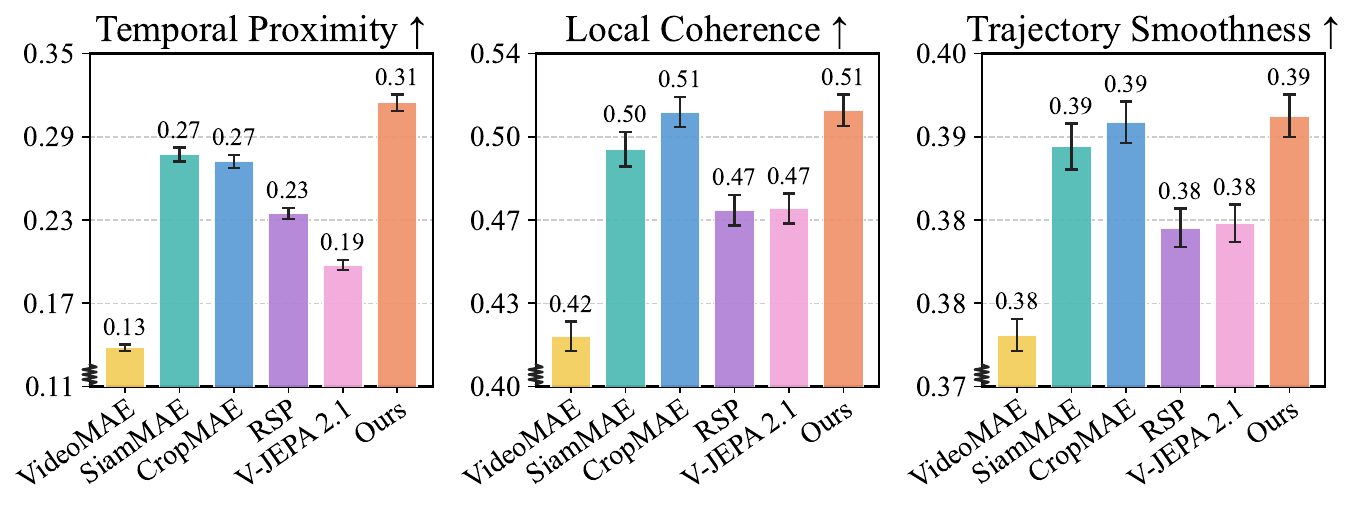}
        \caption{Quantitative comparison of representation quality on SSV2 from static (left) and dynamic (right) perspectives.}
        \label{fig:paper_bars_ssv2_val}
    \end{subfigure}
    \caption{Additional quantitative comparison of representation quality from static and dynamic perspectives across five datasets. Results on Kinetics-400 and SSV2 are evaluated on their validation sets. All metrics are computed in the original representation space. Higher values are preferred for all metrics.}
    \label{fig:representation_metrics_app}
\end{figure*}

Subsequently, we provide a quantitative analysis of the learned video representations in Fig. {\color{blue}12} from static and dynamic perspectives.

\noindent\textbf{1) Static perspective.}
\begin{itemize}
    \item \textbf{Temporal cycle consistency.}
    \methodname{} maintains high cycle-consistency accuracy across datasets, indicating reliable patch-level temporal correspondence.
    Since this metric focuses on short-cycle return accuracy, methods that emphasize local invariance may also obtain high scores.
    \methodname{} uses PRM to stabilize patch correspondence while introducing temporal variation modeling to regularize frame-wise changes.
    The competitive results suggest that modeling these changes remains compatible with reliable local alignment.

    \item \textbf{Effective rank.}
    \methodname{} consistently achieves the highest effective rank, showing that its dense tokens use a richer feature space and are less concentrated in a few dominant dimensions.
    This suggests that diverse token-level responses are retained under the temporal variation constraint.
    Such diversity provides a richer representation basis for distinguishing local visual content in dense prediction tasks.

    \item \textbf{Separation ratio.}
    \methodname{} obtains competitive separation ratios across datasets, indicating strong inter-video separation relative to intra-video variation.
    This suggests that the learned representations accommodate frame-wise changes while preserving distinctions between videos.
    Together with the effective rank results, these findings support a feature space that retains both token-level diversity and video-level organization.
\end{itemize}

\noindent\textbf{2) Dynamic perspective.}
\begin{itemize}
    \item \textbf{Temporal proximity.}
    \methodname{} achieves the highest temporal proximity across the evaluated datasets, indicating a larger average distance gap between nonadjacent and adjacent frames.
    This shows that the learned representations preserve temporal locality while retaining distinguishable states over longer temporal intervals.
    The result is consistent with temporal variation modeling preserving local continuity while accommodating changes across frames.

    \item \textbf{Local coherence.}
    \methodname{} achieves leading or competitive local coherence across datasets, indicating persistent temporal correlation between consecutive representation states.
    Centering by the video-wise mean focuses this analysis on correlations in the temporally varying feature states.
    The results therefore support coherent local evolution of the learned representations, consistent with the temporal variation constraint regularizing relationships among corresponding representations.

    \item \textbf{Trajectory smoothness.}
    \methodname{} achieves leading or competitive trajectory smoothness across datasets.
    These results indicate relatively small average turning angles across temporal intervals, supporting smoother trajectory structure in the original representation space.
    This observation is consistent with the second-order temporal variation constraint regularizing the evolution of aligned representations.
    Together with temporal proximity and local coherence, the smoothness results provide quantitative evidence of coherent and smooth representation trajectories.
\end{itemize}

Overall, these results indicate that \methodname{} maintains a well-organized static representation structure while learning more coherent and smoother dynamic trajectories.

\subsubsection{Visualization of Cross-attention Maps in PRM}

\begin{figure*}
  \centering
    \includegraphics[width=0.94\linewidth]{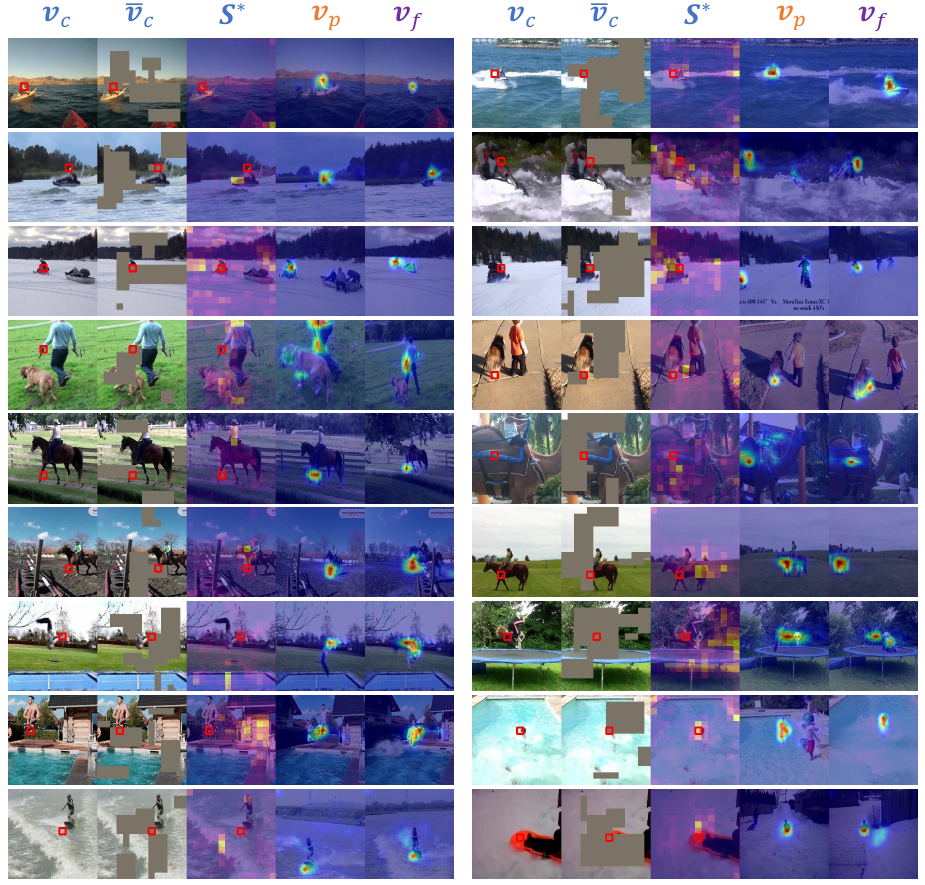}
     \caption{Additional cross-attention and confidence visualization in PRM. We visualize the current frame $\bm v_c$, the masked current frame $\bar{\bm v}_c$, the joint confidence map $\bm S^{\ast}$, and the cross-attention maps from masked current patches to the past and future frames $\bm v_p$ and $\bm v_f$. Best viewed in color.}
    \label{fig:vis_heatmaps_app}
    \vspace{-8pt}
\end{figure*}

To further examine the temporal correspondence established by the patch registration module (PRM), we visualize the cross-attention maps from queried masked patches in the current frame to the auxiliary past and future frames. As shown in \Fgref{fig:vis_heatmaps_app}, the attention responses consistently focus on the corresponding key parts of foreground targets across frames, such as body parts, object regions, and moving instances. This indicates that PRM can identify salient foreground components and establish stable frame-to-frame correspondence for the queried visual content. Such correspondence remains effective even when the queried region is largely or completely masked in the current frame, showing that PRM can infer corresponding regions from temporal references. In addition, the responses are often well localized around object boundaries and contours, suggesting that the established correspondence is spatially precise rather than only coarsely object-level.

The joint confidence map $\bm{S}^{\ast}$ further highlights regions with reliable cross-frame matches. High-confidence responses are mainly concentrated on foreground parts with clear temporal correspondence, while ambiguous background regions and visually uncertain areas produce weaker responses. These observations show that PRM enhances reliable foreground correspondences while suppressing unstable matches from background or ambiguous regions, providing a high-confidence registration basis for latent-space self-distillation alignment and subsequent temporal variation modeling.

\begin{figure*}
  \centering
  \includegraphics[width=0.94\linewidth]{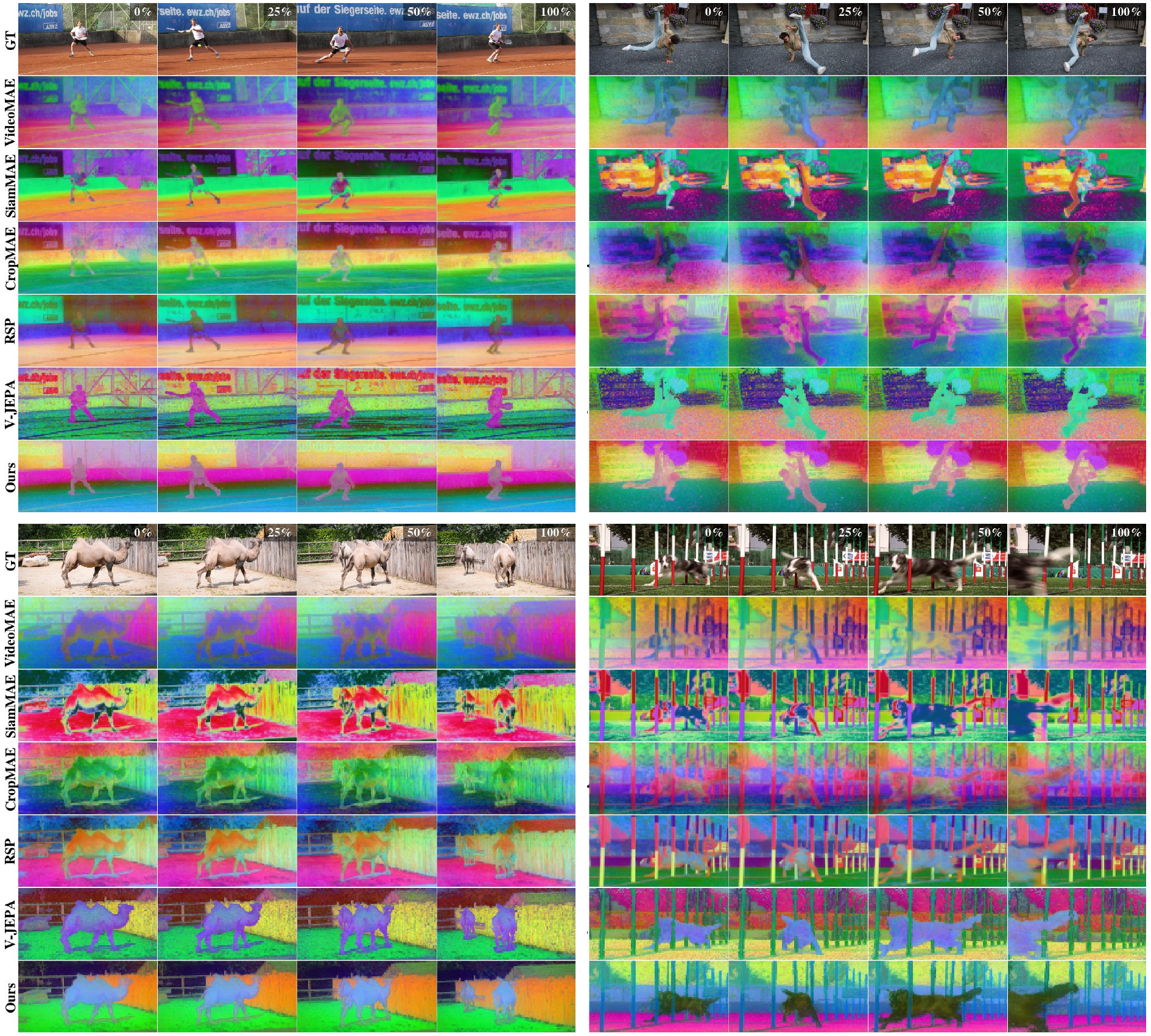}
  \caption{PCA visualization of patch representations on DAVIS-2017. For each video, PCA is fitted using all patch tokens from the entire video, and the first three principal components are mapped to RGB channels. The columns show frames sampled at $0\%$, $25\%$, $50\%$, and $100\%$ of the video. Best viewed in color.}
  \label{fig:vis_pca_app}
\end{figure*}

\subsubsection{PCA Visualization of Patch Representations}

We visualize patch-level representations with PCA to compare the dense feature structures learned by different methods. For each video, we extract last-layer patch tokens from the ViT-B/16 models at 960p resolution and fit PCA to all patch tokens from the entire video.
The first three principal components are min-max normalized and mapped to RGB channels. All methods follow the same resolution, backbone size, and PCA protocol on DAVIS-2017~\cite{DAVIS17}. Since the PCA basis is fitted separately for each method and video, cross-method comparisons focus on whether semantically corresponding regions form coherent and temporally stable feature distributions, rather than on the absolute RGB values.

As shown in \Fgref{fig:vis_pca_app}, we obtain the following observations.
\begin{itemize}
    \item \textbf{Clearer semantic structure.}
    Several masked reconstruction baselines exhibit less distinguishable PCA responses, where different semantic regions tend to collapse into similar responses or are dominated by local appearance variations. In contrast, \methodname{} forms more coherent regions for semantically related patches, indicating a clearer frame-level representation structure.

    \item \textbf{Stronger patch-level discrimination.}
    The compared baselines often produce noisier PCA maps, with fragmented responses around object boundaries and background regions. This indicates that their patch representations are less well separated in local feature space, especially between foreground structures and the surrounding context. \methodname{} yields cleaner and more separable responses between the foreground target and the surrounding scene, suggesting stronger dense semantic discrimination.

    \item \textbf{More stable temporal correspondence.}
    Across different temporal positions, the foreground regions of \methodname{} maintain more consistent PCA responses under changes in pose, scale, and location. This indicates more reliable correspondence of patch representations over time, which is consistent with the proposed temporal variation modeling and is beneficial for dense propagation across time.
\end{itemize}

These visualizations indicate that \methodname{} learns patch representations with clearer semantic organization and more stable temporal correspondence, supporting the role of modeling regular latent variations across video frames.

\subsubsection{Comparison of Representation Trajectories}

\begin{figure*}
  \centering
    \includegraphics[width=0.94\linewidth]{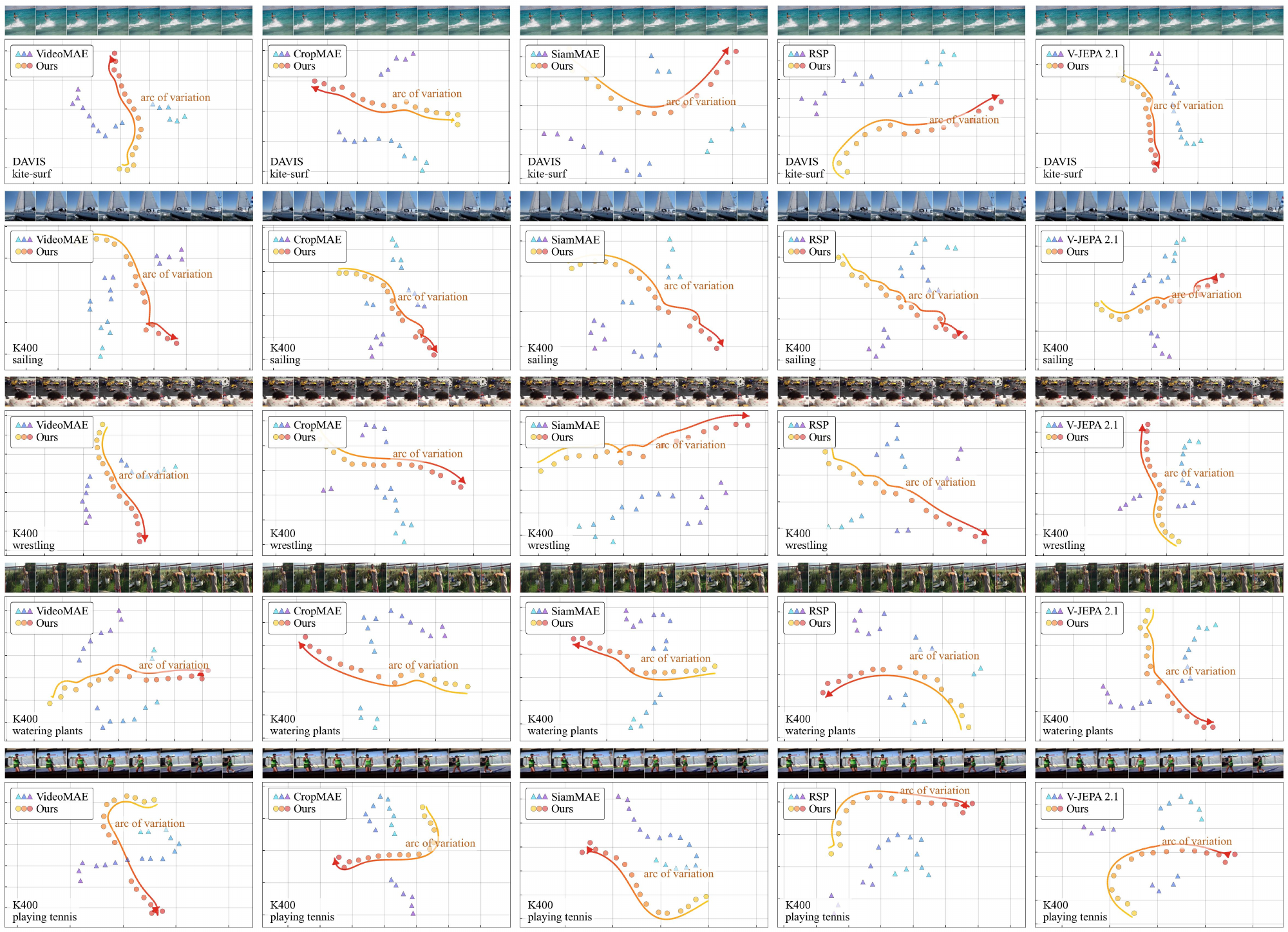}
    \caption{Additional t-SNE visualization of representation trajectories for videos. Each row shows the same video across different methods, with temporal order indicated by color gradients and arrows. Best viewed in color.}
    \label{fig:vis_arc_app}
    \vspace{-8pt}
\end{figure*}

We provide additional t-SNE visualizations of frame-level representation trajectories in \Fgref{fig:vis_arc_app}.
The visualization includes both DAVIS videos with relatively clear object-centric scenes and Kinetics-400 videos with larger motion, more diverse categories, and more complex scene changes.

Compared with previous methods, \methodname{} produces more continuous and ordered trajectories across frames.
On DAVIS, the frame embeddings of \methodname{} form smoother paths, indicating stable temporal organization under relatively clean object-centric videos.
On Kinetics-400, although the visual appearance and motion patterns are more complex, \methodname{} still shows more coherent trajectory structures than the compared baselines.
This suggests that the proposed variation-aware temporal modeling helps organize frame representations according to continuous visual-state changes, rather than producing scattered frame-level embeddings.
Together with the patch registration module, this enables \methodname{} to maintain reliable temporal correspondence while modeling structured representation trajectories in the latent space.

\begin{table*}[t]
\centering
\caption{Direct correspondence evaluation on DAVIS. Matching Accuracy measures instance consistency at temporal gap $g$. Computational costs are reported for the pretrained PRM module from our \methodname{} backbone.}
\label{tab:prm_correspondence}
\vspace{-3pt}
\setlength{\tabcolsep}{5pt}
\renewcommand{\arraystretch}{1.2}
\resizebox{0.95\linewidth}{!}{%
\begin{tabular}{c|l|cccc|ccc}
\toprule
\multirow{2}{*}{Type} & \multirow{2}{*}{Method} & \multicolumn{4}{c|}{Matching Accuracy (\%) $\uparrow$} & \multirow{2}{*}{\makecell{Latency\\(ms/clip) $\downarrow$}} & \multirow{2}{*}{\makecell{Peak Memory\\(MB) $\downarrow$}} & \multirow{2}{*}{\makecell{Parameters\\(M) $\downarrow$}} \\
& & $g=1$ & $g=3$ & $g=5$ & $g=8$ & & & \\
\midrule
\multirow{3}{*}{\makecell{Optical flow\\(supervised)}} & RAFT-Large~\cite{raft} & 90.4 & 84.9 & 78.2 & 72.2 & $70.2 \pm 1.2$ & 115.7 & 5.3 \\
& GMFlow~\cite{gmflow} & 89.2 & 85.0 & 81.3 & 74.1 & $47.3 \pm 0.5$ & 331.8 & 4.7 \\
& SEA-RAFT~\cite{sea_raft} & 91.0 & 85.3 & 82.4 & 76.1 & $48.6 \pm 1.4$ & 156.6 & 19.7 \\
\midrule
\multirow{3}{*}{\makecell{Point tracking\\(supervised)}} & PIPs++~\cite{pips++} & 91.9 & 80.3 & 75.9 & 70.7 & $273.9 \pm 0.3$ & 873.8 & 17.6 \\
& TAPIR~\cite{tapir} & 92.2 & 88.6 & 87.7 & 83.2 & $74.4 \pm 0.5$ & 287.5 & 31.1 \\
& CoTracker~\cite{cotracker} & 90.8 & 86.3 & 83.8 & 80.7 & $156.0 \pm 1.9$ & 3185.1 & 24.1 \\
\midrule
\makecell{Self-supervised\\representation learning} & \textbf{PRM} & 88.9 & 84.0 & 83.4 & 80.6 & $2.9 \pm 0.3$ & 64.2 & 5.3 \\
\bottomrule
\end{tabular}%
}
\end{table*}

\begin{figure*}[htbp]
    \centering
    \begin{minipage}{0.95\linewidth}
        \centering
        \includegraphics[width=\linewidth]{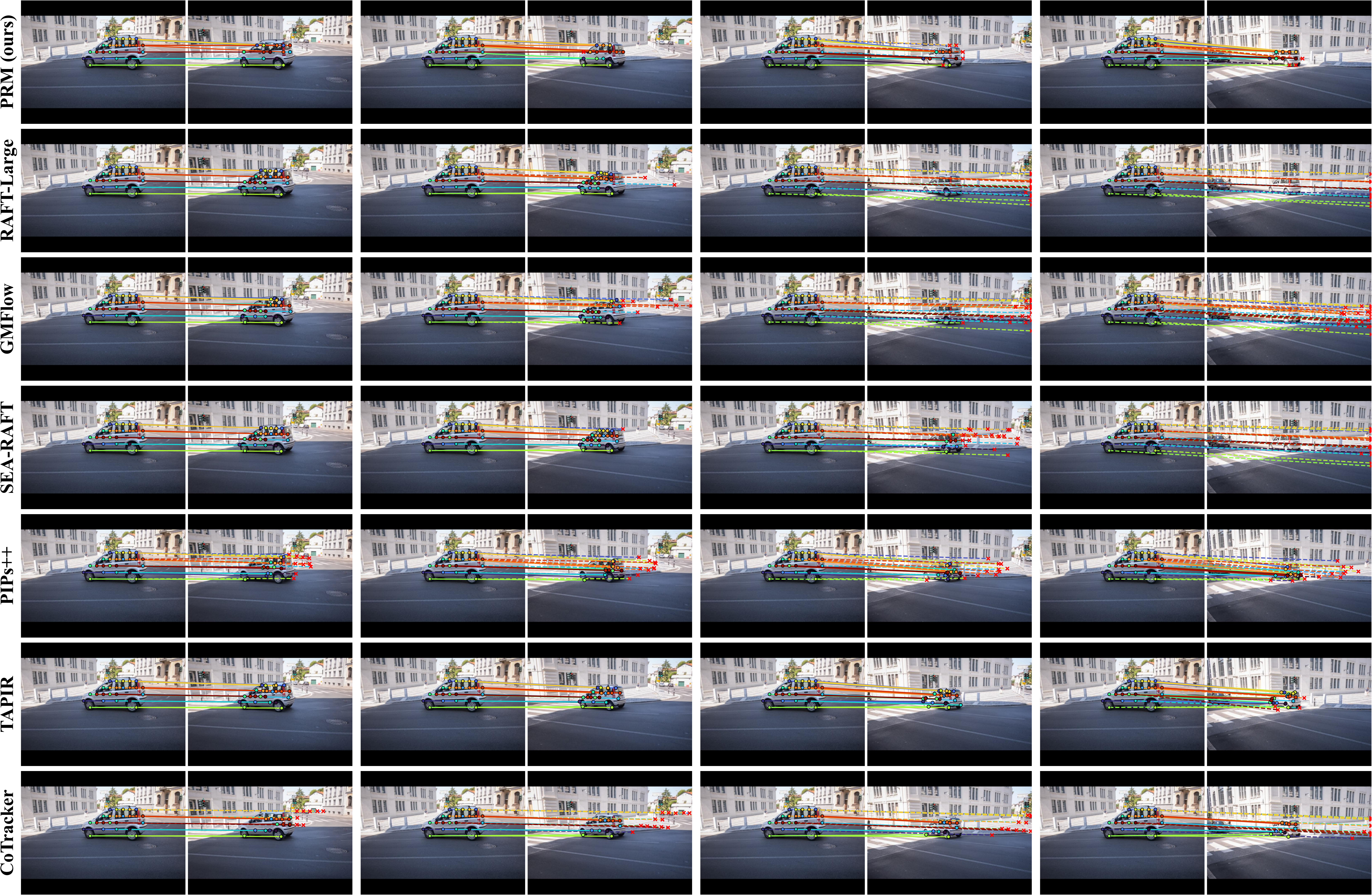}
    \end{minipage}
    \par\vspace{2mm}
    \begin{minipage}{0.95\linewidth}
        \centering
        \includegraphics[width=\linewidth]{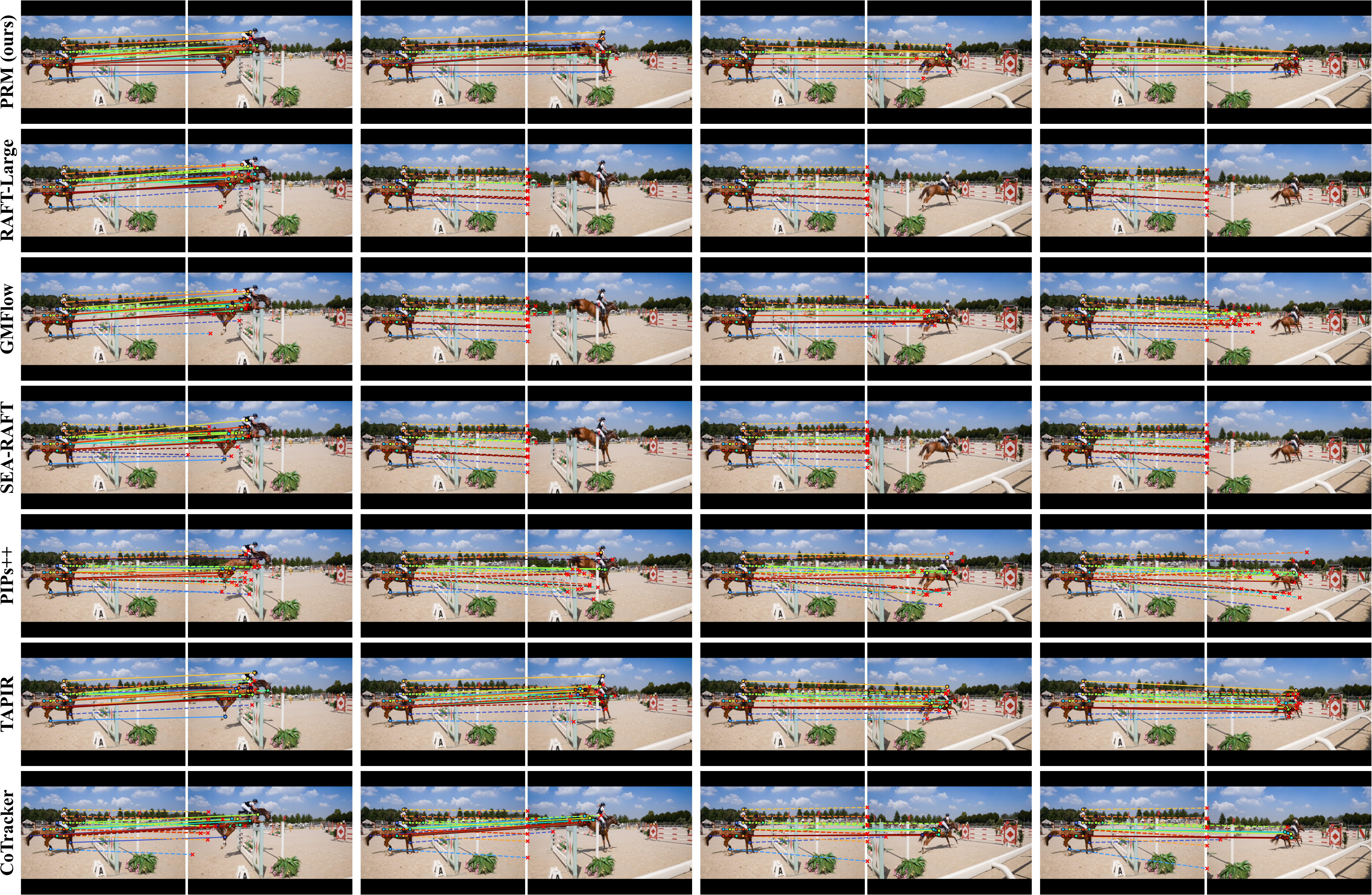}
    \end{minipage}
    \par\vspace{2pt}
    \caption{Sparse correspondences on DAVIS using five frames uniformly sampled across each full video. Rows compare methods, and columns show first-to-target matches at increasing temporal gaps. Solid lines indicate instance-consistent matches, while dashed lines and red crosses indicate errors. Out-of-frame endpoints are clipped for display.}
    \label{fig:prm_correspondence}
\end{figure*}

\subsection{Quality of Learned Temporal Correspondences}
\label{sec:temporal_correspondence_quality}

\subsubsection{Sparse Correspondence Evaluation}
\label{sec:sparse_correspondence}

\noindent\textbf{Evaluation protocol.}
We evaluate PRM correspondence quality on DAVIS-2017 against three optical flow methods (RAFT-Large~\cite{raft}, GMFlow~\cite{gmflow}, and SEA-RAFT~\cite{sea_raft}) and three point tracking methods (PIPs++~\cite{pips++}, TAPIR~\cite{tapir}, and CoTracker~\cite{cotracker}). These baselines use task-specific supervision, whereas PRM learns correspondences through self-supervised video representation learning without flow or tracking annotations. For each video, we select three starting frames near the beginning, middle, and end. For each starting frame $t$, we sample $\{t,t+1,t+3,t+5,t+8\}$ and evaluate correspondences from $t$ to the remaining four frames. All methods use the centers of the same $16\times16$ image patches as source queries. PRM predicts each endpoint by selecting the target patch with the highest correspondence weight. All methods are evaluated without fine-tuning on DAVIS.

\noindent\textbf{Matching Accuracy.}
Let $M_s$ and $M_t$ denote the source and target instance masks, and let $\hat{\mathbf{y}}_i$ be the predicted endpoint of source query $\mathbf{x}_i$. The evaluation set $\mathcal{Q}$ contains source queries belonging to foreground instances present in both frames. Matching Accuracy measures the percentage of endpoints that belong to the same instance as their source queries:
\begin{equation}
\mathrm{MA}(s,t)=\frac{100}{|\mathcal{Q}|}\sum_{i\in\mathcal{Q}}\mathbb{I}\!\left[M_t(\hat{\mathbf{y}}_i)=M_s(\mathbf{x}_i)\right].
\label{eq:prm_matching_accuracy}
\end{equation}
Invalid or out-of-frame endpoints, as well as predictions on background or a different instance, are counted as errors. For each temporal gap, we pool correct and eligible query counts across the three clips within each video, then average the resulting accuracies equally over videos. Ground-truth masks are used only for evaluation. This metric directly assesses whether the predicted correspondences preserve object-instance identity across frames.

\noindent\textbf{Quantitative results.}
As shown in \Tbref{tab:prm_correspondence}, PRM achieves instance-level matching accuracy broadly comparable to that of the supervised optical flow and point tracking methods without manual correspondence annotations.
At gap 8, PRM achieves 80.6\% accuracy, exceeding all three optical flow methods and PIPs++, closely matching CoTracker (80.7\%), and remaining below TAPIR (83.2\%).
PRM identifies corresponding regions through cross-attention based on the similarity of learned visual features.
Joint self-supervised training of the encoder and PRM encourages these feature-based associations to support consistent representations across frames, providing a matching signal without annotated flow fields or point trajectories.

\noindent\textbf{Efficiency analysis.}
We measure correspondence prediction costs using the same five-frame clip and source queries, with FP32 and batch size one. Latency is averaged over 200 measurements after 30 warm-up runs. PRM costs refer to the module and exclude the backbone. As reported in \Tbref{tab:prm_correspondence}, PRM requires only 2.9 ms per clip, 5.3 M parameters, and 64.2 MB of peak memory. Its module latency and peak memory use are lower than those of the compared pipelines, with fewer parameters than the point trackers. These results demonstrate that a lightweight self-supervised module can achieve competitive correspondence quality without a separate supervised tracking network.

\noindent\textbf{Qualitative results.}
\Fgref{fig:prm_correspondence} visualizes correspondences from the first to the remaining four frames of five uniformly sampled frames spanning each video. All methods use the same displayed source queries. PRM maintains object-consistent matches across distant frames despite substantial motion and appearance changes, while several baselines exhibit more background or out-of-frame predictions. Together with the quantitative results in \Tbref{tab:prm_correspondence}, these observations support the effectiveness of the self-supervised correspondences learned by PRM for temporal variation modeling.

\subsubsection{Spatial Resolution and Fine-Grained Correspondence}
\label{sec:attention_tracking}

\noindent\textbf{Effect of input resolution.}
Following the benchmark selection and multi-resolution evaluation strategy of DINOv3~\cite{DINOv3}, we evaluate frozen representations through non-parametric label propagation on DAVIS-2017~\cite{DAVIS17}, YouTube-VOS~\cite{youtube_vos}, and MOSE~\cite{mose}.
The Small/Medium/Large settings use input resolutions of $480\times880$, $960\times1760$, and $1440\times2640$ pixels, respectively.
With a fixed $16\times16$ patch size, these settings yield 30/60/90 native tokens along the short side.
A larger input therefore provides finer spatial sampling, with each patch covering a smaller region of the original scene.
Predictions are evaluated at the original annotation resolution.
All methods use the same fixed label propagation settings across datasets and input resolutions.

As shown in \Tbref{tab:multi_resolution_propagation}, \methodname{} outperforms VideoMAE, SiamMAE, CropMAE, and RSP across all datasets and resolutions under this evaluation protocol.
From Small to Medium, its $\mathcal{J}\&\mathcal{F}$ improves by 4.9, 4.2, and 4.7 percentage points on DAVIS-2017, YouTube-VOS, and MOSE, respectively.
The compared masked reconstruction methods exhibit limited gains or slight declines from Medium to Large in some settings, possibly reflecting remaining spatial granularity constraints with a fixed patch size.
Nevertheless, \methodname{} maintains higher performance than the compared methods at all evaluated resolutions, supporting the adequacy of the adopted token resolution for correspondence-based segmentation.
In principle, reducing the patch size would provide finer spatial sampling and could enable more precise matching.

\begin{figure}[!t]
    \centering
    \includegraphics[width=0.95\linewidth]{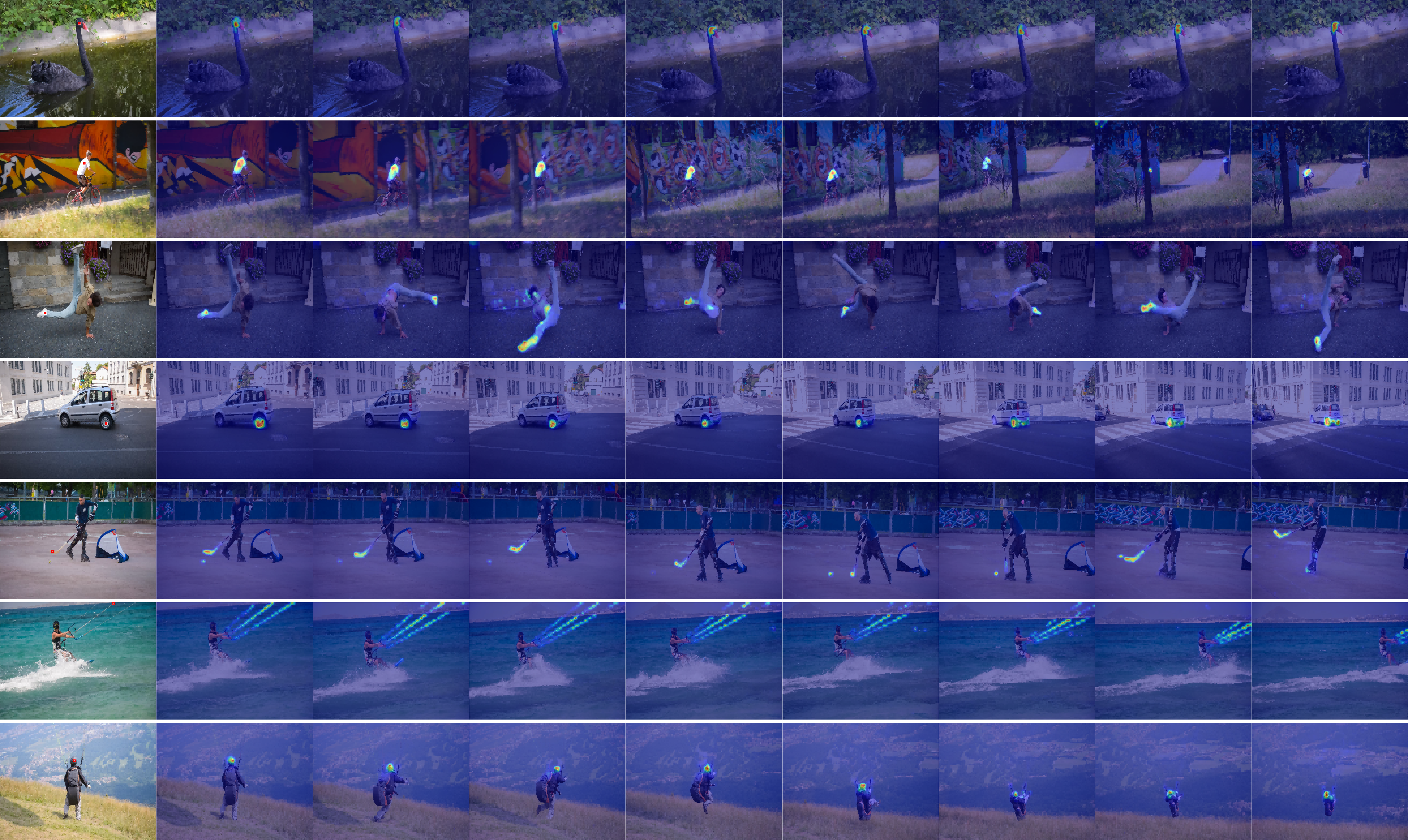}
    \par\vspace{2pt}
    \caption{PRM cross-attention on DAVIS. Each row shows a first-frame query patch marked by a red box, followed by attention maps on eight uniformly sampled subsequent frames. Warmer colors indicate higher attention responses.}
    \label{fig:attention_tracking}
\end{figure}

\noindent\textbf{Fine-grained correspondence visualization.}
\Fgref{fig:attention_tracking} visualizes PRM cross-attention on DAVIS using $1024\times768$ inputs and $16\times16$ patches.
Each row shows a fixed query patch in the first frame and its cross-attention to eight uniformly sampled subsequent frames.
The responses remain localized around corresponding object parts, including car wheels and human heads, and follow thin rope structures across frames.
These examples demonstrate spatially focused correspondences for small regions and thin structures under changes in position, pose, and scale.

Taken together, the segmentation results and PRM visualizations indicate that the adopted token resolution provides useful spatial detail for fine-grained patch-level correspondence.
These results suggest that PRM can generally maintain localized correspondences for small object regions and thin structures in common tracking scenarios.

\begin{table}[htbp]
    \centering
    \caption{Video segmentation results measured by $\mathcal{J}\&\mathcal{F}$ (\%) at different input resolutions. Small/Medium/Large denote input resolutions of $480\times880$, $960\times1760$, and $1440\times2640$ pixels, respectively. All models are pre-trained on Kinetics-400~\cite{Kinetics}, with their backbone architectures and pre-training epochs listed. The best and second-best results are highlighted in \textbf{bold} and \underline{underlined}, respectively.}
    \vspace{-3pt}
    \label{tab:multi_resolution_propagation}
    \setlength{\tabcolsep}{5pt}
    \renewcommand{\arraystretch}{1.1}
    \resizebox{0.95\linewidth}{!}{%
    \begin{tabular}{c|cc|ccc|ccc|ccc}
        \toprule
        \multirow{2}{*}{Method}
        & \multirow{2}{*}{Backbone}
        & \multirow{2}{*}{Epochs}
        & \multicolumn{3}{c|}{DAVIS-2017~\cite{DAVIS17}}
        & \multicolumn{3}{c|}{YouTube-VOS~\cite{youtube_vos}}
        & \multicolumn{3}{c}{MOSE~\cite{mose}} \\
        & &
        & Small & Medium & Large
        & Small & Medium & Large
        & Small & Medium & Large \\
        \midrule
        VideoMAE
        & ViT-L/16 & 1600
        & 45.0 & 40.2 & 39.3
        & 53.0 & 47.8 & 44.3
        & 30.5 & 28.2 & 25.3 \\
        SiamMAE
        & ViT-B/16 & 400
        & \underline{60.9} & \underline{67.6} & \underline{67.9}
        & 64.5 & 67.0 & 66.2
        & 37.7 & 41.6 & 41.4 \\
        CropMAE
        & ViT-B/16 & 400
        & 57.8 & 66.1 & 67.2
        & 61.9 & 66.0 & 65.7
        & 36.6 & 41.2 & 41.5 \\
        RSP
        & ViT-B/16 & 400
        & 60.5 & 65.1 & 65.4
        & \underline{65.1} & \underline{67.7} & \underline{66.9}
        & \underline{39.0} & \underline{42.6} & \underline{42.3} \\
        \methodnameb{}
        & ViT-B/16 & 200
        & \textbf{67.0} & \textbf{71.9} & \textbf{70.7}
        & \textbf{67.5} & \textbf{71.7} & \textbf{69.7}
        & \textbf{41.2} & \textbf{45.9} & \textbf{46.1} \\
        \bottomrule
    \end{tabular}%
    }
\vspace{-8pt}
\end{table}

\subsection{Effects of Temporal Variation Modeling}

\subsubsection{Temporal Regularity Before and After Projection}
\label{sec:projection_trajectory_analysis}

\noindent\textbf{Evaluation protocol.}
We evaluate encoder representations before projection ($\bm A$) and representations after projection ($\bm B$) from the same complete model.
For comparison, we extract encoder representations from a model without projection ($\bm C$).
The two models use otherwise matched training settings.
We evaluate the three representations on DAVIS, VIP, JHMDB, THUMOS14, K400 validation, and SSV2 validation, using identical sampled frames and preprocessing.

We introduce Global Path Efficiency (GPE) to evaluate temporal coherence before and after projection.
It is computed directly in each representation space, with higher values indicating stronger directional coherence of temporal feature changes over the sampled sequence.
For a video with $T$ sampled frames and $P$ patches per frame, let $\bm f_{t,j}$ denote the feature at frame $t$ and fixed spatial patch position $j$.
The metric compares the total endpoint displacement with the accumulated path length across all patches:
\begin{equation}
    \mathrm{GPE}=
    \frac{
    \displaystyle\sum_{j=1}^{P}\|\bm f_{T,j}-\bm f_{1,j}\|_2
    }{
    \displaystyle\sum_{j=1}^{P}\sum_{t=1}^{T-1}\|\bm f_{t+1,j}-\bm f_{t,j}\|_2
    }.
\end{equation}
The metric is computed directly in each representation space, with higher values indicating greater coherence of temporal feature changes.
Results are reported as mean GPE over evaluated videos.

\noindent\textbf{Quantitative results.}
As shown in \Tbref{tab:projection_trajectory_metrics}, the difference in mean GPE between $\bm B$ and $\bm A$ is smaller than that between $\bm B$ and $\bm C$ on all six datasets.
On DAVIS, for example, mean GPE is $0.1577$ for $\bm A$, $0.1690$ for $\bm B$, and $0.1540$ for $\bm C$.
These results show that the encoder features before projection are closer to the projected features in the temporal property measured by GPE than those of the model without projection.

We further compare $\bm A$ and $\bm C$ to assess the effect of temporal modeling on the original encoder representations.
The complete model obtains higher mean GPE on all six datasets.
Since neither $\bm A$ nor $\bm C$ passes through $\phi$ during evaluation, these differences are also present in the learned encoder features.
Together, these comparisons provide evidence that temporal variation modeling in a dedicated projected space also benefits the coherence of temporal feature changes in the original encoder representations.

\begin{table}[htbp]
    \centering
    \caption{Mean GPE of three representations across six datasets. Before projection (A) and after projection (B) are obtained from the same complete model. Without projection (C) denotes encoder representations from a separately trained model without projection.}
    \vspace{-3pt}
    \label{tab:projection_trajectory_metrics}
    \setlength{\tabcolsep}{8pt}
    \renewcommand{\arraystretch}{1.1}
    \resizebox{0.88\linewidth}{!}{%
    \begin{tabular}{l|cccccc}
        \toprule
        Representation & DAVIS & VIP & JHMDB & THUMOS14 & K400 val & SSV2 val \\
        \midrule
        Before projection ($\bm A$) & 0.1577 & 0.0972 & 0.2158 & 0.2541 & 0.1270 & 0.1396 \\
        After projection ($\bm B$) & 0.1690 & 0.0997 & 0.2288 & 0.2674 & 0.1339 & 0.1494 \\
        Without projection ($\bm C$) & 0.1540 & 0.0960 & 0.2131 & 0.2504 & 0.1252 & 0.1379 \\
        \bottomrule
    \end{tabular}%
    }
\end{table}

\begin{figure}[htbp]
    \centering
    \includegraphics[width=0.95\linewidth]{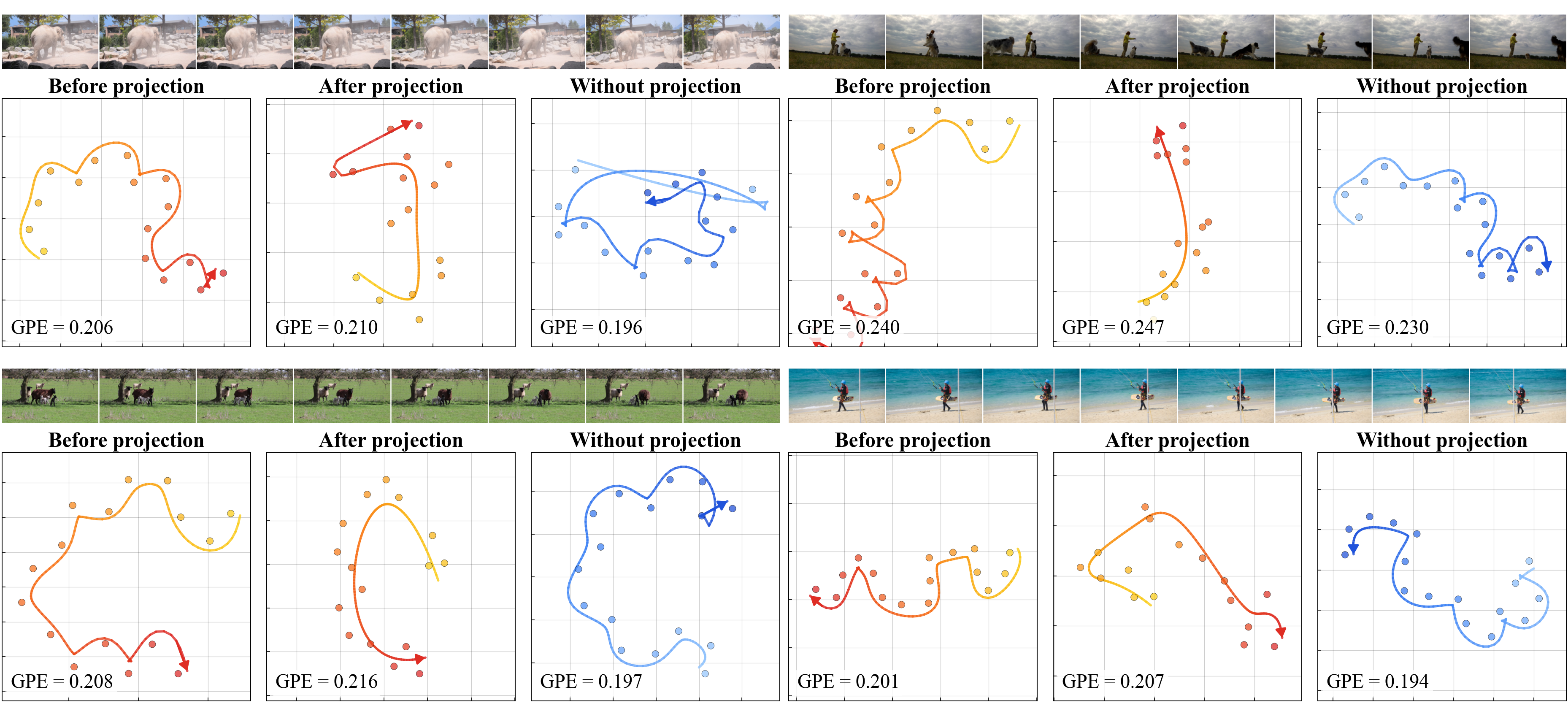}
    \par
    \caption{Temporal trajectory comparisons. Each example shows sampled video frames above three panels: before projection ($\bm A$) and after projection ($\bm B$) from the complete model, and without projection ($\bm C$) from the model without projection. Point colors indicate temporal order, and fitted curves serve as visual guides. For a fair comparison, all three representations use the same frames and embedding settings. The trajectories provide a qualitative reference for GPE computed in the original representation spaces.}
    \label{fig:projection_trajectory_comparison}
    \vspace{-8pt}
\end{figure}

\noindent\textbf{Trajectory visualizations.}
\Fgref{fig:projection_trajectory_comparison} presents the three representation trajectories for the same videos, with point colors indicating temporal order.
Patch features are averaged within each frame for visualization, while the accompanying GPE scores are computed from the individual patch trajectories in the original representation spaces.
The trajectories before projection (A) exhibit relatively orderly temporal progression, close to those after projection (B), whereas the baseline trajectories (C) show less regular temporal organization.
These qualitative observations are consistent with the GPE comparisons and suggest that temporal regularity is also reflected in the original encoder representations.

\subsubsection{Relation to Downstream Performance}
\label{sec:temporal_downstream_analysis}

The relevance of temporal variation modeling to downstream performance depends on the information required by each task and retained by its evaluation protocol. To examine this connection, we compare a \methodname{} variant with temporal variation modeling ablated against the complete \methodname{} model. Both use ViT-B/16 backbones pre-trained for 200 epochs. We evaluate their original encoder representations using our proposed Local Coherence (LC) and Trajectory Smoothness (TS) metrics. LC measures temporal correlation between consecutive frame representations, reflecting the coherence of local temporal evolution. TS characterizes changes in trajectory direction through turning angles across temporal intervals, with higher scores indicating smaller average angles and smoother trajectories. Their definitions are provided in \cref{app:app_quantitative}. \Tbref{tab:temporal_properties_ablation} reports the trajectory metrics, and \Tbref{tab:temporal_downstream_ablation} presents selected downstream evaluations across the four task categories. The complete model obtains higher LC and TS on all six datasets in the trajectory evaluation.

The second-order trajectory constraint links registered representation states to their first- and second-order temporal variation. Together with semantic self-distillation, it explicitly models coherent local state evolution while retaining semantic identity and informative state changes. We discuss how these properties relate to the task requirements below.

\noindent\textbf{Pixel-level tasks.}
Video object segmentation on DAVIS, human part segmentation on VIP, and pose propagation on JHMDB use frozen representations to propagate first-frame annotations. These tasks rely on maintaining object and part identity despite motion and appearance changes. Together, semantic self-distillation and temporal variation modeling encourage semantically consistent and temporally coherent patch representations, which can support reliable feature matching and label propagation. The complete model improves $\mathcal{J}\&\mathcal{F}_{m}$ on DAVIS from 65.3 to 67.0, mIoU on VIP from 39.4 to 40.6, and PCK@0.1 on JHMDB from 46.8 to 47.7. These gains accompany higher LC and TS on all three datasets, consistent with the relevance of coherent representation evolution to correspondence-based propagation.

\noindent\textbf{Frame-level tasks.}
Action segmentation and temporal localization require coherent representations within action stages and distinguishable changes at their boundaries. Temporal variation modeling is intended to organize representation evolution while retaining informative state transitions, providing useful cues for recognizing action stages and delimiting temporal intervals. On EgoProceL, Edit increases from 65.5 to 70.6 and frame accuracy including background (AccB) from 65.8 to 69.6. These improvements are qualitatively consistent with the relevance of structured state evolution to action-stage ordering and frame-level prediction.

\noindent\textbf{Video-level tasks.}
Action classification and video retrieval use spatiotemporal average pooling to obtain global representations. Temporal variation modeling encourages coherent features of successive visual states, which can help the aggregated representation capture consistent action semantics across the video. On HMDB51, R@1 increases from 71.0 to 71.5 and R@10 from 88.5 to 88.8. Chiral recognition concatenates frame-level representations in temporal order, preserving the sequence of visual states for the classifier. The explicit trajectory constraint organizes the evolution of these states, providing useful cues for distinguishing temporally ordered visual patterns. On Chiral SSV2, recognition accuracy increases from 83.23 to 84.19.

\noindent\textbf{Vision-based robotic learning.}
Robotic imitation learning connects visual state evolution to action prediction through behavior cloning with frozen visual representations. In manipulation tasks, changes in object configurations, manipulator positions, and interaction states characterize the progression of task execution and inform the required actions. Temporal variation modeling encourages coherent representations of this progression while retaining distinguishable states. Such representations can help the policy associate successive visual states with appropriate demonstrated actions. For Adroit, we average the peak evaluation success rates over three seeds for each task and report their mean across \texttt{Pen} and \texttt{Relocate}. The success rate increases from 77.3 to 78.7 on \texttt{Pen} and from 30.7 to 41.3 on \texttt{Relocate}, raising the mean from 54.0 to 60.0. These results are consistent with the relevance of representing evolving visual states to action prediction.

In summary, the higher Local Coherence and Trajectory Smoothness scores of the complete \methodname{} model indicate that temporal trajectory regularization improves temporal representation properties, which contribute to better performance across the four downstream task categories.

\begin{table}[htbp]
    \centering
    \caption{Comparison of temporal representation properties between the complete \methodname{} model and a variant without temporal variation modeling (Baseline). Both models use ViT-B/16 backbones pre-trained for 200 epochs.}
    \vspace{-3pt}
    \label{tab:temporal_properties_ablation}
    \setlength{\tabcolsep}{6pt}
    \renewcommand{\arraystretch}{1.15}
    \resizebox{0.98\linewidth}{!}{%
    \begin{tabular}{l|cc|cc|cc|cc|cc|cc}
        \toprule
        \multirow{2}{*}{Model} & \multicolumn{2}{c|}{DAVIS} & \multicolumn{2}{c|}{VIP} & \multicolumn{2}{c|}{JHMDB} & \multicolumn{2}{c|}{THUMOS14} & \multicolumn{2}{c|}{K400 val.} & \multicolumn{2}{c}{SSV2 val.} \\
        \cmidrule(lr){2-3}\cmidrule(lr){4-5}\cmidrule(lr){6-7}\cmidrule(lr){8-9}\cmidrule(lr){10-11}\cmidrule(lr){12-13}
        & LC & TS & LC & TS & LC & TS & LC & TS & LC & TS & LC & TS \\
        \midrule
        Baseline & 0.5698 & 0.4049 & 0.2529 & 0.3601 & 0.6758 & 0.4360 & 0.6797 & 0.4458 & 0.4040 & 0.3792 & 0.5017 & 0.3901 \\
        \methodname{} & \textbf{0.5802} & \textbf{0.4075} & \textbf{0.2609} & \textbf{0.3610} & \textbf{0.6787} & \textbf{0.4375} & \textbf{0.6814} & \textbf{0.4468} & \textbf{0.4129} & \textbf{0.3805} & \textbf{0.5080} & \textbf{0.3913} \\
        \bottomrule
    \end{tabular}%
    }
\end{table}

\begin{table}[htbp]
    \centering
    \caption{Comparison of downstream performance (\%) across four task categories between the complete \methodname{} model and a variant without temporal variation modeling (Baseline). Both models use ViT-B/16 backbones pre-trained for 200 epochs.}
    \vspace{-3pt}
    \label{tab:temporal_downstream_ablation}
    \setlength{\tabcolsep}{4pt}
    \renewcommand{\arraystretch}{1.15}
    \resizebox{0.98\linewidth}{!}{%
    \begin{tabular}{l|ccc|ccc|ccc|ccc}
        \toprule
        \multirow{3}{*}{Model} & \multicolumn{3}{c|}{\textbf{Pixel-level}} & \multicolumn{3}{c|}{\textbf{Frame-level}} & \multicolumn{3}{c|}{\textbf{Video-level}} & \multicolumn{3}{c}{\textbf{Robotic learning}} \\
        \cmidrule(lr){2-4}\cmidrule(lr){5-7}\cmidrule(lr){8-10}\cmidrule(lr){11-13}
        & DAVIS & VIP & JHMDB & \multicolumn{3}{c|}{EgoProceL} & \multicolumn{2}{c}{HMDB51} & Chiral SSV2 & \multicolumn{3}{c}{Adroit} \\
        & $\mathcal{J}\&\mathcal{F}_{m}$ & mIoU & PCK@0.1 & Edit & AccB & F1@10 & R@1 & R@10 & Acc. & \texttt{Pen} & \texttt{Relocate} & Avg. \\
        \midrule
        Baseline & 65.3 & 39.4 & 46.8 & 65.5 & 65.8 & \textbf{60.7} & 71.0 & 88.5 & 83.23 & 77.3 & 30.7 & 54.0 \\
        \methodname{} & \textbf{67.0} & \textbf{40.6} & \textbf{47.7} & \textbf{70.6} & \textbf{69.6} & 60.3 & \textbf{71.5} & \textbf{88.8} & \textbf{84.19} & \textbf{78.7} & \textbf{41.3} & \textbf{60.0} \\
        \bottomrule
    \end{tabular}%
    }
\end{table}

\subsection{Additional Visualizations of Downstream Tasks}

We provide additional qualitative comparisons on four downstream tasks, including video object segmentation, human pose estimation, video retrieval, and vision-based robotic learning.
These visualizations illustrate how the learned representations affect dense correspondence, semantic discrimination, and state-related temporal modeling.

\begin{figure*}
  \centering
    \includegraphics[width=0.98\linewidth]{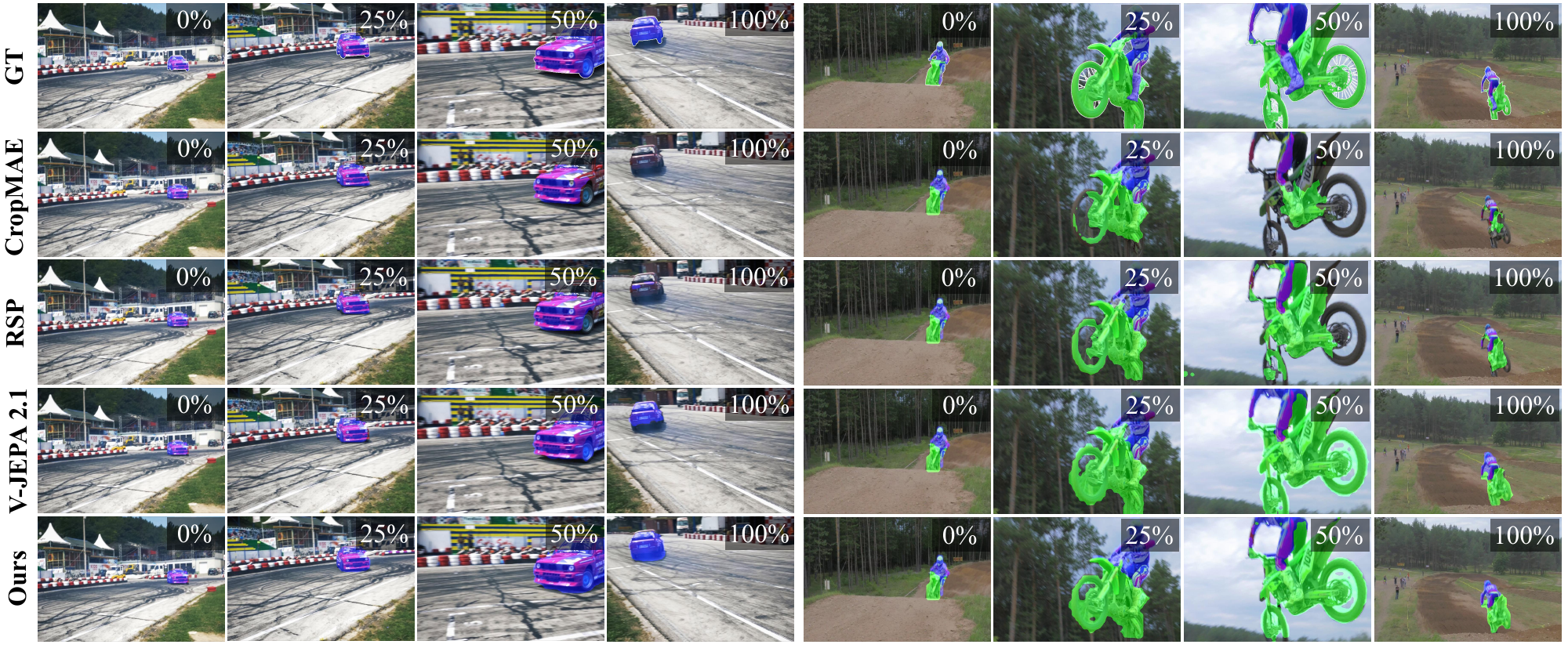}
    \caption{Additional visualization results of video object segmentation on DAVIS-2017.}
    \vspace{-10pt}
    \label{fig:vis_downstream_app_1}
\end{figure*}

For video object segmentation on DAVIS-2017, as shown in \Fgref{fig:vis_downstream_app_1}, \methodname{} produces coherent mask propagation across frames.
Compared with baseline methods, our results show clearer object regions and less drift under object motion, scale changes, and appearance variation.
This indicates that the learned dense tokens preserve foreground discrimination while maintaining reliable cross-frame correspondence.

\begin{figure*}
  \centering
    \includegraphics[width=0.98\linewidth]{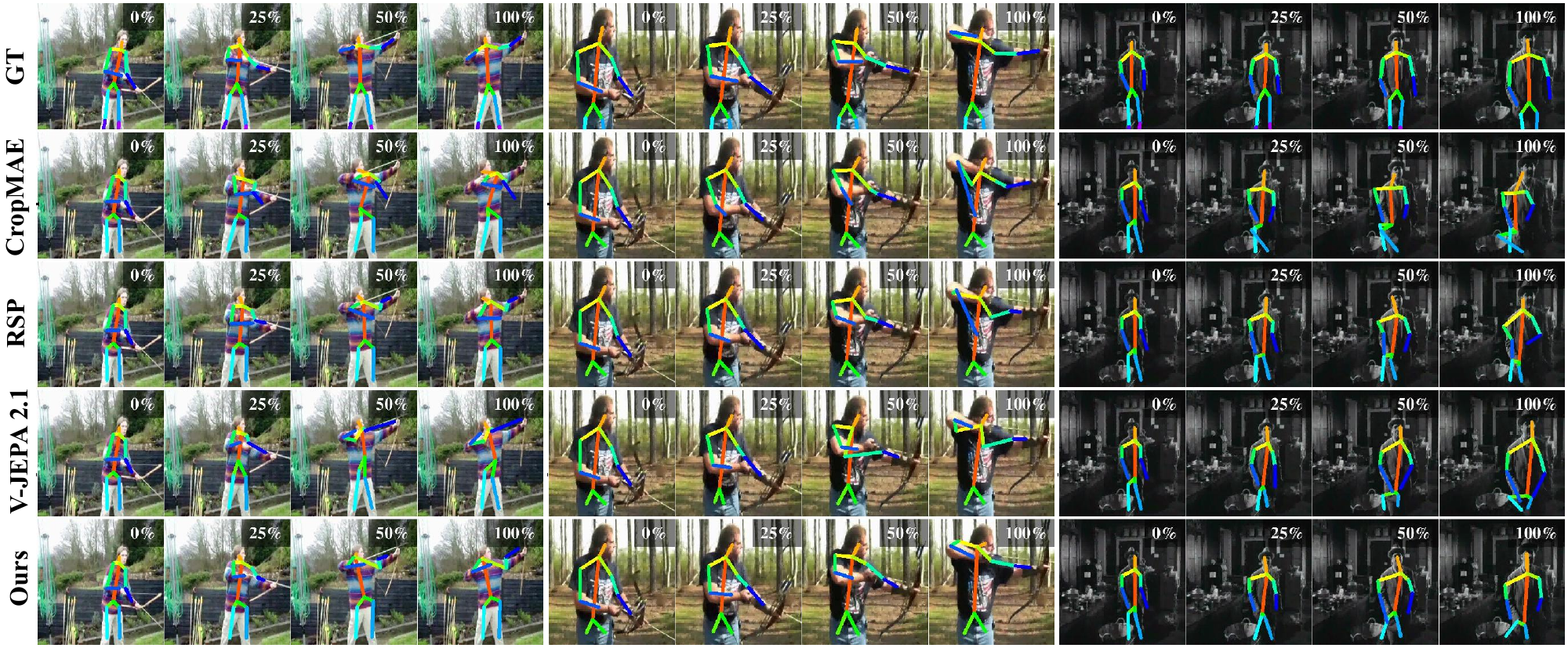}
    \caption{Additional visualization results of human pose estimation on JHMDB.}
    \vspace{-10pt}
    \label{fig:vis_downstream_app_2}
\end{figure*}

For human pose estimation on JHMDB, as shown in \Fgref{fig:vis_downstream_app_2}, \methodname{} yields more reasonable body structures and more stable keypoint propagation.
The predicted skeletons better follow articulated motion and preserve the spatial layout of human parts, especially when limbs undergo continuous deformation.
This suggests that the proposed temporal correspondence modeling benefits fine-grained part-level representation learning.

\begin{figure*}
  \centering
    \includegraphics[width=0.98\linewidth]{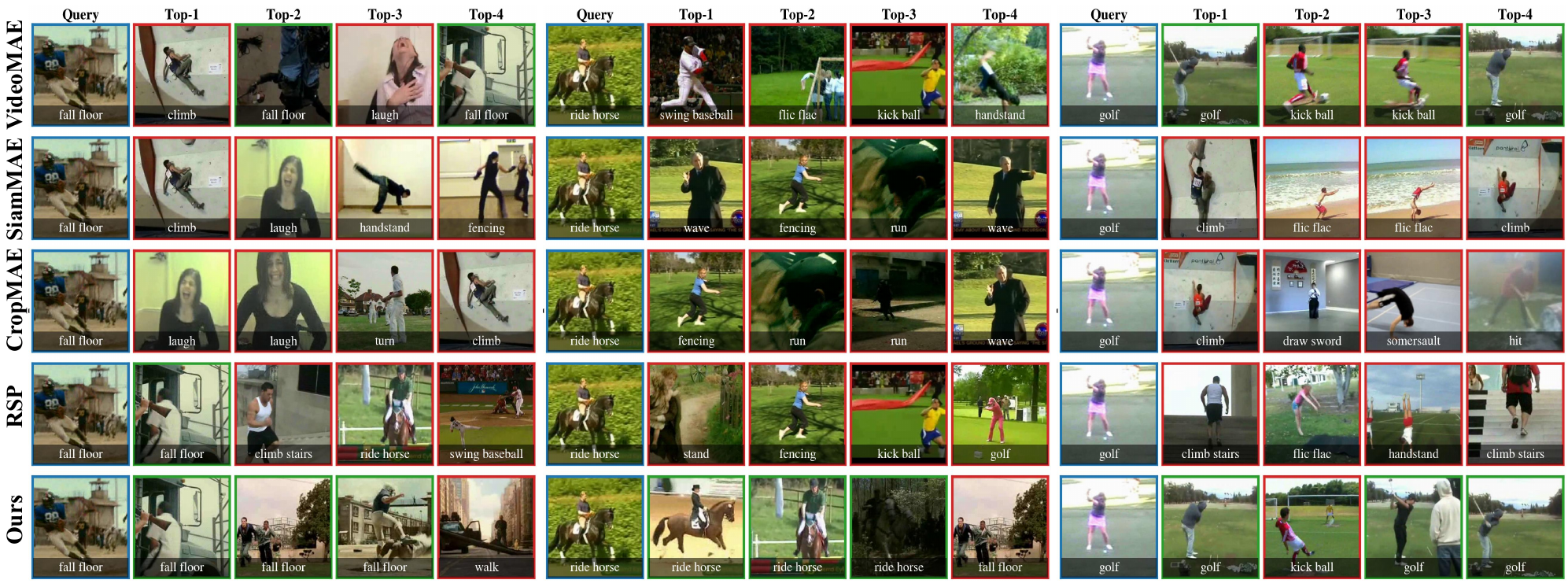}
    \caption{Additional visualization results of video retrieval on HMDB51.}
    \vspace{-10pt}
    \label{fig:vis_downstream_app_3}
\end{figure*}

For video retrieval on HMDB51, as shown in \Fgref{fig:vis_downstream_app_3}, \methodname{} retrieves videos with more consistent action semantics.
The retrieved samples are better aligned with the query in terms of motion patterns and temporal evolution, and are less affected by background similarity.
This shows that the learned representation captures action-related temporal changes beyond static appearance cues.

\begin{figure*}[t]
  \centering
    \includegraphics[width=0.98\linewidth]{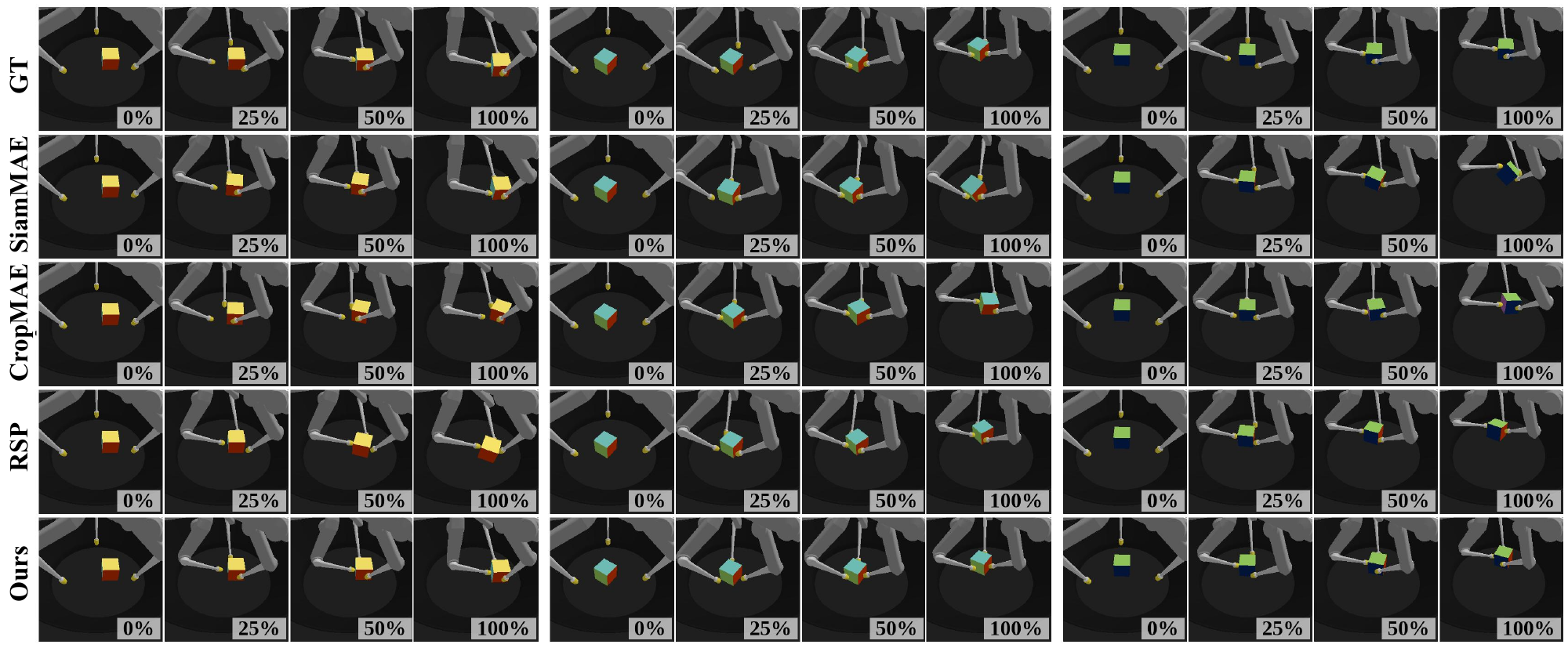}
    \caption{Additional visualization results of vision-based robotic learning on TriFinger.}
    \vspace{-10pt}
    \label{fig:vis_downstream_app_4}
\end{figure*}

For vision-based robotic learning on TriFinger, as shown in \Fgref{fig:vis_downstream_app_4}, \methodname{} produces predictions closer to the ground-truth robot configuration and final object state.
Compared with baseline representations, our results better preserve the state trajectory induced by robot-object interaction.
This suggests that variation-aware temporal modeling can provide state-sensitive representations for vision-based robotic learning.

Overall, these qualitative results show that \methodname{} provides reliable temporal correspondence, dense semantic discrimination, and structured temporal representations across diverse downstream scenarios.

